\documentclass[11pt]{article}
\usepackage[utf8]{inputenc}
\usepackage{amsfonts,fullpage}

\usepackage{amsmath}
\usepackage{graphicx,psfrag,epsf}
\usepackage{enumerate}

\usepackage{hyperref}
\hypersetup{
    colorlinks=true,
    linkcolor=blue,
    citecolor=blue,
    filecolor=magenta,      
    urlcolor=cyan,
}

\usepackage[round,colon,authoryear]{natbib}

\usepackage{url} 

\usepackage{amsthm,amssymb}
\usepackage{mathrsfs}
\usepackage{algorithm}
\usepackage{algpseudocode}
\usepackage{float} 
\usepackage{booktabs}
\usepackage{multirow}
\usepackage{amsfonts}
\usepackage{bbm}
\usepackage{longtable}

\newtheorem{theorem}{Theorem}
\newtheorem{lemma}{Lemma}
\newtheorem{remark}{Remark}

\newtheorem{assumption}{Assumption}

\def\calB{\mathcal{B}}
\def\calC{\mathcal{C}}
\def\calE{\mathcal{E}}
\def\calF{\mathcal{F}}
\def\calG{\mathcal{G}}
\def\calJ{\mathcal{J}}
\def\calL{\mathcal{L}}
\def\calN{\mathcal{N}}
\def\calU{\mathcal{U}}
\def\calX{\mathcal{X}}
\def\calY{\mathcal{Y}}

\def\EE{\mathbb{E}}
\def\RR{\mathbb{R}}

\newcommand{\bfm}[1]{\ensuremath{\mathbf{#1}}}

   \def\bA{\bfm A}  
\def\bb{\bfm b}

\def\be{\bfm e}     \def\EE{\mathbb{E}}
\def\bff{\bfm f}  \def\bF{\bfm F}  
\def\bg{\bfm g}     
   \def\bH{\bfm H}  
   \def\bI{\bfm I}  \def\II{\mathbb{I}}

   \def\MM{\mathbb{M}}
 \def\bmm{\bfm m}

   \def\bP{\bfm P}  \def\PP{\mathbb{P}}
     
\def\br{\bfm r}     \def\RR{\mathbb{R}}

\def\bv{\bfm v}     
   \def\bW{\bfm W}  
\def\bx{\bfm x}   \def\bX{\bfm X}  
\def\by{\bfm y}     
   \def\bZ{\bfm Z}  
\def\bX{\bfm X}

\def\bSigma{\bfm \Sigma}
\def\btheta{\bfm \theta}

\def\bDelta{\bfm\Delta}
\def\bdelta{\bfm\delta}

\def\bdelta{\boldsymbol{\delta}}

\def\tr{\mathsf{tr}}
\def\diag{\mathsf{diag}}
\def\var{\mathsf{Var}}
\def\svar{\mathsf{SVar}}
\def\cov{\mathsf{Cov}}
\def\scov{\mathsf{SCov}}
\def\con{\text{\tiny\sffamily con}}
\def\ppi{\text{\tiny\sffamily PPI}}
\def\moe{\text{\tiny\sffamily MOE}}
\def\grid{\text{\tiny\sffamily grid}}

\def\tilde{\widetilde}
\def\hat{\widehat}

\def\eps{\varepsilon}

\begin{document}



  \title{Prediction-powered Inference by Mixture of Experts}
  \author{Yanwu Gu$^1$, Linglong Kong$^2$, Dong Xia$^1$\thanks{
     Xia's research is partially supported by Hong Kong RGC grant GRF 16303224.}\\
   {\small $^1$ Department of Mathematics,   Hong Kong University of Science and Technology}\\
   {\small $^2$ Department of Mathematical and Statistical Sciences, University of Alberta}
}
\date{(\today)}
  \maketitle

\bigskip
\begin{abstract}

The rapidly expanding artificial intelligence (AI) industry has produced diverse yet powerful prediction tools, each with its own network architecture, training strategy, data-processing pipeline, and domain-specific strengths. These tools create new opportunities for semi-supervised inference, in which labeled data are limited and expensive to obtain, whereas unlabeled data are abundant and widely available. Given a collection of predictors, we treat them as a mixture of experts (MOE) and introduce an MOE-powered semi-supervised inference framework built upon prediction-powered inference (PPI). Motivated by the variance reduction principle underlying PPI, the proposed framework seeks the mixture of experts that achieves the smallest possible variance. 
Compared with standard PPI, the MOE-powered inference framework adapts to the \emph{unknown} performance of individual predictors, benefits from their collective predictive power, and enjoys a best-expert guarantee. The framework is flexible and applies to mean estimation, linear regression, quantile estimation, and general M-estimation. We develop non-asymptotic theory for the MOE-powered inference framework and establish upper bounds on the coverage error of the resulting confidence intervals. Numerical experiments demonstrate the practical effectiveness of MOE-powered inference and corroborate our theoretical findings.

\end{abstract}

\noindent%
{\it Keywords:}  Semi-supervised Learning, Model Averaging, Ensemble Learning, Variance Reduction.
\vfill

\section{Introduction}
\label{sec:intro}

Semi-supervised learning \citep{zhu2005semi,chapelle2009semi} is a machine learning paradigm that aims to integrate both labelled and unlabelled data to improve overall prediction accuracy and inferential efficiency. Suppose we observe i.i.d. labelled data $\calL=\{(X_i, Y_i)\}_{i=1}^n\subset \calX\times \calY$, drawn from a joint distribution $\PP_{X,Y}$, and i.i.d. unlabelled data $\calU=\{\tilde X_i\}_{i=1}^N$, where $X_i$ and $\tilde X_i$ share the same marginal distribution $\PP_X$. Since unlabelled data are often much cheaper to obtain than labelled data in many applications, the unlabelled sample size typically far exceeds the labelled sample size; that is, $N \gg n$. Because the unlabelled data provide abundant information about $\PP_X$, inferential efficiency can potentially be improved whenever the target parameter depends on the covariate distribution \citep{chakrabortty2018efficient}. More specifically, let $\theta_{\ast}=\theta(\PP_{X,Y})=\theta\big(\PP_X,\PP_{Y\mid X}\big)$ denote the target parameter. Then, whenever $\theta$ depends on $\PP_X$, one may \emph{potentially} improve inferential efficiency by leveraging the unlabelled data. For example, \citet{zhang2019semi} studied semi-supervised inference for the population mean $\EE Y$ and showed that even a simple linear-model-based correction can outperform the labelled-only sample mean, especially when $Y$ and $X$ are strongly linearly related. When the covariate $X$ is high-dimensional, inferential efficiency for $\EE Y $ can still be improved by incorporating unlabelled data through sparse linear regression methods \citep{zhang2022high}. See also \citet{tony2020semisupervised, liu2025semi, xu2025unified}.

The effectiveness of semi-supervised inference also depends on the postulated model for $\PP_{Y \mid X}$. For instance, \citet{zhang2019semi} and \citet{zhang2022high} studied linear models and showed that their proposed semi-supervised inference methods achieve lower variance when the linear relationship between $Y$ and $X$ is strong. This observation suggests that the efficiency of semi-supervised inference may be further improved when more powerful predictive models for $Y|X$ are available. Deep learning methods, or artificial intelligence more broadly, provide a rich class of such predictive tools. Prediction-powered inference (PPI; \citealp{angelopoulos2023prediction}) is a general semi-supervised inference framework that leverages both unlabelled data and a powerful predictor $f: \calX \to \calY$. The predictor is regarded as \emph{powerful} in the sense that $Y \approx f(X)$. For example, this can be formalized by requiring that $\var\!\bigl(Y - f(X)\bigr)$ be much smaller than $\var(Y)$ in mean value inference; that is, $f(X)$ explains a substantial fraction of the variation in $Y$.
There are two key components in PPI: the \emph{imputation} of outcomes for the unlabelled data, yielding pseudo-pairs $\{(\tilde X_i, \tilde Y_i)\}_{i=1}^N$ with $\tilde Y_i = f(\tilde X_i)$, and the \emph{rectifier}, constructed from the labelled data through the residuals $\{Y_i - f(X_i)\}_{i=1}^n$. Because the imputed values $f(\tilde X_i)$ may be biased, the rectifier serves to debias the resulting imputation-assisted estimators. PPI is a flexible framework that applies to a variety of classical statistical problems, including mean estimation, linear regression, logistic regression, and general $M$-estimation. See also \citet{angelopoulos2023ppi++, zrnic2024cross}.

It is often the case that the predictor $f(\cdot)$ is obtained from an off-the-shelf machine learning method trained on a massive dataset whose distribution does not necessarily match $\PP_{X,Y}$. Because of the potential distribution shift, there is no guarantee that a predictor $f(\cdot)$ that performs well on the training data will still  be powerful on the target labelled data $\calL$. This issue is especially pronounced in deep learning methods,  where networks are often highly overparameterized \citep{allen2019convergence}.  Although a preliminary check of $\var\big(Y-f(X)\big)$ may be convenient before applying PPI in some settings (e.g., mean estimation), it can be prohibitively time-consuming in others (e.g., M-estimation). Moreover, the rapidly evolving and expanding AI industry has produced a diverse array of powerful prediction tools, many of which are publicly or commercially accessible. These tools may be built on distinct network architectures tailored to specific data types, practical applications, or domain knowledge, and they often produce different predictions for the same covariate input. Large language models (LLMs), for example, illustrate this diversity well: GPT \citep{achiam2023gpt}, Claude, Gemini \citep{team2023gemini}, Qwen \citep{yang2025qwen3}, DeepSeek, and Kimi each have their own strengths, i.e., being an expert in different domains. This raises natural and crucial questions: \emph{is it possible to enhance the inference efficiency when multiple predictors $\{f_k(\cdot)\}_{k=1}^K$ are available whose performance on the target labelled data $\calL$ is unknown or expensive to evaluate? Is it possible to generalize the PPI framework so that it can robustly incorporate additional predictors whenever they are available? }

Treating each predictor $f_k(\cdot)$ as a potentially domain-specific expert, we propose a mixture-of-experts (MOE) PPI framework. MOE \citep{jacobs1991adaptive,dai2024deepseekmoe} is a general framework for predictive modeling in which multiple specialized predictors $\{f_k(\cdot)\}_{k=1}^K$ are combined adaptively to produce stronger predictions. For example, a linear mixture of experts takes the form $
F_{\beta}(x)=\sum_{k=1}^K \beta_k(x) f_k(x)$, where the weight vector $\beta(x)=(\beta_1(x),\cdots,\beta_K(x))^{\top}$ varies with the input covariate. The central idea of MOE is to decompose a complex prediction problem into simpler subproblems, each handled by an expert with localized competence.

Motivated by the variance reduction principle underlying PPI, our MOE-powered inference framework seeks a mixture of experts $F_{\beta}$ such that PPI achieves the smallest possible variance,
$
\var\big(Y - F_{\beta}(X)\big).
$
This MOE-PPI framework enjoys several advantages including collective prediction power, best-expert guarantee, and safe expert expansion, which we shall detail in Section~\ref{sec:m-est}.  Learning the optimal weight function $\beta(x)$ is challenging and may itself introduce bias and substantial variability, especially when the dimension of $\calX$ is high. For simplicity, we therefore focus primarily on global weighting, that is, $\beta(x)\equiv \beta$. We discuss localized weighting in Section~\ref{sec:discuss}. 
Note that global weighting reduces MOE to classical model averaging in the statistical literature \citep{claeskens2008model,raftery1997bayesian}.

Our contributions are as follows. We propose a flexible MOE-powered inference framework for prediction-powered inference that leverages the collective predictive power of multiple predictors. We define the oracle mixture of experts as the combination that minimizes the variance of PPI-based estimators. Notably, unlike classical model averaging, which typically focuses on mean squared error \citep{claeskens2008model}, the oracle MOE is designed specifically to minimize variance. This distinction is natural in the PPI framework, since the rectification step already removes bias. The oracle MOE generally achieves smaller variance than PPI based on any single predictor.  We estimate the oracle mixture weights by minimizing the sample (co)variance  on the labeled data and then construct confidence sets using the resulting sample MOE. Unlike standard PPI-based estimators, which are unbiased, the MOE-powered estimator is generally biased because the oracle MOE must be estimated from the labelled data. Nevertheless, we show that this bias is negligible and, in many applications, is dominated by the standard deviation. 
Our MOE-powered inference framework is highly flexible and applies to a broad range of problems, including M-estimation, mean estimation, linear regression, and quantile estimation. We develop a non-asymptotic theory for the proposed framework. Under mild conditions, we show that the bias of the MOE-powered estimator converges at the rate $O(n^{-1})$. We also establish a Berry--Esseen bound for the normal approximation of the MOE-powered estimator, which yields non-asymptotic guarantees for the coverage probabilities of the resulting confidence sets. Moreover, the constructed confidence intervals have widths comparable to those based on the oracle MOE. Finally, we present comprehensive numerical experiments demonstrating that the proposed MOE framework generally outperforms standard PPI in terms of inferential efficiency.

The rest of the paper is organized as follows. In Section~\ref{sec:m-est}, we introduce the MOE-powered inference framework for general M-estimation, describe the construction of the confidence set, and provide theoretical guarantees for its coverage probability. Section~\ref{sec:app} presents specific applications to mean estimation, quantile estimation, linear regression, and logistic regression, accompanied with corresponding theoretical guarantees. Comprehensive simulation studies and real-world data experiments are presented in Section~\ref{sec:experiments}. We discuss possible extensions in Section~\ref{sec:discuss}. All proofs and technical lemmas are deferred to the Appendix.

\section{MOE-powered Inference for M-Estimation}\label{sec:m-est}

\subsection{Variance reduction by PPI}

M-estimation \citep{geer2000empirical} is a unified statistical framework for parametric estimation. Let $(X, Y)\in \calX\times \calY\subset \RR^d\times \RR$ be a pair of random variables and $\ell_{\theta}: \calX\times \calY\mapsto \RR$ be a loss function in $\theta\in\Theta\subset \RR^p$.  Denote the population minimizer
\begin{align}\label{eq:m-est-def}
\theta_{\ast}=\underset{\theta\in\Theta}{\arg\min}\ \EE \ell_{\theta}(X, Y),
\end{align}
where we assume $\Theta$ is a convex set.  Note that $\theta_{\ast}$ may represent a set if the minimizer is not unique. 
For simplicity, we assume that $\ell_{\theta}(\cdot,)$ is continuously differentiable with respect to $\theta$ and denote its gradient by $\bg_{\theta}(\cdot):=\nabla_{\theta}\ell_{\theta}(\cdot)$. Under the optimality, we can equivalently view $\theta_{\ast}$ as the solution to 
\begin{align}\label{eq:z-est-def}
\EE\big[ \bg_{\theta}(X, Y)\big]=0.
\end{align}
Let $\calL:=\{(X, Y), (X_1, Y_1),\cdots, (X_n, Y_n)\}$ be a collection of \emph{labelled} i.i.d. observations and $\calU:=\{\tilde X, \tilde X_1,\cdots, \tilde X_N\}$ be a collection of \emph{unlabelled} i.i.d. observations. Our goal is to construct valid $100(1-\alpha)\%$ confidence sets for $\theta_{\ast}$. Following the practice in \cite{angelopoulos2023ppi++}, we search for the candidate solutions from a \emph{fine-grained} grid $\Theta_{\grid}\subset \Theta$.

Let $f: \calX\mapsto \calY$ be a given ``powerful" \emph{expert} or \emph{predictor}. Motivated by (\ref{eq:z-est-def}) and inspired by \cite{angelopoulos2023prediction}, we define the sample gradient and \emph{imputed} gradient by 
$$
\hat \bg_{\theta}:=\frac{1}{n}\sum_{i=1}^n \bg_{\theta}(X_i, Y_i)\quad {\rm and}\quad \tilde \bg_{\theta, f}=\frac{1}{N}\sum_{i=1}^N \bg_{\theta}\big(\tilde X_i, f(\tilde X_i)\big),
$$
respectively. While the sample gradient $\hat \bg_{\theta}$ is unbiased,  the imputed one may have a non-zero (negative) bias $\bDelta_{\theta, f}:=\EE\big[\bg_{\theta}(X, Y)\big]-\EE \big[\bg_{\theta}(X, f(X))$, which is referred to as the \emph{rectifier} in \cite{angelopoulos2023ppi++}.  The \emph{empirical rectifier} is thus defined by
$$
\hat\bDelta_{\theta, f}:=\frac{1}{n}\sum_{i=1}^n\big(\bg_{\theta}(X_i, Y_i)-\bg_{\theta}(X_i, f(X_i))\big).
$$
Both the sample gradient $\hat\bg_{\theta}$ and the rectified imputed gradient $\tilde \bg_{\theta, f}+\hat\bDelta_{\theta,f}$ are unbiased,  yet their covariances can be strikingly different.  Denote $\bW_{\theta,Y}=\cov\big(\bg_{\theta}(X, Y)\big)$,  $\bW_{\theta,f}=\cov\big(\bg_{\theta}(X, f(X))\big)$, and $\bW_{\theta,Y-f}=\cov\big(\bg_{\theta}(X,Y)-\bg_{\theta}(X,f(X))\big)$, respectively. Then,
$$
\cov\big(\hat\bg_{\theta}\big)=\frac{\bW_{\theta, Y}}{n}\quad {\rm and}\quad \cov\big(\tilde \bg_{\theta,f}+\hat\bDelta_{\theta,f}\big)=\frac{\bW_{\theta, f}}{N}+\frac{\bW_{\theta, Y-f}}{n},
$$
implying that the rectified imputed gradient achieves a smaller covariance if $N\gg n$ and $\bW_{\theta,Y- f}\leq \bW_{\theta, Y}$. For brevity, we abuse the notation ``$\leq$" for matrices which, depending on the context, may represent comparison w.r.t. their traces or  diagonal entries one-by-one.

Motivated by the asymptotic normality of $n^{1/2}\hat\bg_{\theta}$ and $n^{1/2}\hat\bDelta_{\theta,f}$, the conventional and PPI-based approaches construct the confidence sets by 
\begin{align}\label{eq:m-est-con-ppi-ci}
\calC_{\alpha}^{\con}:=&\Big\{\theta\in\Theta_{\grid}: \big|(\hat\bg_{\theta})_j \big|\leq  z_{\alpha/(2p)}\sqrt{(\hat\bW_{\theta, Y})_{jj}/n},\ \forall j\in[p]\Big\};\notag\\
\calC_{\alpha,f}^{\ppi}:=&\Big\{\theta\in\Theta_{\grid}: \big|\big(\tilde\bg_{\theta,f}+\hat\bDelta_{\theta,f}\big)_{j} \big|\leq z_{\alpha/(2p)}\sqrt{(\hat\bW_{\theta,Y-f})_{jj}/n},\ \forall j\in[p] \Big\},
\end{align}
where the $z$-critical value $z_{\alpha}:=\Phi^{-1}(1-\alpha)$ and $\hat\bW_{\theta, Y}$ and $\hat\bW_{\theta,Y-f}$ represent the sample estimates of $\bW_{\theta, Y}$ and $\bW_{\theta,Y-f}$, respectively. See Section~\ref{sec:m-est-oracle} for more details. 
 It was shown by \cite{angelopoulos2023ppi++} that, under mild conditions, the coverage probability of $\calC_{\alpha,f}^{\ppi}$ is at least $1-\alpha$ asymptotically. The advantage of PPI-based approach is reliant on the assumption that the expert is powerful in predicting $Y$ so that $\bW_{\theta,Y-f}\leq \bW_{\theta, Y}$.  This assumption may not hold in practice, or it may be costly to verify. Fortunately, it is common that multiple experts are available,  e.g., AI models released by different companies. Our MOE-powered inference framework can leverage the prediction power of multiple experts and outperform the conventional one as long as one expert provides accurate prediction.

\subsection{Oracle MOE and its estimate}\label{sec:m-est-oracle}

Let $f_1(\cdot),\cdots, f_K(\cdot): \calX\mapsto \calY$ be a collection of available experts. 
We consider the mixture of experts (MOE) through an ensemble operator  $\mathscr G(\cdot; \beta)$, parameterized by $\beta \in \mathcal B \subseteq \RR^q$, 
\begin{align}\label{equ:function_space_def}
	\calF_{\calB}:=&\big\{F_\beta=\mathscr G(f_1,\cdots, f_K;\beta):F_\beta(x)\in C^2(\mathcal B^{\circ}),\forall x\in\mathcal{X}\big\}.
\end{align}
Here we assume the set $\calB$ is convex and $\mathscr{G}(f_1,\cdots,f_K, \beta)$ is twice continuously differentiable w.r.t. the weight vector $\beta\in\calB$.  
Note that $\calB^{\circ}$ denotes the interior if $\calB$ is closed, and $q$ is a constant much smaller than $n$, indicating the complexity of the candidate space is limited. The function family $\calF_{\calB}$ covers many important examples when choosing $q=K$:
\begin{itemize}
	\item[1).] {\it Linear mixture:} $F_{\beta}(\cdot)=\sum_{k=1}^K\beta_k f_k(\cdot)$ for $\beta\in\calB=\RR^K$;
	\item[2).] {\it $\ell_1$-compressed sensing:} $F_{\beta}(\cdot)=\sum_{k=1}^K\beta_k f_k(\cdot)$ for $\calB:=\{\beta\in\RR^K: \|\beta\|_{\ell_1}\leq L\}$;
	\item[3).] {\it Soft max}: $F_{\beta}(\cdot)=\log\big(\sum_{k=1}^K \exp(\beta_k f_k(\cdot))\big)$ for $\beta\in\calB=\RR^K$;
	\item[4).] {\it Logistic linear mixture:} $F_{\beta}(\cdot)=\psi\big(\sum_{k=1}^K \beta_kf_k(\cdot)\big)$ for $\beta\in\calB=\RR^K$ with a sigmoid function $\psi(u)=1/(1+e^{-u})$. 
\end{itemize}

Inspired by the variance-reduction principle, we call the weight vector that minimizes the variance of PPI-based estimators the \emph{oracle} MOE weight vector. More precisely, we define
\begin{align}\label{eq:m-est-beta-opt}
\beta_{\ast}:=\underset{\beta\in\calB}{\arg\min}\ \tr\big(\bW_{\theta, Y-F_{\beta}}\big).
\end{align}
For notational brevity, we suppress the dependence of $\beta_{\ast}$ on $\theta$. Correspondingly, we refer to $F_{\beta_{\ast}}$ as the \emph{oracle mixture of experts}, which enjoys several benign properties:
\begin{itemize}
\item[1).] {\it Collective prediction power:} By definition, the oracle MOE attains the smallest variance among all possible mixtures in the class. In particular, its variance is never larger than that of any PPI-based estimator that uses a single expert; that is $\bW_{\theta, Y-F_{\beta_{\ast}}}\leq \bW_{\theta, Y-f_k}$ for $\forall k\in[K]$. This variance-reduction guarantee highlights the key advantage of MOE-powered methods: they can exploit the collective predictive power of multiple experts. This advantage is especially appealing when different experts capture domain knowledge that is not widely shared. See the numerical experiments in Section~\ref{sec:experiments}.
\item[2).] {\it Best-expert guarantee:} The oracle MOE automatically achieves variance reduction at least as good as that of the PPI-based estimator built on the best single expert. Unlike standard PPI, the MOE approach does not require prior knowledge of the experts’ relative predictive performance. Put differently, the oracle MOE is robust to \emph{expert misspecification} in the sense that it can still deliver effective variance reduction as long as at least one expert provides strong predictions. 
\item[3).] {\it Safe expert expansion:} The variance of the oracle MOE does not increase as the expert pool expands. As the rapidly evolving AI industry continues to produce new expert models, these models can be safely incorporated into the MOE framework without requiring confidence check in their trustworthiness or predictive quality.
\end{itemize}

We define the \emph{unbiased} sample covariance matrix by 
\begin{align*}
\hat\bW_{\theta, Y-f}:=\frac{1}{n-1}\sum_{i=1}^n \big(\bg_{\theta}(X_i, Y_i)-\bg_{\theta}(X_i, f(X_i))\big) \big(\bg_{\theta}(X_i, Y_i)-\bg_{\theta}(X_i, f(X_i))\big)^{\top}-\frac{n}{n-1}\cdot\hat\bDelta_{\theta, f}\hat\bDelta_{\theta, f}^{\top},
\end{align*}
and estimate the oracle MOE weight by $\hat\beta_n:=\arg\min_{\beta\in\calB} \tr(\hat \bW_{\theta, Y-F_{\beta}})$.  It remains to establish non-asymptotic error rate for $\hat\beta_n-\beta_{\ast}$. 

For any fixed $\theta$,  let $\bmm_{\theta}(\beta, X,Y):=\bg_{\theta}(X, Y)-\bg_{\theta}(X, F_{\beta}(X))$.  As a result, we have $\hat\bDelta_{\theta,f}=n^{-1}\sum_{i=1}^n \bmm_{\theta}(\beta, X_i, Y_i)$  and $\EE \hat\bW_{\theta, Y-f}=\bW_{\theta, Y-f}$. 
Then we can write $\bW_{\theta, Y-F_{\beta}}=\cov\big(\bmm_{\theta}(\beta, X, Y)\big)$ and define $Q(\beta)=\tr\big(\bW_{\theta, Y-F_{\beta}}\big)$, where we suppress the dependence of $Q({\beta})$ on $\theta$ for notational brevity. Similarly, we write $Q_{n}(\beta):=\tr\big(\scov\big(\bmm_{\theta}(\beta, X, Y)\big)\big):=\tr\big(\hat\bW_{\theta,Y-F_{\beta}}\big)$. 

\begin{assumption}\label{assump:m-est}
Suppose that $\calB\subset \RR^q$ is compact with a fixed parameter dimension $q$ and a diameter $D>0$, and, for any $\theta\in\Theta$, $(X, Y)\in \calX\times \calY$, the function $\bmm_{\theta}(\beta, X_, Y)$ satisfies:
\begin{itemize}
\item[(a)] $\bmm_{\theta}(\beta, X, Y)\in C^2(\calB^{\circ})$ and $\|\bmm_{\theta}(\beta, X, Y)\|\leq U$;
\item[(b)] $\big\|\bmm_{\theta}(\beta_1, X, Y)-\bmm_{\theta}(\beta_2, X, Y) \big\|\leq \tau_1\|\beta_1-\beta_2\|$, for $\forall \beta_1, \beta_2\in \calB$;
\item[(c)] $Q(\beta)-Q(\beta_{\ast})\geq \tau_0\|\beta-\beta_{\ast}\|^2$ for $\forall \beta\in\calB$;
\end{itemize}
where $\tau_0, \tau_1, U>0$ are absolute constants.
\end{assumption}

Assumption~\ref{assump:m-est} is standard for  obtaining high-probability upper bound in empirical risk minimization. The uniform boundedness condition on $\bmm_{\theta}(\beta, X, Y)$ can be relaxed, e.g., one may instead assume that $\bmm_{\theta}(\beta, X, Y)$ is sub-Gaussian for any fixed $\theta$ and $\beta$.   The assumption of global strong convexity can also be weakened to a local condition, provided additional requirements hold, for instance, that $\beta_{\ast}$ is the unique minimizer of $Q(\beta)$. See Assumption~\ref{assump:quantile} in Section~\ref{sec:quantile}. We shall explore these refinements in future work.  

\begin{lemma}\label{lem:m-est-beta}
Suppose Assumption~\ref{assump:m-est} holds and $n\geq C_0(\tau_1^2/\tau_0)p\big(q+\log(Dn)\big)$ for a large $C_0>1$.  There exists a constant $C_1>0$ such that 
$$
\|\hat\beta_n-\beta_{\ast}\|^2\leq C_1U^2\bigg(\frac{\tau_0^{-1}\log n}{n}+\frac{\tau_1^2}{\tau_0^2}\cdot\frac{p(q+\log(Dn))}{n}\bigg),
$$
which holds with probability at least $1-n^{-9}$. 
\end{lemma}

\subsection{MOE-powered confidence sets}

Based on the estimated oracle MOE $F_{\hat\beta_n}(\cdot)$,  we construct the confidence set following the PPI approach (\ref{eq:m-est-con-ppi-ci}). For notational simplicity, let $\tilde\bg_{\theta, \hat\beta_n}$ and $\hat\bDelta_{\theta, \hat\beta_n}$ denote $\tilde\bg_{\theta, F_{\hat\beta_n}}$ and $\hat\bDelta_{\theta, F_{\hat\beta_n}}$, respectively. We then define the MOE-powered confidence set as
$$
\calC^{\moe}_{\alpha}:=\Big\{\theta\in\Theta_{\grid}: \big|\big(\tilde\bg_{\theta, \hat\beta_n}+\hat\bDelta_{\theta,\hat\beta_n}\big)_{j} \big|\leq z_{\alpha/(2p)}\sqrt{\big(\hat\bW_{\theta, Y-F_{\hat\beta_n}}\big)_{jj}/n}, \forall j\in[p] \Big\},
$$
where $\Theta_{\grid}$ denotes a fine-grained grid of $\Theta$. We note that the estimated MOE weight vector $\hat\beta_n$, as well as\footnote{As noted above, we suppress the dependence of $\hat\beta_n$ on $\theta$ for notational simplicity.} the oracle $\beta_{\ast}$, depends on $\theta$.  Therefore,   $\beta_{\ast}(\theta)$ must be re-estimated as $\theta$ ranges over $\Theta_{\grid}$. 

The detailed steps are summarized as in Algorithm~\ref{algo:convex_estimation}. If a point estimate of $\theta_{\ast}$ is desired, we propose 
$$
\hat\theta^{\moe}:=\arg\min_{\theta\in\Theta_{\grid}} \big\| \tilde\bg_{\theta, \hat\beta_n} +\hat\bDelta_{\theta,\hat\beta_n}\big\|,
$$
although we do not pursue the theoretical analysis of $\hat\theta^{\moe}$ here. 

\begin{algorithm}
\caption{MOE-powered Inference for M-Estimation}  
\label{algo:convex_estimation}
\begin{algorithmic}[1]
    \State {\bf Input:} labeled and unlabelled dataset $\{(X_i,Y_i)\}_{[n]}$, $\{\tilde X_i\}_{[N]}$, MOE family $\calF_{\calB}:=\{F_{\beta}, \beta\in\calB\}$, the gradient of loss function $\bg_{\theta}(\cdot, \cdot)$,  initial set   $\calC^{\moe}_{\alpha}=\emptyset$;
    \For{$\theta\in\Theta_{\text{grid}}$}
            \State{solve $\hat \beta_n=\arg\min_{\beta\in\mathcal B} \tr\big(\scov\big(\bg_{\theta}(X,Y)-\bg_{\theta}(X,F_{\beta}(X))\big)\big)$;}
        \State{imputed gradient: $\tilde \bg_{\theta, \hat\beta_n}\gets N^{-1}\sum_{i=1}^N \bg_\theta(\tilde X_i, F_{\hat\beta_n}(\tilde X_i))$;}
        \State{rectifier: $\hat \bDelta_{\theta,\hat\beta_n} \gets n^{-1}\sum_{i=1}^n \Big(\bg_\theta(X_i, Y_i)-\bg_\theta(X_i, F_{\hat\beta_n}(X_i))\Big)$;}
        
        \State{sample covariance: $\hat\bW_{\theta, Y-F_{\hat\beta_n}} \gets \scov\big(\bg_{\theta}(X,Y)-\bg_{\theta}(X, F_{\hat\beta_n}(X))\big)$;}
    \If{$\big|\be_s^\top\big(\tilde\bg_{\theta, \hat\beta_n}+\hat\bDelta_{\theta,\hat\beta_n}\big) \big|\leq z_{\alpha/(2p)}\sqrt{\be_s^\top\hat\bW_{\theta, Y-F_{\hat\beta_n}}\be_s/n}$ for $\forall s\in[p]$}
    \State{update: $\calC_{\alpha}^{\moe}\gets \calC_{\alpha}^{\moe}\cup \{\theta\}$;}
    \EndIf
        \EndFor
    \State \textbf{Output:} confidence set $\calC_{\alpha}^{\moe}$
\end{algorithmic}
\end{algorithm}

For a sequence of random variables $\{X_n\}$ and a deterministic sequence $\{\alpha_n\}$,   we write $X_n=\tilde O_p(\alpha_n)$ if $\PP\big(|X_n|\geq C\alpha_n\big)=O(n^{-2})$ for some constant $C>0$.  Recall that the classical notation $X_n=O_p(\alpha_n)$ means that,  $\forall \eps>0$,  there exists an $n_{\eps}>0$ such that $\PP\big(|X_n|\geq C\alpha_n\big)\leq \eps$ for all $n\geq n_{\eps}$.

\begin{theorem}[M-estimation]\label{thm:m-est}
Suppose Assumption~\ref{assump:m-est} holds and $n\geq C_1 p^2\big(q^2+\log^2(Dn)\big)$ and $N\geq C_1 pn\log n$ for a large constant $C_1>0$. Then, for any $\theta\in\Theta$, 
\begin{align}\label{eq:m-est-main}
    \sqrt n\Big(\tilde\bg_{\theta, \hat\beta_n} + \hat \bDelta_{\theta, \hat\beta_n} -\EE\bg_\theta(X,Y)\Big)=\bZ_{n,\theta,\beta_\ast}+\tilde O_p\bigg(\frac{p(q+\log(Dn))}{\sqrt{n}}+\sqrt{\frac{pn\log n}{N}}\bigg),
\end{align}
where $\bZ_{n,\theta, \beta_\ast}=n^{-1/2}\sum_{i=1}^n \big[\big(\bg_\theta(X_i,Y_i) - \bg_\theta(X_i, F_{\beta_\ast}(X_i))\big)-\big(\EE\bg_\theta(X,Y) - \EE\bg_\theta(X, F_{\beta_\ast}(X))\big)\big]$.

Given $\theta_\ast\in\Theta^\grid$, if $n\geq C_2p^4\big(q^2+\log^2(Dn)\big)$ and $N\geq C_2p^3n\log n$ for a large constant $C_2>0$, then the coverage probability satisfies
\begin{align}\label{eq:m-est-coverage}
    \PP(\theta_{\ast}\in\calC_{\alpha}^{\moe})\geq 1-\alpha+O\bigg(\frac{p^2(q+\log(Dn))}{\sqrt{n}}+\sqrt{\frac{p^3n\log(n)}{N}}\bigg).
\end{align}
\end{theorem}

We remark that the global uniqueness of $\theta_\ast$ is not required. The estimating equation $\EE[\bg_{\theta}(X,Y)]=0$ may have multiple solutions, the asymptotic normality and coverage statement in Theorem~\ref{thm:m-est} remain valid for all regular solutions. 
Moreover, the complexity of $\calF_{\calB}$, as measured by its dimension $q$, should be kept under control. In particular, 
$q$ should not much larger than $\log n$; otherwise, the resulting procedure may suffer from increased estimation error and reduced statistical efficiency.

By taking into consideration of the variance contributed by the unlabelled data, we can construct a slightly larger confidence set, which improves the remainder term $O(\sqrt {(pn/N)\log n})$ in (\ref{eq:m-est-main}) and allows a wider range of the size of unlabelled data. 

\begin{theorem}[M-estimation+]\label{thm:m-est+}
Suppose Assumption~\ref{assump:m-est} holds and $n\geq C_1 p^2\big(q^2+\log^2(Dn)\big)$ and $N\geq C_1p\log n$ for a large $C_1>0$. Then, for any $\theta\in\Theta$, 
\begin{align}\label{eq:m-est-main+}
    \sqrt n\Big(\tilde\bg_{\theta, \hat\beta_n} + \hat \bDelta_{\theta, \hat\beta_n} -\EE\bg_\theta(X,Y)\Big)=\bZ_{n,\theta,\beta_\ast}+\sqrt{\frac nN}\tilde\bZ_{N,\theta,\beta_\ast}+\tilde O_p\bigg(\frac{p\big(q+\log(Dn)\big)}{\sqrt{n}}\bigg).
\end{align}
where $\tilde \bZ_{N,\theta, \beta_\ast}=N^{-1/2}\sum_{i=1}^N \Big[\bg_\theta(\tilde X_i, F_{\beta_\ast}(\tilde X_i))- \EE\bg_\theta(X_i, F_{\beta_\ast}(X_i))\Big]$.
Let $\hat \bW_{\theta, F_{\hat\beta_n}}=\scov\big(\bg_\theta(X, F_{\hat\beta_n}(X))\big)$ and construct the confidence set as
\begin{align*}
    \tilde\calC_{\alpha}^{\moe} = \bigg\{\theta_\ast \in\Theta^\grid : \frac{\Big|\be_s^\top\Big(\tilde\bg_{\theta, \hat\beta_n} + \hat \bDelta_{\theta, \hat\beta_n}\Big) \Big|}{\sqrt{\be_s^\top \big(n^{-1}\hat \bW_{\theta, Y-F_{\hat\beta_n}}+N^{-1}\hat \bW_{\theta, F_{\hat\beta_n}}\big)}\be_s}\leq z_{\alpha/(2p)},\forall s\in[p]\bigg\}.
\end{align*}

Given $\theta_\ast\in\Theta^\grid$, if $n\geq C_2 p^4\big(q^2+\log^2(Dn)\big)$ and $N\geq C_2 p^2\log n$ for a large $C_2>0$, then the coverage probability satisfies 
$$
\PP(\theta_{\ast}\in\tilde\calC_{\alpha}^{\moe})\geq 1-\alpha+O\bigg(\frac{p^2\big(q+\log(Dn)\big)}{\sqrt{n}}+\sqrt{\frac{p^2\log(n)}{N}}\bigg).
$$
\end{theorem}

\begin{remark}\label{rmk:refined_version}
The oracle weight vector $\beta_\ast$ and its estimator $\hat\beta_n$ minimize the trace of $\bW_{Y-F_\beta}$ and $\hat\bW_{Y-F_\beta}$, respectively. From the perspective of Theorem~\ref{thm:m-est+}, a more refined choice is to minimize the trace of
    $$
    n^{-1}\bW_{Y-F_\beta}+N^{-1}\bW_{F_\beta}
    \quad\text{and}\quad
    n^{-1}\hat\bW_{Y-F_\beta}+N^{-1}\hat\bW_{F_\beta},
    $$
including the covariance contributed by the unlabelled data. 
 The corresponding oracle and empirical minimizers are denoted by $\beta_\ast^+$ and $\hat\beta_n^+$, respectively.  Theorem~\ref{thm:m-est+} continues to hold if these refined empirical estimates are employed.
\end{remark}

\section{Applications}\label{sec:app}

For ease of exposition, we focus on the linear mixture of experts throughout this section. More specifically, given $K$ experts $f_1(\cdot),\cdots, f_K(\cdot)$, the candidate MOE resides in 
$$
\calF:=\bigg\{F_{\beta}:=\sum_{k=1}^k\beta_k f_k, \beta\in\RR^K\bigg\}.
$$
Let $\calL:=\{(X, Y), (X_1, Y_1),\cdots, (X_n, Y_n)\}\subset \calX\times \calY$ be a collection of \emph{labelled} i.i.d. observations and $\calU:=\{\tilde X, \tilde X_1,\cdots, \tilde X_N\}$ be a collection of \emph{unlabelled} i.i.d. observations, which is independent from $\calL$.  Moreover,  all covariates $\tilde X$ and $X$ are identically distributed and $\calX\subset \RR^d$.  

\subsection{Mean Value Inference}\label{sec:moe-mean}
Mean value estimation is perhaps the simplest yet most classical application of semi-supervised learning \citep{angelopoulos2023prediction}.  The goal is to estimate the mean response $\theta_{\ast}:=\EE Y$, by leveraging the unlabelled observations and multiple experts $f_1,\cdots, f_K: \calX\mapsto \calY$ which provide ``reliable" prediction of the response.  Without loss of generality, we focus on the case of univariate response where $\calY\subset \RR$ and is compact.  For any $X$, denote $\bff(X)=(f_1(X),\cdots,f_K(X))^{\top}$, and we write $\bff$ for brevity. 

\begin{assumption}\label{assump:fK}
There exist constants $c_0, C_0>0$ such that $|Y|\leq C_0$ and $|f_k(X)|\leq C_0$ {\it almost surely} for all $k=1,\cdots,K$, and $\cov(\bff)\succcurlyeq c_0\cdot \bI_K$.
\end{assumption}

The uniformly boundedness condition in Assumption~\ref{assump:fK} is imposed for technical simplicity, which is relaxable. We assume that the $K$ experts are not overly correlated with each other, otherwise, the optimal mixture may not be unique and our MOE procedure becomes unstable.

The conventional estimator $\hat\theta^{\con}$ and the PPI-based \citep{angelopoulos2023prediction} estimator $\hat\theta^{\ppi}$, given an expert $f(\cdot)$, are defined by
\begin{align}
\hat\theta^{\con}: &= \frac{1}{n}\sum_{i=1}^nY_i, \label{eq:mean-con}\\
\hat\theta^{\ppi}_f :&= \frac{1}{N}\sum_{i=1}^Nf(\tilde X_i)-\frac{1}{n}\sum_{i=1}^n\left(f(X_i)-Y_i\right),\label{eq:mean-ppi}
\end{align}
respectively. These two estimators are both unbiased, and as shown in \cite{angelopoulos2023prediction}, their covariances are 
$$
\var(\hat\theta^{\con})\asymp \frac{1}{n}\cdot \var(Y)\quad {\rm and}\quad \var(\hat\theta^{\ppi})\asymp \frac{1}{n}\cdot \var\big(Y-f(X)\big)+\frac{1}{N}\cdot \var\big(f(X)\big).
$$
The PPI-based estimator achieves a smaller variance than the conventional one as long as the expert is accurate in that $\var\big(Y-f(X)\big)\ll \var(Y)$ and the unlabelled data is abundant in that $N\gg n$. 

Given that multiple experts $f_1(\cdot),\cdots, f_K(\cdot)$ are available, our MOE-powered approach aims to exploit a mixture of these experts to achieve a robust PPI-based estimator. More specifically, given $\beta\in\RR^K$, define the MOE-powered estimator by
\begin{align}\label{eq:mean-moe-general}
\hat\theta^{\moe}_{\beta} &= \frac{1}{N}\sum_{i=1}^N F_{\beta}(\tilde X_i)-\frac{1}{n}\sum_{i=1}^n\left(F_{\beta}(X_i)-Y_i\right),
\end{align}
where the linear mixture of experts $F_{\beta}(x):=\sum_{k}\beta_kf_k(x)$. For any fixed $\beta$, the MOE-powered estimator $\hat\theta^{\moe}_{\beta}$ is unbiased. Moreover, $\hat\theta_{\be_k}^{\moe}=\hat\theta_{f_k}^{\ppi}$ where $\be_k$ denotes the $k$-th canonical basis vector. To this end, the MOE-powered approach targets the linear mixture which achieves the smallest variance and define the \emph{oracle} MOE weight vector by
$
\beta_{\ast}:=\arg\min_{\beta}\ \var\big(F_{\beta}(X)-Y\big).
$
Its sample version based on labelled data is given by 
\begin{equation}\label{eq:mean-moe-beta}
\hat\beta_n:=\underset{\beta}{\arg\min}\ \svar\big(F_{\beta}(X)-Y\big):=\frac{1}{n}\sum_{i=1}^n \big(F_{\beta}(X_i)-Y_i\big)^2-\bigg(\frac{1}{n}\sum_{i=1}^n\big(F_{\beta}(X_i)-Y_i\big)\bigg)^2,
\end{equation}
which admits a closed-form solution. 

For each $X_i$ and $\tilde X_i$, define the $K$-dimensional predictor vectors by  $\bff_i=\big(f_1(X_i), \cdots,f_K(X_i)\big)^{\top}$ and $\tilde\bff_i=\big(f_1(\tilde X_i), \cdots,f_K(\tilde X_i)\big)^{\top}$, respectively.  We assume, without loss of generality, that $\EE (\bff_1\bff_1^{\top})$ is invertible. Collectively, we define the predictor matrices by $\bF=(\bff_1,\cdots,\bff_n)^{\top}\in\RR^{n\times K}$ and $\tilde\bF=(\tilde\bff_1,\cdots,\tilde\bff_N)^{\top}\in\RR^{N\times K}$. As a result, the sample MOE weight vector $\hat\beta_n$ from (\ref{eq:mean-moe-beta}) actually minimizes $(\bF\beta-\by)^{\top}\bP_n (\bF\beta-\by)$, where $\by=(Y_1,\cdots, Y_n)^{\top}$ and $\bP_n=\bI_n-n^{-1}{\bf 1}_n{\bf 1}_n^{\top}$. Here, ${\bf 1}_n$ denotes the $n$-dimensional all-one vector.  The solution to (\ref{eq:mean-moe-beta}) is $\hat\beta_n=(\bF^{\top}\bP_n\bF)^{-1}\bF^{\top}\bP_n\by$. Similarly, the oracle MOE weight vector is $\beta_{\ast}=\big(\cov(\bff)\big)^{-1}\cov(\bff, Y)$ and
\begin{equation}\label{eq:mean-moe-ppi-rel}
\hat\theta_{\beta_{\ast}}^{\moe}=\hat\theta^{\ppi}_{f_{\ast}},\qquad {\rm where}\ f_{\ast}(X)=\langle \beta_{\ast}, \bff\rangle=\sum_{k=1}^K \beta_{\ast, k} f_k(X).
\end{equation}
The function $f_{\ast}$ is the oracle mixture of experts.

Finally, our MOE-powered mean value estimator can be written as
\begin{align}
\hat\theta^{\moe}_{\hat\beta_n}=&N^{-1}{\bf 1}_N^{\top}\tilde \bF\hat\beta_n-n^{-1}{\bf 1}_n^{\top}\big(\bF\hat\beta_n-\by\big)\notag\\
=&\big(N^{-1}{\bf 1}_N^{\top}\tilde \bF-n^{-1}{\bf 1}_n^{\top}\bF\big)(\bF^{\top}\bP_n\bF)^{-1}\bF^{\top}\bP_n\by+n^{-1}{\bf 1}_n^{\top}\by.  \label{eq:mean-moe-beta-1}
\end{align} 
Hereafter, we shall write $\hat\theta^{\moe}$ for brevity whenever its dependence on $\hat\beta_n$ is clear from the context.  Unlike the conventional and PPI-based estimator,  the MOE-powered estimator can be biased.  The following lemma provides an upper bound of the bias of $\hat\theta^{\moe}$. 

\begin{lemma}\label{lem:mean-moe-bias}
Suppose that Assumption~\ref{assump:fK} holds and $n\geq C_1K^2\log n$ for a large absolute constant $C_1>0$. The MOE-powered estimator by (\ref{eq:mean-moe-beta-1}) is biased and 
$$
\Big|\EE\hat\theta^{\moe}-\theta_{\ast}  \Big|=O\bigg(\frac{K^2\log n}{n}\bigg)
$$
\end{lemma}

Lemma~\ref{lem:mean-moe-bias} shows that the bias of $\hat\theta^{\moe}_{\hat\beta_n}$ is negligible.  The MOE-powered inference procedure is detailed in Algorithm~\ref{algo:mean_estimation}. For any $\alpha\in(0,1)$,  we denote $z_{\alpha}:=\Phi^{-1}(1-\alpha)$ where $\Phi(\cdot)$ is the cumulative distribution function of a standard normal random variable. 

\begin{algorithm}[H]
\caption{MOE-powered  Inference on Mean Value}
\label{algo:mean_estimation}
\begin{algorithmic}[1]
    \State \textbf{Input}: labelled and unlabelled predictor matrices $\bF=(\bff_1,\cdots,\bff_n)^{\top}$ and $\tilde\bF=(\tilde\bff_1,\cdots,\tilde\bff_N)^{\top}$, the observed response vector $\by=(Y_1,\cdots, Y_n)^{\top}$.
    \State{Mean response and predictors: $\bar Y\gets n^{-1}{\bf 1}^{\top}_ n \by$ and $\bar \bff^{\top}\gets n^{-1}{\bf 1}_n^{\top}\bF$}.
    \State{Estimate MOE weight vector: $\hat\beta_n\gets (n^{-1}\bF^{\top}\bF-\bar \bff\bar \bff^{\top})^{-1}(n^{-1}\bF^{\top}\by-\bar \bff\bar Y)$}
    \State{MOE-powered point estimate: $\hat\theta^{\moe}\gets N^{-1}{\bf 1}_N^{\top}\tilde \bF\hat\beta_n-n^{-1}{\bf 1}_n^{\top}\big(\bF\hat\beta_n-\by\big)$}
    \State{Estimated the variance: $\hat \sigma_{Y-f_{\ast}}^2\gets n^{-1}\|\bF\hat\beta_n-\by\|^2-\big(n^{-1}{\bf 1}_n^{\top}(\bF\hat\beta_n-\by)\big)^2$}
    \State \textbf{Output}: $\hat\theta^{\moe}$ and its $100(1-\alpha)\%$ confidence interval $\mathcal C_\alpha:=\Big(\hat\theta^{\moe}\pm z_{\alpha/2}n^{-1/2}\hat\sigma_{Y-f_{\ast}}\Big)$
\end{algorithmic}
\end{algorithm}

\begin{theorem}[mean value inference]\label{thm:mean_estimation}
Suppose that Assumption~\ref{assump:fK} holds,  $n\gg K^4\log^2 n$, and $N\gg n\log n$.   The MOE-powered estimator $\hat\theta^{\moe}$ produced by Algorithm \ref{algo:mean_estimation} satisfies
\begin{equation}\label{equ:normality_mean}
 \sqrt{n}\big(\hat\theta^{\moe}-\theta_{\ast}\big)=-Z_{n,f_{\ast}}+\tilde O_p\bigg(\frac{K^2\log n}{\sqrt{n}}+\sqrt{\frac{n\log n}{N}}\bigg),
\end{equation}
where $n^{1/2}Z_{n,f_{\ast}}:=\sum_{i=1}^n\big(f_{\ast}(X_i)-Y_i-\EE\big(f(X)-Y\big)\big)$. Moreover, the coverage probability of $\calC_{\alpha}$ output by Algorithm~\ref{algo:mean_estimation} satisfies
\begin{equation}\label{equ:CI_mean}
    \PP\big(\theta_{\ast}\in\calC_\alpha\big)= 1-\alpha+O\bigg(\frac{K^2\log n}{\sqrt{n}}+\sqrt{\frac{n\log n}{N}}\bigg).
\end{equation}
\end{theorem}

\begin{remark}
	The confidence interval constructed by Algorithm~\ref{algo:mean_estimation} has a width characterized by $\sigma^2_{Y-f_{\ast}}=\var\big(Y-f_{\ast}(X)\big)$. It is determined by the accuracy of the optimal mixture of experts. Since $\var\big(Y-f_{\ast}(X)\big)\leq \min_k \var\big(Y-f_k(X)\big)$, our MOE-powered confidence interval is at least as short as the PPI-based confidence interval by any individual expert. This property holds for all the other algorithms in this section.
\end{remark}

The confidence interval constructed in Algorithm~\ref{algo:mean_estimation} ignores the variance contributed by the unlabelled data, which also introduces the remainder term $(n/N)\log n$ in (\ref{equ:normality_mean}) and (\ref{equ:CI_mean}). While this remainder term is negligible for large $N$, we can still eliminate it by slightly modifying Algorithm~\ref{algo:mean_estimation} and using a slightly wider confidence interval. 

\begin{theorem}[mean value inference+]\label{thm:mean-value+}
Suppose that Assumption~\ref{assump:fK} holds and  $n\gg K^4\log^2 n$. The MOE-powered estimator  $\hat\theta^{\moe}$ produced by Algorithm \ref{algo:mean_estimation} satisfies
$$
\sqrt{n}\big(\hat\theta^{\moe}-\theta_{\ast}\big)=-Z_{n,f_{\ast}}+\tilde Z_N+\tilde O_p\bigg(\frac{K^2\log n}{\sqrt{n}}\bigg),
$$
where $n^{1/2}Z_{n,f_{\ast}}:=\sum_{i=1}^n\big(f_{\ast}(X_i)-Y_i-\EE\big(f(X)-Y\big)\big)$ and $\tilde Z_N:=n^{1/2}N^{-1} \sum_{i=1}^N\big(f_{\ast}(\tilde X_i)-\EE f_{\ast}(X)\big)$.  Let $\hat\sigma_{f_{\ast}}^2:=N^{-1}\|\tilde\bF\hat\beta_n\|^2-(N^{-1}{\bf 1}_N^{\top}\tilde\bF\hat\beta_n)^2$ be an estimate of $\sigma_{f_{\ast}}^2=\var\big(f_{\ast}(X)\big)$ and construct 
$$
\tilde \calC_{\alpha}:=\Bigg(\hat\theta^{\moe}-z_{\alpha/2}\sqrt{\frac{\hat\sigma_{Y-f_{\ast}}^2}{n}+\frac{\hat\sigma_{f_{\ast}}^2}{N}},\ \hat\theta^{\moe}+z_{\alpha/2}\sqrt{\frac{\hat\sigma_{Y-f_{\ast}}^2}{n}+\frac{\hat\sigma_{f_{\ast}}^2}{N}}\Bigg). 
$$
Then,  $\PP\big(\theta_{\ast}\in\tilde \calC_{\alpha}\big)=1-\alpha+O\big(K^2n^{-1/2}\log n+N^{-1/2}\log n\big)$.
\end{theorem}

\subsection{Quantile inference}\label{sec:quantile}

Quantiles play a central role in statistical inference because they provide a complete description of the underlying distribution. For any fixed $q\in(0,1)$, we define the $q$-quantile of $Y$ as $\theta_\ast := \inf\big\{\theta\in\RR:\PP(Y\leq \theta)\geq q\big\}$, where we suppress the dependence of $\theta_{\ast}$ on $q$ for brevity. Similarly as in Section~\ref{sec:m-est}, we search for the candidate solutions from a fine-grained grid $\Theta_{\grid}\subset\Theta$. 

Under mild continuity assumption on the distribution function of $Y$, we have $\EE\big[\II\{Y\leq \theta_{\ast}\}\big]=q$. Motivated by this equation (see \cite{angelopoulos2023prediction} for more details), the conventional and PPI-based approaches construct the confidence sets by studying the following estimating functions: 
\begin{align*}
    \hat m^{\con}(\theta) &:= n^{-1}\sum_{i=1}^n\II\big\{Y_i\leq \theta\big\}-q,\\
    \hat m_f^{\ppi}(\theta) &:= n^{-1}\sum_{i=1}^n\big(\II\big\{Y_i\leq \theta\big\}-\II\big\{f(X_i)\leq \theta\big\}\big) +N^{-1}\sum_{i=1}^N\II\big\{f(\tilde X_i)\leq \theta\big\}-q,
\end{align*}
where $f(\cdot)$ is a given predictor. Both of them are unbiased in that $\EE \hat m^{\con}(\theta)=\EE \hat m_f^{\ppi}(\theta)=\EE\big[\II(Y\leq \theta)\big]-q$. Moreover, if $\theta=\theta_{\ast}$, by central limit theorem, these two estimating functions can be normally approximated by
\begin{equation*}
    \calN\bigg(0,\frac{\var\big(\II\big\{Y\leq \theta_\ast\big\}\big)}{n}\bigg)\quad\text{and}\quad \calN\bigg(0,\frac{\var\big(\II\big\{Y\leq \theta_\ast\big\}-\II\big\{f(X)\leq \theta_\ast\big\}\big)}{n}+\frac{\var\big(\II\big\{f(\tilde{X})\leq \theta_\ast\big\}\big)}{N}\bigg),
\end{equation*}
respectively. If $N\gg n$ and $f(\cdot)$ is a powerful predictor so that $\var\big(\II\{Y\leq \theta\}-\II\{f(X)\leq \theta\}\big)\leq \var\big(\II\{Y\leq \theta\}\big)$, then $\var\big(\hat m_f^{\ppi}(\theta)\big)\leq \var\big(\hat m^{\con}(\theta)\big)$ and the PPI-based approach may yield a tighter confidence set. 

Note that $\var\big(\II\big\{Y\leq \theta_\ast\big\}\big)=q-q^2$ and 
\begin{align*}
    &\var\big(\II\big\{Y\leq \theta_\ast\big\}-\II\big\{f(X)\leq \theta_\ast\big\}\big)=\PP\big(\big(f(X)-\theta_\ast\big)\big(Y-\theta_{\ast}\big)<0 \big)-\big(q-\PP(f(X)\leq \theta_{\ast})\big)^2.
\end{align*}
By the shape of the function $t-t^2$ for $t\in[0,1]$, a necessary condition for $\var\big(\II\{Y\leq \theta_{\ast}\}-\II\{f(X)\leq \theta_{\ast}\}\big)\leq \var\big(\II\{Y\leq \theta_{\ast}\}\big)$ is that $\big| q-\PP(f(X)\leq \theta_{\ast})\big|\leq \min(q, 1-q)$.

However, due to the non-smoothness of indicator function, the general M-estimation framework in Section~\ref{sec:m-est} is not immediately applicable. There are several common approaches to smooth the indicator function, such as linearization at the jump point \citep{koltchinskii2016perturbation}. Here, we adopt the sigmoid function $S_h(t)=(1+\exp\{-t/h\})^{-1}, \forall t\in\RR$ for its higher-order smoothness, where $h>0$ is a bandwidth parameter.  Note that $\big| S_h'(t)\big|\leq h^{-1}$. The bandwidth controls the tradeoff between bias and variance. 

Based on the smoothed indicator function, we define the conventional, PPI-based, and MOE-powered estimating functions by 
\begin{align}
\hat m^{\con}(\theta) &:= \frac{1}{n}\sum_{i=1}^nS_{h}\big(\theta-Y_i\big)-q,\notag\\
    \hat m^{\ppi}_f(\theta) &:= \frac{1}{n}\sum_{i=1}^n\big[S_{h}\big(\theta-Y_i\big)-S_{h}\big(\theta-f(X_i)\big)\big] +\frac{1}{N}\sum_{i=1}^NS_h\big(\theta-f(\tilde X_i)\big)-q,\notag\\
    \hat m^{\moe}_\beta(\theta) &:= \frac{1}{n}\sum_{i=1}^n\big[S_h\big(\theta-Y_i\big)-S_h\big(\theta-\bff_i^\top \beta\big)\big] +\frac{1}{N}\sum_{i=1}^NS_h\big(\theta-\tilde{\bff}_i^\top\beta\big)-q\label{eq:quantile-moe-beta-1},
\end{align}
where we suppress the dependence of $h_n$ on $n$ for notational brevity and $\bff_i, \tilde\bff_i$ are as defined in Section~\ref{sec:moe-mean}. Hereafter, we take
\begin{equation}\label{equ:h_order}
    h:=n^{-\gamma},\quad \gamma\in\Big(\frac14,\ \frac12\Big).
\end{equation}
The oracle MOE and its sample counterpart are defined as the minimizers of the population and sample variances, respectively. Denote $Q(\theta, \beta): = \var\big(S_h(\theta-Y)-S_h(\theta-\bff^\top \beta)\big)$ and $Q_n(\theta, \beta): = \svar\big(S_h(\theta-Y)-S_h(\theta-\bff^\top \beta)\big)$. More specifically, we have
\begin{align*}
    Q_n(\theta,\beta): &= \frac{1}{n}\sum_{i=1}^n \Big(S_h(\theta-Y_i)-S_h(\theta-\bff_i^\top \beta)\Big)^2 - \bigg(\frac{1}{n}\sum_{i=1}^n \Big(S_h(\theta-Y_i)-S_h(\theta-\bff_i^\top \beta)\Big)\bigg)^2.
\end{align*}
Finally, the oracle MOE weight vector and its sample counterpart are defined by 
\begin{equation}\label{eq:quantile-beta-def}
  \beta_\ast(\theta):= \underset{\beta\in\calB}{\arg\min}\ Q(\theta,\beta) \quad \textrm{and} \quad \hat\beta_n(\theta):= \underset{\beta\in\calB}{\arg\min}\, Q_n(\theta,\beta),
\end{equation}
where $\calB\subset \RR^K$ is a compact set. Hereafter, we will suppress the dependence of $\beta_{\ast}$ and $\hat\beta_n$ on $\theta$ for simplicity.


\begin{assumption}\label{assump:quantile}
There exists a constant $U>0$ such that $\|\bff(X)\|\leq U$ almost surely.  Moreover, let $\beta_{\ast}=\beta_{\ast}(\theta_{\ast})$ and 
\begin{itemize}
    \item[(A1)] there exist $\delta>0$ and $C_0>0$ such that 
    $
        \sup_{\beta\in\calB_{\ast}(\delta)} \big|p_\beta(\cdot)\big|\le C_0,
    $
    where $\calB_{\ast}(\delta):=\{\beta\in\calB:  \|\beta-\beta_{\ast}\|\leq \delta\}$ and $p_{\beta}$ is the probability density function of $F_{\beta}(X)$;
    \item[(A2)] there exist $\delta_0>0$, $c_0>0$ and $\eta_0>0$ such that
    \begin{align*}
        \inf_{\beta\in\calB_{\ast}(\delta_0) } \lambda_{\min}\!\bigl(h\nabla^2 Q(\theta_\ast,\beta)\bigr) \ge c_0 \qquad {\rm and}\qquad 
        \inf_{\beta\notin\calB_{\ast}(\delta_0) }\big\{Q(\theta_\ast,\beta) - Q(\theta_\ast,\beta_\ast)\big\}\geq\eta_0.
    \end{align*}

    {\item[(A3)] there exists a neighborhood $\calN_0$ of $\theta_\ast$ such that the probability density function $f_Y$ is differentiable on $\calN_0$, and $f_Y'$ is continuous at $\theta_\ast$.}

\end{itemize}
\end{assumption}

Note that $\beta_{\ast}(\theta)$ changes with respect to $\theta$, but the conditions ({\it A1}) and ({\it A2}) in Assumption~\ref{assump:quantile} are imposed only on $\beta_{\ast}(\theta_{\ast})$. The condition ({\it A1}) characterizes the smoothness of $Q(\theta_{\ast}, \beta)$ with respect to $\beta\in\calB_{\ast}(\delta)$. It implies that $\big|\EE\big(S_h(\theta_{\ast}-\bff^{\top}\beta_1)-\EE\big(S_h(\theta_{\ast}-\bff^{\top}\beta_2)\big)\big) \big|=O(\|\beta_1-\beta_2\|)$ for all $\beta_1, \beta_2\in\calB_{\ast}(\delta)$. 
The condition ({\it A2}) imposes locally strong convexity and global separation condition around $\beta_\ast$, which ensure the uniqueness of $\beta_\ast$. This condition is weaker than the global strongly convexity in Assumption~\ref{assump:m-est}. 
And the condition ({\it A3}) ensures the feasibility of smoothing approach.


\begin{lemma}\label{lem:quantile-est-beta}
    Suppose Assumption~\ref{assump:quantile} holds and $n\geq C_1(K+\log n)$ for a large constant $C_1>0$. Then, there exists a constant $C_2>0$ and an event $\calE$ with $\PP(\calE)\geq 1-n^{-10}$ on which
    \begin{equation}
        \big\|\hat\beta_n - \beta_\ast\big\| \leq C_2\sqrt{\frac{K+\log n}{n}},
    \end{equation}
    where $\hat\beta_n=\hat\beta_n(\theta_{\ast})$ and $\beta_\ast=\beta_{\ast}(\theta_{\ast})$ are as defined in (\ref{eq:quantile-beta-def}).
\end{lemma}


For any fixed $\beta$, the MOE-powered estimating function $\hat m_{\beta}^{\moe}(\theta_{\ast})$ is biased due to the kernel smoothing. 
Indeed, $\EE \hat m_{\beta}^{\moe}(\theta_{\ast})=\EE S_h(\theta_{\ast}-Y)-q=O(h^2)$, provided that $Y$ has a smooth density in a neighborhood of $\theta_{\ast}$. The following lemma shows that a similar bias remains even if the sample MOE weight $\hat\beta_n$ is employed.

\begin{lemma}\label{lem:quantile-moe-bias}
Suppose that Assumption~\ref{assump:quantile} holds and $n\geq C_1(K+\log n)$ for a large constant $C_1>0$. The MOE-powered estimating function by (\ref{eq:quantile-moe-beta-1}) satisfies 
$$
\Big|\EE\hat m^\moe_{\hat\beta_n}(\theta_\ast)\Big|=O\bigg(h^2+\frac{K+\log n}{nh}\bigg).
$$
\end{lemma}

Since the  variance of $\hat m^\moe_{\hat\beta_n}(\theta_\ast)$ is at the order of $O(n^{-1})$, Lemma~\ref{lem:quantile-moe-bias} implies that a valid inference of $\theta_{\ast}$ requires $1/4< \gamma< 1/2$ if $h\asymp n^{-\gamma}$. 

The MOE-powered inference procedure is detailed in Algorithm~\ref{algo:quantile_estimation}. The coverage probability of the MOE-powered confidence set is guaranteed by Theorem~\ref{thm:quantile_estimation}.

\begin{algorithm}[H]
\caption{MOE-powered Inference on Quantile}
\label{algo:quantile_estimation}
\begin{algorithmic}[1]
    \State \textbf{Input}: labelled and unlabelled predictor vectors $\bff_i,1\leq i\leq n$ and $\tilde\bff_i,1\leq i\leq N$, the observed responses $Y_i,1\leq i\leq n$, quantile level $q\in(0,1)$, bandwidth $h$, sigmoid function $S_h(t)=1/(1+e^{-t/h})$, initial set   $\calC^{\moe}_{\alpha}=\emptyset$;
    \For{$\theta\in\Theta_{\text{grid}}$}
        \State{solve $\hat \beta_n=\arg\min_{\beta} Q_n(\theta, \beta)$;}
        \State{imputed gradient: $\tilde g_{\theta, \hat\beta_n}\gets N^{-1}\sum_{i=1}^N S_h(\theta-\tilde \bff_i^\top \hat \beta_n)$;}
        \State{rectifier: $\hat \Delta_{\theta,\hat\beta_n} \gets n^{-1}\sum_{i=1}^n \big(S_h({\theta-Y_i})-S_h(\theta-\bff_i^\top \hat \beta_n)\big)$;}
        
        \State{sample variance: $\hat Q_n(\theta,\hat\beta_n) \gets n^{-1}\sum_{i=1}^n \big(S_h({\theta-Y_i})-S_h(\theta-\bff_i^\top \hat \beta_n)\big)^2-\Big(\hat\Delta_{\theta,\hat\beta_n}\Big)^2$;}

        \If{$\big|\tilde g_{\theta, \hat\beta_n}+\hat \Delta_{\theta,\hat\beta_n}-q \big|\leq z_{\alpha/2}\sqrt{\hat Q_n(\theta,\hat\beta_n)/n}$}
            
            \State{update: $\calC_{\alpha}^{\moe}\gets \calC_{\alpha}^{\moe}\cup \{\theta\}$;}
        \EndIf
    \EndFor
    \State \textbf{Output:} confidence set $\calC_{\alpha}^{\moe}$.
\end{algorithmic}
\end{algorithm}

\begin{theorem}[quantile inference]\label{thm:quantile_estimation}
Suppose that Assumption~\ref{assump:quantile} and the conditions of Lemma~\ref{lem:quantile-moe-bias} hold.   
The MOE-powered estimating function satisfies
\begin{equation}\label{equ:normality_quantile}
\sqrt{n}\,\hat m^\moe_{\hat\beta_n}(\theta_\ast) =Z_{n,f_{\ast}}+\tilde O_p\bigg(\sqrt{n}h^2+\frac{K+\log n}{\sqrt {n}h}+\sqrt\frac{K+\log n}{{N}h}+\sqrt{\frac{n\log n}{N}}\bigg),
\end{equation}
where 
\begin{align*}
    n^{1/2}Z_{n,f_{\ast}}:=\frac{1}{\sqrt{n}}\sum_{i=1}^n\Big(S_h({\theta_\ast-Y_i})-S_h({\theta_\ast-\bff_i^\top\beta_\ast})- \EE\big[S_h({\theta_\ast-Y_i})-S({\theta_\ast-\bff_i^\top\beta_\ast})\big]\Big).
\end{align*}
Moreover, the coverage probability of $\calC_{\alpha}^\moe$ output by Algorithm~\ref{algo:quantile_estimation} satisfies
\begin{equation}\label{equ:CI_quantile}
    \PP\big(\theta_{\ast}\in\calC_\alpha^\moe\big)= 1-\alpha+O\bigg(\sqrt{n}h^2+\frac{K+\log n}{\sqrt {n}h}+\sqrt {\frac{K+\log n}{Nh}}+\sqrt{\frac{n\log n}{N}}\bigg).
\end{equation}
\end{theorem}
\begin{remark}
    The optimality choice of the bandwidth in (\ref{equ:h_order}) is $h_n = n^{-1/3}$, which implies 
    \begin{align*}
         \sqrt{n}\,\hat m^\moe_{\hat\beta_n}(\theta_\ast) &= Z_{n,f_{\ast}}+\tilde O_p\bigg(n^{-1/6}(K+\log n)+\sqrt{\frac{n\log n}{N}}\bigg),\\
         \PP\big(\theta_{\ast}\in\calC_\alpha^\moe\big) &= 1-\alpha+O\bigg(n^{-1/6}(K+\log n)+\sqrt{\frac{n\log n}{N}}\bigg).
    \end{align*}
\end{remark}

The confidence interval constructed in Algorithm \ref{algo:quantile_estimation} ignores the variance contributed by the unlabelled data, which also introduces the remainder term $(n/N) \log n$ in (\ref{equ:normality_quantile}) and (\ref{equ:CI_quantile}). While this remainder term is negligible for large $N$, we can still eliminate it by slightly modifying Algorithm \ref{algo:quantile_estimation} and using a slightly wider confidence interval.

\begin{theorem}[quantile inference+]\label{thm:quantile_estimation+}
Suppose that Assumption~\ref{assump:quantile} and conditions of Lemma~\ref{lem:quantile-moe-bias} hold. 
The MOE-powered estimating function produced by Algorithm~\ref{algo:quantile_estimation} satisfies
\begin{equation}\label{equ:normality_quantile+}
 \sqrt{n}\,\hat m^\moe_{\hat\beta_n}(\theta_\ast) = Z_{n,f_{\ast}}+\tilde Z_{N, f_\ast}+\tilde O_p\bigg(\sqrt{n}h^2+\frac{K+\log n}{\sqrt nh}\bigg),
\end{equation}
where 
\begin{align*}
    n^{1/2}Z_{n,f_{\ast}}&:=\sum_{i=1}^n\Big(S_h({\theta_\ast-Y_i})-S_h({\theta_\ast-\bff_i^\top\beta_\ast})- \EE\big[S_h({\theta_\ast-Y})-S({\theta_\ast-\bff^\top\beta_\ast})\big]\Big).\\
    n^{1/2}\tilde Z_{N,f_\ast}&:=n^{1/2}N^{-1}\sum_{i=1}^N\Big(S_h({\theta_\ast-\tilde \bff_i^\top\beta_\ast})- \EE S_h({\theta_\ast-\bff^\top\beta_\ast})\Big).
\end{align*}
Let $\tilde Q_n(\theta, \hat\beta_n):=N^{-1}\sum_{i=1}^NS_h(\theta-\tilde\bff_i^\top\hat\beta_n)^2-\big[N^{-1}\sum_{i=1}^NS_h({\theta-\tilde\bff_i^\top\hat\beta_n})\big]^2$ be an estimate of $\tilde Q(\theta, \beta_\ast)=\var\big(S_h({\theta-\bff^\top \beta_\ast})\big)$, then give the refined confidence set
\begin{align*}
    \tilde \calC_\alpha^\moe := \bigg\{\theta\in\Theta_\grid:\Big|\hat 
    m_{\hat\beta_n}^\moe(\theta)\Big| \leq z_{\alpha/2}\sqrt{\frac{Q_n(\theta, \hat\beta_n)}{n}+\frac{\tilde Q_n(\theta, \hat\beta_n)}{N}} \bigg\}.
\end{align*}
Then the coverage probability holds that
\begin{align*}
    \PP\big(\theta_\ast\in\tilde\calC_\alpha^\moe\big)=1-\alpha+O\Bigg(\sqrt{n}h^2+\frac{K+\log n}{\sqrt nh}+\frac{1}{\sqrt N}\Bigg).
\end{align*}
\end{theorem}

\subsection{Linear regression and its inference}\label{sec:moe-mlr}
Given the pair of covariates and response $(X, Y)\in \calX\times \calY\subset \RR^d\times \RR$,  the  best linear model and its coefficients are defined by 
\begin{equation}\label{eq:mlr-theta_star}
    \btheta_{\ast}= \underset{\btheta}{\arg\min}\ \EE(Y-X^{\top}\btheta)^2=(\EE XX^{\top})^{-1}\EE(XY). 
\end{equation}
For simplicity, we assume $\theta_{\ast}$ is unique (see Assumption~\ref{assump:mlr}). 
Motivated similarly as in Section~\ref{sec:moe-mean},  we consider the conventional,  PPI,  and our MOE-powered estimators defined as 
\begin{equation}\label{eq:mlr-ppi-moe}\begin{split}
\hat\theta^{\con} &:= \big(\bX^{\top}\bX\big)^{-1}\bX^{\top} \by\\
\hat\theta^{\ppi}_{f} &:= \big(\tilde \bX^{\top}\tilde \bX\big)^{-1}\tilde \bX^{\top}f(\tilde \bX)-\big(\bX^{\top}\bX\big)^{-1}\bX^{\top}(f(\bX)-\by)\\
\hat\theta^{\moe}_{\hat\beta_n} &:= \big(\tilde \bX^{\top}\tilde \bX\big)^{-1}\tilde \bX^{\top}\tilde \bF\hat \beta_n-\big(\bX^{\top}\bX\big)^{-1}\bX^{\top}(\bF^{\top}\hat \beta_n-\by),\\
\end{split}\end{equation}
respectively,  where the  feature matrices $\bX=(X_1, \cdots, X_n)^{\top}\in\RR^{n\times d}$ and $\tilde \bX =\big( \tilde X_1, \cdots, \tilde X_N)^{\top}\in\RR^{N\times d}$ are constructed using labelled and unlabelled data,  respectively.  The matrices $\bF\in\RR^{n\times K}$ and $\tilde\bF\in\RR^{N\times K}$ are constructed by applying experts to the feature matrices,  similarly as in Section~\ref{sec:moe-mean}.   Here,  $f(\bX)\in\RR^n$ denotes the operation of applying the expert $f(\cdot)$ to the rows of $\bX$.  The MOE-powered estimator is reliant on $\hat\beta_n$, i.e.,  the estimation of the oracle mixture of experts., which we shall unfold shortly. 

The asymptotic properties of $\hat\theta^{\con}$ and $\hat\theta^{\ppi}_f$ are well-documented in the literature.  It is clear that both estimators are biased,  yet consistent and asymptotically normal.  It was shown  by \cite{white1980nonlinear} that 
\begin{equation*}
    \sqrt{n}(\hat\theta^{\con}-\theta_{\ast})\stackrel{d.}{\longrightarrow}  \calN\big(0, \bW_{Y-X^{\top}\theta_{\ast}}\big),\qquad {\rm as}\ n\to\infty,
\end{equation*}
where $\bW_{Y-X^{\top}\theta_{\ast}}:=\bSigma^{-1}\EE\big[(Y-X^{\top} \theta_{\ast})^2XX^{\top}\big]\bSigma^{-1}$ and $\bSigma:=\EE XX^{\top}$.  Note that we used the simple fact $\EE (Y-X^{\top}\theta_{\ast})X=0$ by the definition in (\ref{eq:mlr-theta_star}).  Similarly,  \cite{angelopoulos2023prediction} shows that 
\begin{equation*}
    \sqrt{n}(\hat\theta^{\ppi}_{f}-\theta_{\ast})\stackrel{d.}{\longrightarrow} \calN\big(0, \bW_{Y-f}\big),\qquad {\rm as}\ n, N\to\infty\ {\rm and}\ n/N\to 0,  
\end{equation*}
where $\bW_{Y-f}:= \bSigma^{-1}\EE\big[\big(Y-f(X)-X^{\top}\bdelta_f\big)^2XX^{\top}\big]\bSigma^{-1}$ with the rectifier defined by $\bdelta_{f}:=\bSigma^{-1}\EE\big[X\big(Y-f(X)\big)\big]$.  
Note that $\bW_{Y-f}\neq \bSigma^{-1}\cov\big(X(Y-f(X))\big)\bSigma^{-1}$ in general.  The rectifier serves as a debiasing treatment in the sense that $\EE\big(X(Y-f(X)-\bdelta_f^{\top}X)\big)=0$.  The PPI-based estimator enjoys a smaller variance than the conventional one when the expert $f(\cdot)$ is accurate,  e.g.,  if $(Y-f(X)-X^{\top}\bdelta_f)^2\ll (Y-X^{\top}\theta_{\ast})^2$ almost surely.  
Note that a finite-sample estimate of $\cov(\hat\theta_f^{\ppi})$ is given by 
\begin{equation}\label{eq:hat-cov-ppi}
\hat\bW_{Y-f}:=n(\bX^{\top}\bX)^{-1}\cdot \bigg(\sum_{i=1}^n \big(Y_i-f(X_i)-X_i^{\top}\hat\bdelta_{f}\big)^2X_iX_i^{\top}\bigg)\cdot (\bX^{\top}\bX)^{-1},
\end{equation}
where $\hat \bdelta_{f}:=(\bX^{\top}\bX)^{-1}\cdot \sum_{i=1}^n X_i(Y_i-f(X_i))=(\bX^{\top}\bX)^{-1}\bX^{\top}\big(\by-f(\bX)\big)$.  

Our MOE-powered estimator (\ref{eq:mlr-ppi-moe}) intends to minimize the (co)variance of $\hat\theta^{\moe}_{\hat\beta_n}$ through a carefully chosen $\hat\beta_n$.   As discussed in Section~\ref{sec:moe-mean},  given a fixed mixture assignment $\beta$,  the MOE-powered estimator can be written as a PPI-based estimator via $\hat\theta^{\moe}_{\beta}=\hat\theta^{\ppi}_{F_\beta}$,  where $F_{\beta}:=\sum_{k=1}^K \beta_k f_k$.  Moreover,  $\cov(\hat\theta^{\moe}_{\beta})$ is given by 
\begin{equation}\label{eq:mlr-moe-pop-cov}
    \bW_{Y-F_{\beta}}=\bSigma^{-1}\cdot \EE\big[\big(Y-F_{\beta}(X)-X^{\top}\bdelta_{F_{\beta}}\big)^2XX^{\top}\big]\cdot \bSigma^{-1}.
\end{equation}
Since $\hat\theta_{\beta}^{\moe}$ is multivariate, its variability can be viewed from several perspectives. For instance, for a fixed index $s\in[d]$,  one may consider the entrywise variance $\var\big((\hat\theta^{\moe}_{\beta})_s\big)=\be_s^{\top}\bW_{Y-F_{\beta}}\be_s$.  Here, $\be_s$ denotes the $s$-th canonical basis vector, whose dimension may vary at different appearances.  Without loss of generality and for ease of exposition, we focus primarily on the total variance of $\hat\theta^{\moe}$, defined as  $\tr\big(\bW_{Y-F_{\beta}}\big)$. 

Define the oracle MOE weight vector by  $\beta_{\ast}:=\arg\min_{\beta} \tr(\bW_{Y-F_{\beta}})$, which aims to minimize the total variance of our MOE-powered estimator. Accordingly, define the oracle MOE by $f_{\ast}(X)=F_{\beta_{\ast}}(X)$. 
 Its sample version is simply defined by 
\begin{align}
\hat\beta_n:=&\underset{\beta}{\arg\min}\ \tr\big(\hat\bW_{Y-F_{\beta}}\big),\label{eq:MOE-hat-beta}
\end{align}
where $\hat\bW_{Y-F_{\beta}}=n(\bX^{\top}\bX)^{-1}\cdot \big(\bX^{\top}\diag^2(\by-F_{\beta}(\bX)-\bX\hat\bdelta_{F_{\beta}})\bX\big)\cdot (\bX^{\top}\bX)^{-1}$ is as defined in (\ref{eq:hat-cov-ppi}).  Here,  $\diag(\bv)$ represents a diagonal matrix constructed from the vector $\bv$.  Note that $F_{\beta}(\bX)=\bF\beta$ with the matrix $\bF\in\RR^{n\times K}$ being defined as in Section~\ref{sec:moe-mean}.    Hereafter, we shall write $\hat\theta^{\moe}$ for brevity whenever its dependence on $\hat\beta_n$ is clear from the context.  

Define $\bSigma:=\EE XX^{\top}$ and 
$$
\bH:=\EE\big[(X^{\top}\bSigma^{-2}X)\bv(X)\bv^{\top}(X)\big]\quad {\rm where}\ \bv^{\top}(X)=X^{\top}\bSigma^{-1}\EE(X\bff^{\top})-\bff^{\top},
$$
with $\bff:=\bff(X)=\big(f_1(X),\cdots,  f_K(X)\big)^{\top}$.  

\begin{assumption}\label{assump:mlr}
There exist constants $c_0,  C_0>0$ such that $|Y|\leq C_0,  \|X\|\leq C_0,  $ and $|f_k(X)|\leq C_0$ {\it almost surely} for all $k=1,\cdots,K$,  $\bSigma\succcurlyeq c_0\cdot \bI_d$,  and $\bH\succcurlyeq c_0\cdot \bI_K$.
\end{assumption}
The matrix $\bSigma$ is invertible,  so $\theta_{\ast}$ is uniquely defined,  whereas $\bH$ is invertible to ensure the uniqueness of $\beta_{\ast}$. 

\begin{lemma}\label{lem:mlr-moe-bias}
Suppose that Assumption~\ref{assump:mlr} holds and $n\geq C_1K^2d\log n$ for a large absolute constant $C_1>0$.  The bias of the MOE-powered estimator by (\ref{eq:mlr-ppi-moe}) is bounded by 
$$
\Big\|\EE\hat\theta^{\moe}-\theta_{\ast}  \Big\|=O\bigg(\frac{(K^2\sqrt{d}+d)\log n}{n}\bigg).
$$
\end{lemma}

Lemma~\ref{lem:mlr-moe-bias} shows that the bias of MOE-powered estimator is negligible if sample size is large.  The inference procedure is detailed in Algorithm~\ref{algo:linear_regression}.

\begin{algorithm}[H]
\caption{MOE-powered Inference on Linear Regression}  
\label{algo:linear_regression}
\begin{algorithmic}[1]
    \State{ {\bf Input}: labelled and unlabelled predictor matrices $\bF=(\bff_1,\cdots,\bff_n)^{\top}$ and $\tilde\bF=(\tilde\bff_1,\cdots,\tilde\bff_N)^{\top}$,  labelled and unlabelled covariate matrices $\bX=(X_1,\cdots, X_n)^{\top}$ and $\tilde\bX=(\tilde X_1,\cdots, \tilde X_N)^{\top}$,  the observed response vector $\by=(Y_1,\cdots, Y_n)^{\top}$, an index $s\in[d]$. }
    \State{$\hat\bSigma\gets n^{-1}\bX^{\top}\bX$,  $\hat \bA \gets (\bX^{\top}\bX)^{-1}\bX^{\top}\bF$,  $\hat \bb \gets  (\bX^{\top}\bX)^{-1}\bX^{\top}\by$;}
        \State{$\hat \bH\gets n^{-1}\sum_{i=1}^n (X_i^{\top} \hat\bSigma^{-2}X_i) (\bff_i-\hat \bA^{\top}X_i)(\bff_i-\hat \bA^{\top}X_i)^{\top}$;\\
        $\hat \br\gets n^{-1}\sum_{i=1}^n(X_i^{\top}\hat\bSigma^{-2}X_i)(\bff_i-\hat \bA^{\top}X_i)(Y_i-X_i^{\top}\hat \bb)$;}
    \If{$\lambda_{\min}(\hat\bH)\leq 0.01c_0$}
        \State{$\hat\beta_n\gets0$;}
    \Else
        \State{estimate MOE weight vector: $\hat\beta_n \gets \hat \bH^{-1}\hat \br$;}
    \EndIf
    \State{MOE point estimate $\hat\btheta^{\moe}\gets \big(\tilde \bX^{\top}\tilde \bX\big)^{-1}\tilde \bX^{\top}\tilde \bF\hat \beta_n-\big(\bX^{\top}\bX\big)^{-1}\bX^{\top}(\bF^{\top}\hat \beta_n-\by)$;}
    \State{covariance $\hat \bW=n^{-1}\hat\bSigma^{-1}\cdot \big(\bX^{\top}\diag^2(\by-\bF\hat\beta_n-\bX\hat\bdelta_{F_{\hat\beta_n}})\bX\big)\cdot \hat\bSigma^{-1}$ with $\hat\bdelta_{F_{\hat\beta_n}}=(\bX^{\top}\bX)^{-1}\bX^{\top}(\by-\bF\hat\beta_n)$.}
    \State \textbf{Return} MOE point estimate $\hat\theta^{\moe}$ and entrywise confidence interval $\calC_{\alpha,s}=\Big(\hat\theta^{\moe}_{s}\pm z_{\alpha/2}\sqrt{\be_s^{\top}\hat\bW \be_s/n}\Big)$.
\end{algorithmic}

\end{algorithm}
 
The non-asymptotic performance of the MOE-powered estimator and the coverage guarantee for its confidence intervals are given in Theorem~\ref{thm:mlr-MOE}. For simplicity, we focus on entrywise inference for fixed entries. Extending the results to simultaneous inference for all entries is possible, but it would substantially complicate the presentation of our method; therefore, we omit it.

\begin{theorem}[linear regression]\label{thm:mlr-MOE}
  Suppose Assumption~\ref{assump:mlr} holds,  $n\geq C_0 dK^3\log^2n$,  and $N\geq C_0dn\log n$ for a large constant $C_0>0$.  Let $\hat\btheta^{\moe}$ be the MOE-powered estimator output by Algorithm~\ref{algo:linear_regression}.  Then,  for any fixed index $s\in[d]$,  
\begin{equation}\label{equ:normality_linear}
    \sqrt{n}\big(\hat\btheta^{\moe}_s-\theta_{\ast, s}\big)=Z_{n,f_{\ast},s}+\tilde O_p\bigg(\sqrt{\frac{dn\log n}{N}}+\frac{\sqrt{dK^3
    }\log n}{\sqrt{n}}\bigg),
\end{equation}
where $Z_{n,f_{\ast}, s}=\be_s^{\top}\bSigma^{-1}\cdot n^{-1/2}\sum_{i=1}^n\big(Y_i-f_{\ast}(X_i)-X_i^{\top}\bdelta_{f_{\ast}}\big)X_i$.   Moreover,  the coverage probability of $\calC_{\alpha, s}$ output by Algorithm~\ref{algo:linear_regression} satisfies 
\begin{equation}\label{equ:CI_linear}
    \PP\big(\theta_{\ast,s}\in\calC_{\alpha,s}\big)= 1-\alpha+O\bigg(\sqrt{\frac{dn\log n}{N}}+\frac{\sqrt{dK^3}\log n}{\sqrt{n}}\bigg).
\end{equation}
\end{theorem}

\subsection{Logistic regression and its inference}

Given the pair of covariates and response $(X,Y)\in \calX\times \calY \subset \RR^d\times \{0,1\}$ for binary classification, and target of inference for logistic regression is defined by 
$$
\theta_{\ast}=\underset{\theta\in\Theta}{\arg\min}\ \EE \big[-Y\theta^{\top}X+\log\big(1+\exp(\theta^{\top}X)\big)\big],
$$
where $\Theta\subset\RR^d$. The population estimating function becomes $\bmm(\theta):=\EE [\big(Y-S(\theta^{\top}X)\big)X]$ with the sigmoid function $S(t)=(1+e^{-t})^{-1}$. Therefore, $\theta_{\ast}$ solves the equation $\bmm(\theta)=\boldsymbol{0}$. Let $\Theta_{\grid}$ be a fine-grained grid of $\Theta$. 

Let $\bX, \tilde \bX,\by, \bff,\tilde\bff, \bF$, and $\tilde \bF$ be constructed similarly as in Section~\ref{sec:moe-mlr}. 
The conventional, PPI-based, and MOE-powered estimation functions are defined by 
\begin{equation}\label{eq:logistic-ppi-moe}\begin{split}
\hat \bmm^{\con}(\theta) &:= n^{-1}\bX^\top \big(\by-S(\bX\theta)\big),\\
\hat \bmm^{\ppi}_f(\theta) &:= N^{-1}\tilde \bX^\top \big(f(\tilde\bX)-S(\tilde \bX\theta)\big) - n^{-1}\bX^\top \big(f(\bX)-\by\big),\\
\hat \bmm^{\moe}_{\beta}(\theta) &:= N^{-1}\tilde \bX^\top \big(\tilde\bF \beta-S(\tilde \bX \theta)\big) - n^{-1}\bX^\top \big(\bF\beta-\by\big),
\end{split}\end{equation}
respectively, where $S(\bX\theta) = (S(X_1^\top\theta), \cdots, S(X_n^\top\theta))^{\top}\in\RR^{n}$.  

Note that both $\hat\bmm^{\con}$ and $\hat\bmm_f^{\ppi}$ are both unbiased estimators of the population estimating function. Denote $\bW_{Y-X^{\top}\theta}:=\cov\big(X(Y-S(X^{\top}\theta))\big)$ and $\bW_{Y-f}:=\cov\big(X(Y-f(X))\big)$. Then, under mild conditions, CLT dictates that
\begin{align*}
    n^{1/2}\big(\hat \bmm^{\con}(\theta)-\bmm(\theta)\big)&\overset{d.}{\to} \calN\big(0, \bW_{Y-X^{\top}\theta}\big), \textrm{ as } n\to\infty; \\
    n^{1/2}\big(\hat \bmm_f^\ppi(\theta)-\bmm(\theta)\big) &\overset{d.}{\to} \calN\big(0, \bW_{Y-f}\big), \textrm{ if } n/N\to 0 \textrm{ as } n\to\infty.
\end{align*}
For any given $\theta$, the oracle MOE weight vector is defined as the minimizer of the total variance of $\hat \bmm_{\beta}^{\moe}(\theta)$. Its empirical counterpart is defined in a similar fashion. More exactly, let
\begin{align}\label{eq:logistic-moe-beta}
    \beta_\ast &:=\underset{\beta}{\arg\min}\, \tr\big(\bW_{Y-F_{\beta}}\big)\quad \text{and} \quad \hat\beta_n :=\underset{\beta}{\arg\min}\, \tr\big(\hat\bW_{Y-F_{\beta}}\big),
\end{align}
where the $\beta$-index mixture of experts $F_{\beta}(X)=\bff^{\top}\beta=\sum_k \beta_k f_k(X)$. The sample covariance matrix $\hat\bW_{Y-F_{\beta}}=\scov\big(X(\bff^\top \beta-Y))\big)$ with an explicit form as
$$
n\cdot\hat\bW_{Y-F_{\beta}}=\bX^{\top}\diag(\by-\bF\beta)\bP_n\diag(\by-\bF\beta) \bX,
$$
where $\bP_n=\bI_n-n^{-1}{\bf 1}_n{\bf 1}_{n}^{\top}$.  Note that these covariance matrices $\bW_{Y-f}$ and $\hat\bW_{Y-f}$ are irrelevant to $\theta$. 

\begin{assumption}\label{assump:logistic} 
    There exist constant $c_0, C_0$ such that $\|X\|\leq C_0$, $|Y|\leq C_0$ and $|f_k(X)|\leq C_0$ for all $k\in [K]$ almost surely, and $\EE\big[\big(\bff X^\top-\EE\bff X^\top\big)\big(X\bff^\top-\EE X\bff^\top\big)\big]\succeq c_0\bI_d$. 
\end{assumption}
The last condition of Assumption~\ref{assump:logistic} guarantees that the minimizer $\beta_{\ast}$ is unique. See the proof of Lemma~\ref{lem:logistic-moe-bias} for the explicit form of $\beta_{\ast}$. 
The following lemma provides the high-probability upper bound for $\hat\beta_n-\beta_{\ast}$. 
\begin{lemma}\label{lem:logistic-moe-bias}
Suppose that Assumption~\ref{assump:logistic} holds, $n\geq C_1(d+K)\log n$, and $N\geq C_1 n$ for a large constant $C_1>0$. Then,
\begin{align*}
    \|\hat\beta_n-\beta_{\ast}\|=\tilde O_p\bigg(\frac{{(d+K)\log n}}{n}\bigg) \quad {\rm and}\quad    \Big\|\EE\hat \bmm_{\hat\beta_n}^\moe(\theta)-\bmm(\theta))\Big\| = O\Big(\frac{{dK\log n}}{n}\Big),
\end{align*}
for any fixed $\theta\in\Theta$. 
\end{lemma}

The inference procedure is detailed in Algorithm~\ref{algo:logistic_regression}.  Note that the estimated MOE weight vector $\hat\beta_n$ is irrelevant of $\theta$, whereas the imputed gradient must be recalculated for each $\theta\in\Theta_{\grid}$. Algorithm~\ref{algo:logistic_regression} outputs only the MOE-powered confidence set. If a point estimate of $\theta_{\ast}$ is desired, we suggest $\hat\theta^{\moe}:=\arg\min_{\theta\in\Theta_{\grid}} \|\tilde \bg_{\theta, \hat\beta_n}-\hat\bDelta_{\hat\beta_n}\|$, but its theoretical property is skipped. 

\begin{algorithm}
\caption{MOE-powered Inference for Logistic Regression}  
\label{algo:logistic_regression}
\begin{algorithmic}[1]
    \State{ {\bf Input}: labelled and unlabelled predictor matrices $\bF=(\bff_1,\cdots,\bff_n)^{\top}$ and $\tilde\bF=(\tilde\bff_1,\cdots,\tilde\bff_N)^{\top}$,  labelled and unlabelled covariate matrices $\bX=(X_1,\cdots, X_n)^{\top}$ and $\tilde\bX=(\tilde X_1,\cdots, \tilde X_N)^{\top}$,  the observed response vector $\by=(Y_1,\cdots, Y_n)^{\top}$,
    initial Set $\calC_\alpha^\moe=\emptyset$; }

    \State{$\hat \bH\gets n^{-1}\sum_{i=1}^n \bff_iX_i^\top X_i\bff_i^\top - \big(n^{-1}\sum_{i=1}^n\bff_iX_i^\top\big) \big(n^{-1}\sum_{i=1}^nX_i\bff_i^\top\big)$;}
    \State{$\hat \br\gets n^{-1}\sum_{i=1}^n \bff_iX_i^\top X_iY_i - \big(n^{-1}\sum_{i=1}^n\bff_iX_i^\top\big) \big(n^{-1}\sum_{i=1}^nX_iY_i\big)$;}
    \If{$\lambda_{\min}(\hat\bH)\leq 0.01c_0$}
        \State{$\hat\beta_n\gets 0$;}
    \Else
        \State{estimate MOE weight vector: $\hat\beta_n \gets \hat \bH^{-1}\hat \br$;}
    \EndIf
    \State{rectifier: $\hat \bDelta_{\hat\beta_n} \gets n^{-1}\sum_{i=1}^n \big(\bff_i^\top\hat \beta_n-Y_i\big)X_i$;}
    \State{sample covariance: $\hat \bW_{Y-F_{\hat\beta_n}} \gets n^{-1}\sum_{i=1}^n (\bff_i^{\top}\hat \beta_n-Y_i)^2X_iX_i^\top-\hat\bDelta_{\hat\beta_n} \hat\bDelta_{\hat\beta_n} ^\top$;}
    \For{$\theta\in\Theta_{\text{grid}}$}
        \State{imputed gradient: $\tilde \bg_{\theta,\hat\beta_n}\gets N^{-1}\sum_{i=1}^N\big(\tilde \bff_i^\top\hat \beta_n-S(\tilde X_i^{\top}\theta)\big)\tilde X_i$;}
        \If{$\big|\be_s^\top(\tilde \bg_{\theta,\hat\beta_n}-\hat\bDelta_{\hat\beta_n})\big|\leq z_{\alpha/(2d)}\sqrt{\be_s^{\top}\hat\bW_{Y-F_{\hat\beta_n}}\be_s/n},\forall s\in[d]$}
            \State{$\calC_\alpha^\moe\gets \calC_\alpha^\moe \cup \{\theta\}$;}
        \EndIf
    \EndFor
    \State \textbf{Return} confidence set $\mathcal C_{\alpha}^\moe$.
\end{algorithmic}
\end{algorithm}

\begin{theorem}[Logistic regression]\label{thm:logistic_estimation}
Suppose that Assumption~\ref{assump:logistic} holds, $n\geq C_1(d+K)^2\log^2 n$, and $N\geq C_1(d+K)n\log n$ for a large constant $C_1>0$. Then, for any fixed $\theta\in\Theta$, 
\begin{equation}\label{equ:normality_logistic}
 \sqrt{n}\, \big(\hat \bmm^{\moe}_{\hat\beta_n}(\theta)-\bmm(\theta)\big) = -\bZ_{n,f_{\ast}}+\tilde O_p\bigg(\frac{(d+K)\log n}{\sqrt{n}}+\sqrt{\frac{(d+K)n\log n}{N}}\bigg),
\end{equation}
where $n^{1/2}\bZ_{n,f_{\ast}}:=\sum_{i=1}^n\Big[X_i\big(F_{\beta_{\ast}}(X_i)-Y_i\big)-\EE X\big(F_{\beta_\ast}(X)-Y\big)\Big]$. Moreover, the coverage probability of $\calC_{\alpha}^\moe$ output by Algorithm~\ref{algo:logistic_regression} satisfies
\begin{align}\label{equ:CI_logistic}
    \PP\Big(\theta_\ast \in\calC_\alpha^\moe\Big) \geq 1-\alpha+O\bigg({\frac{{d(d+K)\log n}}{\sqrt n}}+\sqrt{\frac{{d^2(d+K)n\log n}}{N}}\bigg),
\end{align}
given $\theta_\ast\in\Theta_{\grid}$.
\end{theorem}

The confidence interval constructed in Algorithm~\ref{algo:logistic_regression} ignores the variance contributed by the unlabelled data, which also introduces the remainder term $(n/N)\log n$ in (\ref{equ:normality_logistic}) and (\ref{equ:CI_logistic}). While this remainder term is negligible for large $N$, we can still eliminate it by slightly modifying Algorithm~\ref{algo:logistic_regression} and using a slightly wider confidence interval. Denote 
$$
\tilde\bZ_{N,f_{\ast}}:=N^{-1/2}\sum_{i=1}^N\Big[ \tilde X_i\Big(f_{\ast}(\tilde X_i)-S\big(\tilde X_i^\top \theta\big)\Big)-\EE X\big(f_{\ast}(X)-S(X^\top\theta)\big)\Big],
$$
where $f_{\ast}(\cdot)=F_{\beta_\ast}(\cdot)$. It is clear that $\tilde\bZ_{N,f_{\ast}}$ is asymptotically normal with covariance matrix $\bW_{f_{\ast}-X^{\top}\theta}=\cov\big(X\big(f_{\ast}(X)-S(X^{\top}\theta)\big)\big)$. Its empirical version $\tilde\bW_{f_{\ast}-X^{\top}\theta}=\scov\big(\tilde X\big(f_{\ast}(\tilde X)-S(\tilde X^{\top}\theta)\big)\big)$ is constructed solely from the unlabelled data.

\begin{theorem}[Logistic regression+]\label{thm:logistic+}
Suppose that Assumption~\ref{assump:logistic} holds, $n\geq C_1(d+K)^2\log^2 n$, and $N\geq C_1n\log^2 n$ for a large constant $C_1>0$. For any $\theta$, 
the estimating function $\hat \bmm^{\moe}_{\hat\beta_n}(\theta)$ produced by Algorithm \ref{algo:logistic_regression} satisfies
$$
    \sqrt n\,  \Big(\hat \bmm_{\hat\beta_n}^\moe(\theta)-\bmm(\theta))\Big) =\, -\bZ_{n,f_\ast} + \sqrt{\frac nN}\tilde \bZ_{N, f_\ast}+\tilde O_p\Big({\frac{{(d+K)\log n}}{\sqrt n}}\Big).
$$
Construct the confidence set by
$$
\tilde \calC_{\alpha}^{\moe}:=\Bigg\{\theta\in\Theta_{\grid}:\bigg|\frac{\sqrt n\,  \be_s^{\top}\hat \bmm_{\hat\beta_n}^\moe(\theta)}{\sqrt{\be_s^{\top}\hat\bW_{Y-F_{\hat\beta_n}}\be_s/n+\be_s^{\top}\tilde\bW_{f_{\ast}-X^{\top}\theta}\be_s/N}}\bigg|\leq z_{\alpha/(2d)},\ \forall s\in[d]\Bigg\}. 
$$
If $\theta_{\ast}\in\Theta_{\grid}$, then
$$
\PP\big(\theta_{\ast}\in\tilde \calC_{\alpha}^{\moe}\big)\geq 1-\alpha+O\bigg(\frac{d(d+K)\log n}{\sqrt{n}}+\frac{d(d+K)\log n}{\sqrt{N}}\bigg).
$$
\end{theorem}

\section{Numerical Experiments}
\label{sec:experiments}

This section empirically evaluates the performance of the MOE-powered inference framework through a comprehensive set of experiments. These experiments aim to demonstrate that: (i) MOE outperforms or at least matches PPI equipped with the best single predictor; (ii) MOE is robust under model misspecification; (iii) MOE-powered confidence sets achieve the desired coverage probabilities; and (iv) MOE requires the fewest labeled samples among all methods to achieve the same statistical efficiency.

\subsection{Simulation studies}

\subsubsection{Data modeling}

We consider two settings in which the labeled data are generated from a linear model and a nonlinear model, respectively. More specifically,
\begin{itemize}
    \item[(1)] \textit{Linear model.} The labeled data \((X, Y)\) satisfy
    $
    Y = 10 + X^{\top}\beta + \varepsilon,
    $
    where $\varepsilon \sim \mathcal{N}(0, \sigma^2)$. The covariate $X \in \mathbb{R}^{20}$ has i.i.d.\ entries drawn from the standard normal distribution, and $\beta \in \mathbb{R}^{20}$ has 10 nonzero entries in the first 10 coordinates, each sampled from $U(0,100)$. Here, we set $\sigma = 10$. The labeled data set $\calL$ is drawn from fixed $50{,}000$ samples. In addition, we generate another $50{,}000$ samples to train the experts (predictors), which will be specified in Section~\ref{sec:num-pred}.
    
    \item[(2)] \textit{Nonlinear model.} The response variable $Y$ contains a nonlinear component of $X$:
    $
    Y = X^{\top}\beta + \gamma g(X) + \varepsilon,
    $
    where
    $
    g(X) = 10 \sin(\pi X_{11}X_{12}) + 20 (X_{13}-0.5)^2,
    $
    and $X$ and $\beta$ are generated in the same way as above. The nonlinear function $g(X)$ includes sinusoidal, polynomial, and interaction terms and corresponds to the nonlinear component of the Friedman1 model \citep{friedman1991multivariate}. The parameter $\gamma$ controls the strength of the nonlinearity.
\end{itemize}

\subsubsection{Candidate expert predictors}\label{sec:num-pred}

We train six classical machine learning models to serve as the expert predictors $\{f_1, \ldots, f_6\}$:
(1) \textit{Deep neural network (DNN):} 3-layer fully connected network (MLP) with ReLU activations, hidden dimensions $[64, 32, 16]$, and trained with Adam optimizer for $100$ epochs;
    (2) \textit{Linear model:} the least squares estimator;
    (3) \textit{Quadratic regression:} Degree-2 polynomial regression;
    (4) \textit{Random forest:} 100 trees with max depth 10, minimum samples split 5;
    (5) \textit{XGBoost:} Gradient boosting with 100 estimators, learning rate 0.1, max depth 6;
    (6) \textit{LightGBM:} Light gradient boosting with 100 estimators, learning rate 0.1, max depth 6. 

All experts are trained independently on an independent auxiliary labeled training sample with size $50{,}000$. 
Table \ref{tab:model_perf} summarises the validation-set RMSE for each trained expert. 
\begin{table}[ht]
\centering\small
\caption{Validation RMSE of the $K=6$ trained expert predictors}
\label{tab:model_perf}
\begin{tabular}{lcc}
\toprule
Experts & Linear & Nonlinear \\
\midrule
Linear / Polynomial & 10.01 / 10.03 & 26.21 / 16.73 \\
DNN                 & 11.26         & 18.37 \\
Random Forest       & 54.08         & 67.33 \\
XGBoost             & 30.45         & 35.87 \\
LightGBM            & 23.96         & 30.55 \\
\bottomrule
\end{tabular}
\end{table}
It highlights strikingly different performances across predictors, especially under nonlinear models. All the subsequent simulation studies focus on linear mixture of experts $\mathcal F=\left\{F_{\beta}(x)=\sum_{k=1}^K \beta_k f_k(x):\beta\in\RR^K\right\}$.
The empirical weight vector $\hat\beta$ is learned by the variance-minimization principle detailed in previous sections. 

For each inferential task, we compare five methods: (1) {\it Conventional}: classical estimator based on labeled data; (2) {\it PPI-worst}: PPI with the single worst-performing expert; (3): {\it PPI-mean}: PPI with the average of $K$ experts; (4) {\it PPI-best}: PPI with the single best-performing expert; (5): {\it PPI-MOE}: our proposed MOE-powered inference method. 
Note that ``best" and ``worst" are realized in hindsight, consequently infeasible in practice, evaluated by the width of confidence interval.
The results of point estimates are aggregated over 100 Monte Carlo runs. 


\subsubsection{Inference tasks}

We study four inference tasks as summarized in Table~\ref{tab:task_and_metric}, together with their population target (estimating function) and their empirical counterparts. For each task, we evaluate the variance of empirical inference target or estimating function by $B=1{,}000$ bootstrap replicates.

\begin{table}[ht]
\centering\scriptsize
\caption{Inference tasks, population targets, and their sample estimates.}
\label{tab:task_and_metric}
\begin{tabular}{p{0.15\textwidth}p{0.24\textwidth}p{0.55\textwidth}}
\toprule
Inference Task & Population Target & Sample Estimate (MOE-powered) \\
\midrule
Mean value & $\theta_{\ast}=\mathbb{E}[Y]$ & $N^{-1}{\bf 1}_N^{\top}\tilde \bF\hat\beta_n-n^{-1}{\bf 1}_n^{\top}\big(\bF\hat\beta_n-\by\big)$ \\
\midrule
Quantiles &
$\theta_{\ast}:\inf\left\{\theta\in\mathbb{R}:\mathbb{P}(Y\leq\theta)\geq q\right\}$ & $\frac{1}{n}\sum_{i=1}^n\big[S_h\big(\theta-Y_i\big)-S_h\big(\theta-\bff_i^\top \hat\beta_n\big)\big] +\frac{1}{N}\sum_{i=1}^NS_h\big(\theta-\tilde{\bff}_i^\top\hat\beta_n\big)-q$ \\
\midrule
Linear regression & $\theta_{\ast}:=\left[\mathbb{E}(X^{\top}X)\right]^{-1}\mathbb{E}(XY)$ & $\big(\tilde \bX^{\top}\tilde \bX\big)^{-1}\tilde \bX^{\top}\tilde \bF\hat \beta_n-\big(\bX^{\top}\bX\big)^{-1}\bX^{\top}(\bF^{\top}\hat \beta_n-\by)$\\
\midrule
Logistic regression  &
$\theta_{\ast}:\mathbb{E}\!\left[\left(S(X^{\top}\theta)-Y\right)X\right]=0$ & $N^{-1}\tilde \bX^\top \big(\tilde\bF \hat\beta_n-S(\tilde \bX \theta)\big) - n^{-1}\bX^\top \big(\bF\hat\beta_n-\by\big)$
\\
\bottomrule
\end{tabular}
\end{table}

\subsubsection{Variance reduction}

\begin{figure}[ht]
\centering
\includegraphics[width=\textwidth]{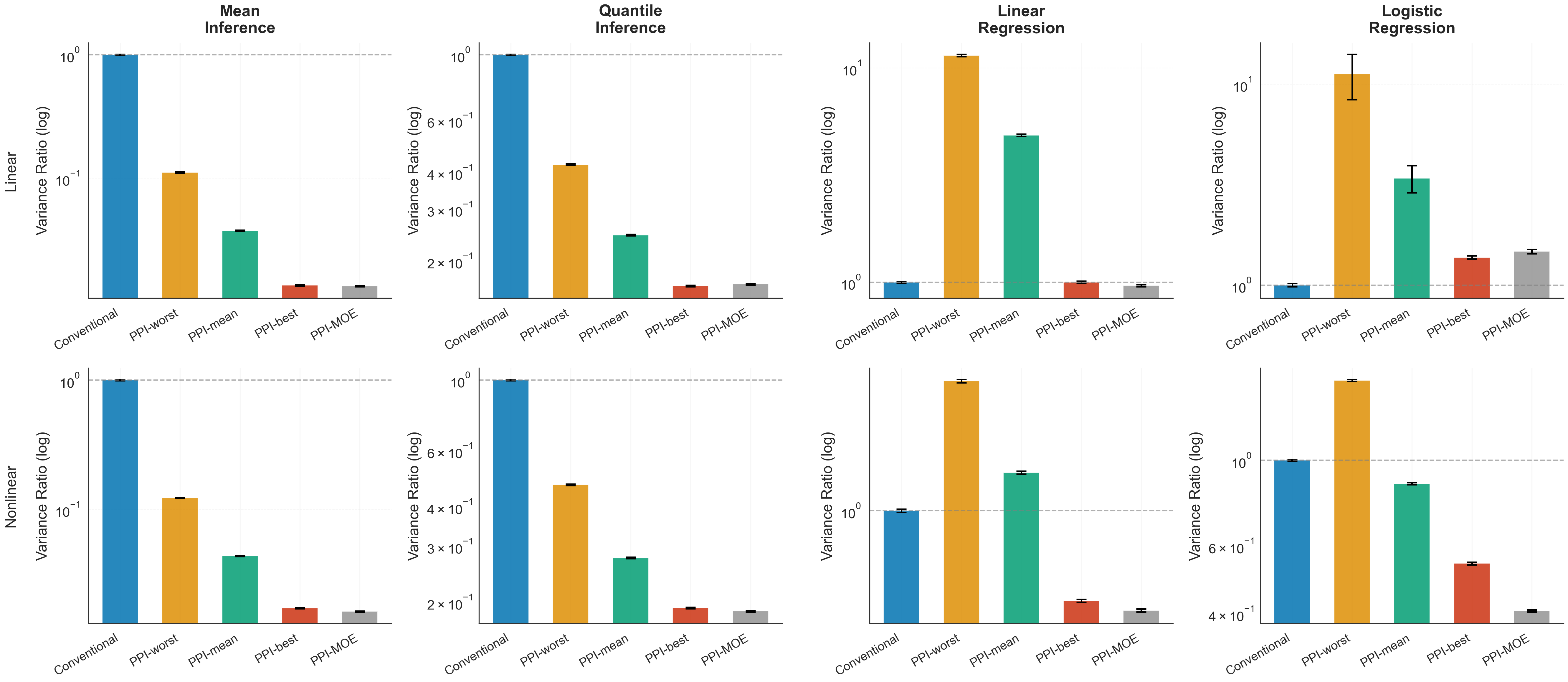}
\vspace{-4pt}
\caption{Variance reduction under Linear and Nonlinear regimes across inference tasks.}
\label{fig:variance_reduction_comp}
\end{figure}

Figure~\ref{fig:variance_reduction_comp} summarizes variance ratios across tasks and model regimes under fixed $n=200$ and $N=20{,}000$. Lower values indicate stronger efficiency gains over the conventional estimator. And black error bars denote the bootstrap standard deviation of the variance ratio estimates.

Across mean and quantile inference, PPI methods consistently reduce variance, and PPI-MOE closely tracks---and occasionally exceeds---PPI-best. 
Under the linear (well-specified) setting, PPI-best always chooses Linear Model as its predictor since it is the oracle model. While in nonlinear (misspecified) setting, when Linear Model is no longer the oracle one, PPI-MOE can choose a better ensembled predictor and consequently exceeds PPI-best.

For coefficient inference, performance is more sensitive to predictor quality: in linear settings weak single predictors can inflate variance, whereas under misspecification predictor diversity allows PPI-MOE and PPI-best to recover substantial gains. 

Overall, PPI-MOE is the most stable method across tasks because it adaptively combines complementary experts instead of relying on a single model.

\begin{figure}[H]
\centering
\includegraphics[width=0.6\textwidth]{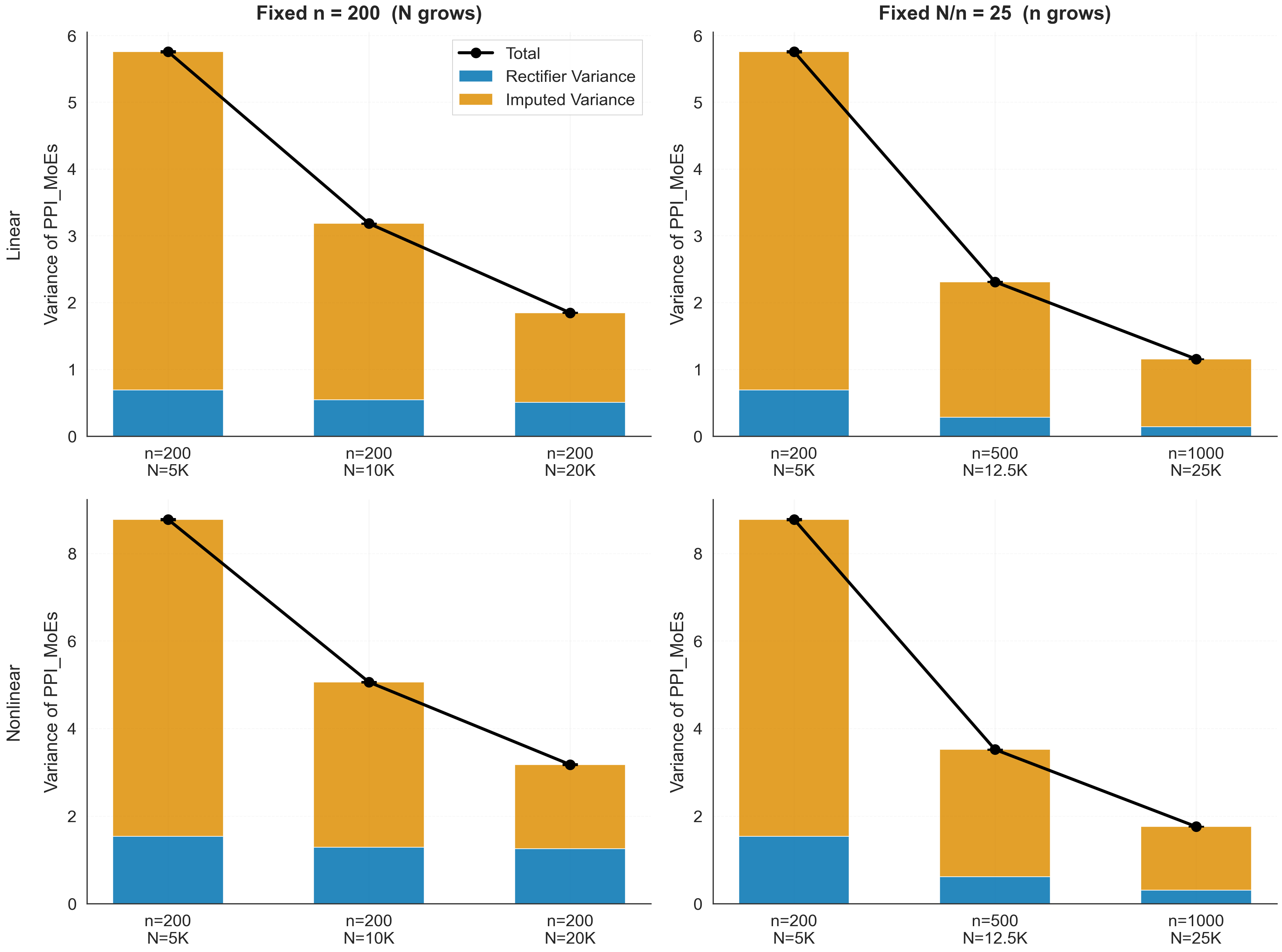}
\vspace{-4pt}
\caption{Effect of the growth rates of $n$ and $N$ on mean-inference variance.}
\label{fig:variance_reduction_n_N}
\end{figure}

Figure~\ref{fig:variance_reduction_n_N} decomposes the PPI-MOE variance into a rectifier variance term $\mathrm{Var}(Y-\bff^\top\hat\beta)/n$ and an imputed variance term $\mathrm{Var}(\bff^\top\hat\beta)/N$. As $N$ increases with fixed $n$, total variance decreases quickly at first and then plateaus at the labeled-data floor. When $n$ and $N$ grow proportionally, both components decrease, confirming that PPI benefits from larger unlabeled samples but remains fundamentally constrained by labeled-sample information.

\subsubsection{Coverage and interval width}

\begin{figure}[H]
\centering
\includegraphics[width=0.8\textwidth]{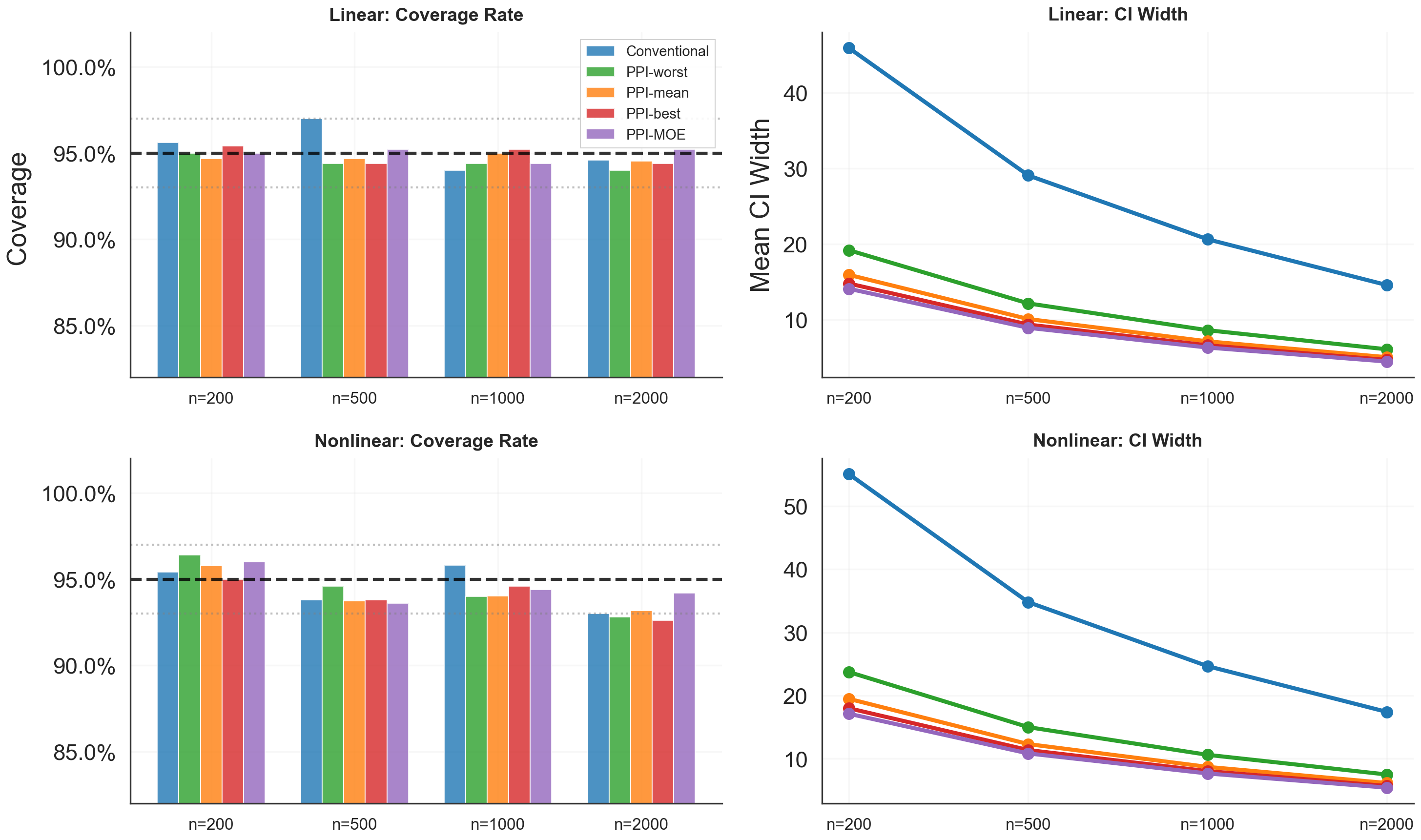}
\vspace{-0.4pt}
\caption{Coverage and confidence-interval width under linear and nonlinear settings.}
\label{fig:coverage_comparison}
\end{figure}

For coverage probability guarantee, we sample constant number of labeled samples and repeat the procedure of mean inference for $500$ times, and focus on whether the true value of mean falls in constructed confidence interval. Comparison of coverage rate and interval width will be shown in Figure~\ref{fig:coverage_comparison} to the certainty will be improved when $n$ increases and $N/n$ is fixed to $10$. 

PPI-MOE delivers the narrowest confidence intervals in both regimes, with strong gains in precision relative to Conventional and robust performance against weak predictors. Coverage is near nominal in linear settings and remains competitive under misspecification, with a mild under-coverage trade-off at moderate $n$ due to adaptive in-sample weight estimation.

{\begin{table*}[t]
\centering
\caption{Width comparison and coverage of PPI-MOE at \(n=500\).}
\label{tab:moes_summary_n500}
\resizebox{\textwidth}{!}{%
\begin{tabular}{lcccc|cc}
\toprule
Task-setting & MOE/Best & MOE/Mean & MOE/Worst & MOE/Conv & Coverage & Code \\
\midrule
Mean (L)           & 0.98 & 0.82 & 0.56 & 0.20 & 0.940 & ** \\
Mean (N)          & 0.98 & 0.81 & 0.56 & 0.21 & 0.940 & ** \\\hline
Quantile (L)       & 1.01 & 0.79 & 0.55 & 0.32 & 0.950 & *** \\
Quantile (N)      & 0.98 & 0.78 & 0.56 & 0.35 & 0.950 & *** \\\hline
Linear Reg. (L)    & 0.99 & 0.51 & 0.30 & 0.99 & 0.930 & . \\
Linear Reg. (N)   & 0.97 & 0.53 & 0.31 & 0.64 & 0.950 & *** \\\hline
Logistic Reg. (L)  & 0.50 & $<0.01$ & $<0.01$ & 1.72 & 0.970 & .\\
Logistic Reg. (N) & 0.85 & 0.68 & 0.49 & 0.64 & 0.960 & ** \\
\bottomrule
\end{tabular}}
\vspace{0.5em}

\parbox{0.97\textwidth}{\footnotesize
\textit{Notes.}
L/N in Task-setting column refers to data generation mode Linear/Nonlinear. Ratio columns report \(\mathrm{Width(PPI\mbox{-}MOE)}/\mathrm{Width(comparator)}\), with values below 1 indicating shorter intervals for PPI-MOE; values below \(0.01\) are shown as \(<0.01\).
The last column gives a coverage-agreement code, where more stars mean closer agreement of empirical coverage \(\hat p\) with the nominal target \(0.95\).
With $z={|\hat p-0.95|}/{\sqrt{0.95(1-0.95)/R}},$
we use \(***\) for \(z\le1.00\), \(**\) for \(1.00<z\le1.64\), \(*\) for \(1.64<z\le1.96\), \(. \) for \(1.96<z\le2.56\), and leave the entry blank otherwise.
}
\end{table*}

More extensive experiments and comparisons are summarized in Table~\ref{tab:moes_summary_n500}, which show that PPI-MOE delivers a robust bias--variance trade-off across all inference tasks. 
\begin{itemize}
    \item For mean and quantile estimation, it achieves near-nominal coverage while yielding confidence intervals that are essentially as short as those of PPI-best and markedly shorter than those of the conventional estimator, PPI-mean, and PPI-worst. 
    \item For linear regression, PPI-MOE remains close to PPI-best in both settings. Its gain over the conventional estimator is limited in the well-specified linear setting (MOE/Conv \(=0.99\)), where the conventional procedure is already near variance-optimal and leaves little room for further improvement; under nonlinear misspecification, however, PPI-MOE substantially shortens the interval width relative to the conventional estimator (MOE/Conv \(=0.64\)). 
    \item A similar contrast appears in logistic regression. In the linear setting, PPI-MOE may offer little gain relative to the conventional estimator, and can even be slightly worse (MOE/Conv \(=1.72\)), likely because the Bernoulli response introduces additional finite-sample variability that offsets the variance reduction from auxiliary predictions. 
    Under nonlinear misspecification, by contrast, PPI-MOE again yields a clear improvement over the conventional estimator (MOE/Conv \(=0.64\)) while remaining far more stable than PPI-mean and PPI-worst. Overall, these results show that PPI-MOE stays close to the best-performing baseline while avoiding the severe instability of more aggressive alternatives, and that its gains are especially pronounced under misspecification.
\end{itemize}

Overall, these results suggest that PPI-MOE preserves the efficiency gains of the strongest PPI baseline while being markedly more robust than simple averaging- or worst-case aggregation strategies. Full results are avaliable in Section~\ref{sec:result_coverage}.}

\subsubsection{Sample size efficiency}

A central practical question is how many labeled samples are needed to reach a target testing power when unlabeled data are abundant. We consider a mean-inference hypothesis test, $H_0:\mu=0$ versus $H_1:\mu\neq0$, with true mean $\mu=10$. For each method, we increase the labeled sample size $n$ (while fixing $N$ for PPI methods) until the empirical rejection probability (power) reaches a target level, such as 80\%. A method is considered more label-efficient if it requires fewer labeled samples to attain this target power.

\begin{figure}[ht]
\centering
\includegraphics[width=\textwidth]{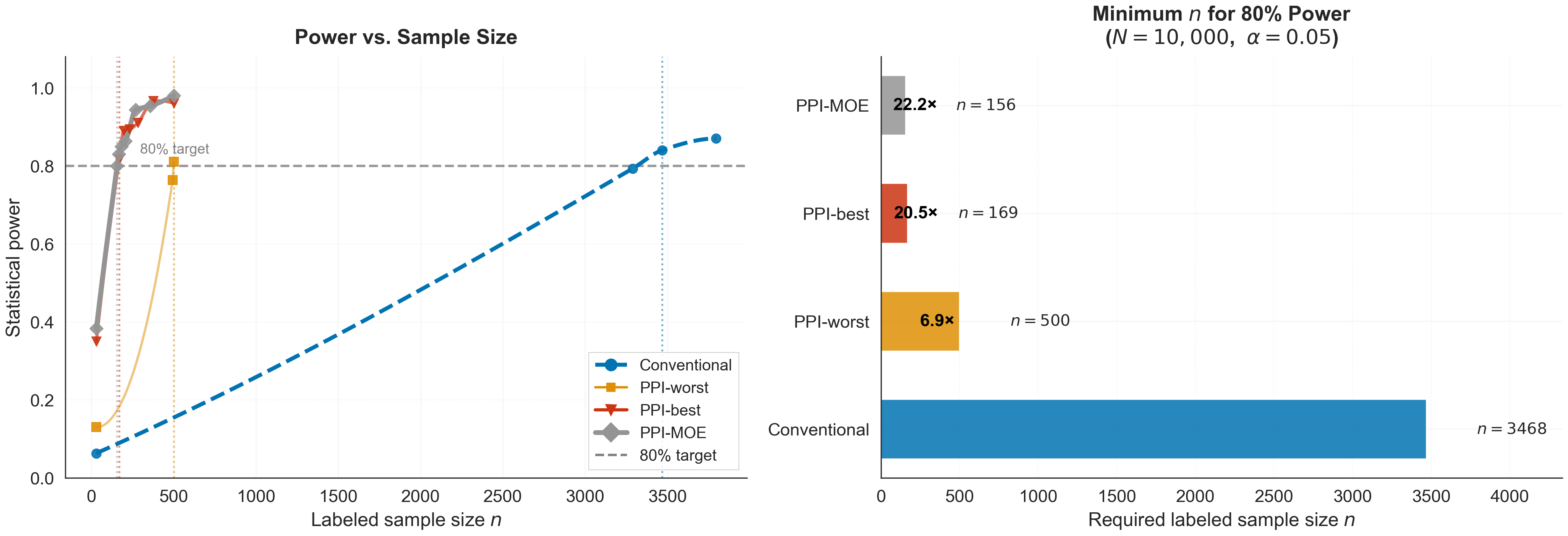}
\caption{Sample-size efficiency for mean inference under the linear setting.}
\label{fig:sample_efficiency}
\end{figure}

Figure~\ref{fig:sample_efficiency} shows that all PPI variants improve power efficiency over Conventional, with the strongest gains from PPI-MOE and PPI-best. The leading term still scales as $n^{-1}$, but its coefficient is substantially reduced through rectification (from $\mathrm{Var}(Y)$ toward $\mathrm{Var}(Y-\hat Y)$), while the additional $N^{-1}$ imputation term is negligible when $N\gg n$. In our experiments, PPI-MOE reaches target power with the fewest labeled samples and in some configurations slightly outperforms PPI-best.

\subsection{Real data experiments}

\subsubsection{Setup}

We evaluate the methods on California Housing and Bike Sharing, using the same predictor pool and the same method definitions as in simulation. We report variance-based metrics for mean, median, and linear-coefficient inference, and then compare sample-size efficiency via power analysis under fixed unlabeled budgets.

\subsubsection{Variance and power results}

\begin{figure}[ht]
\centering
\includegraphics[width=0.8\linewidth]{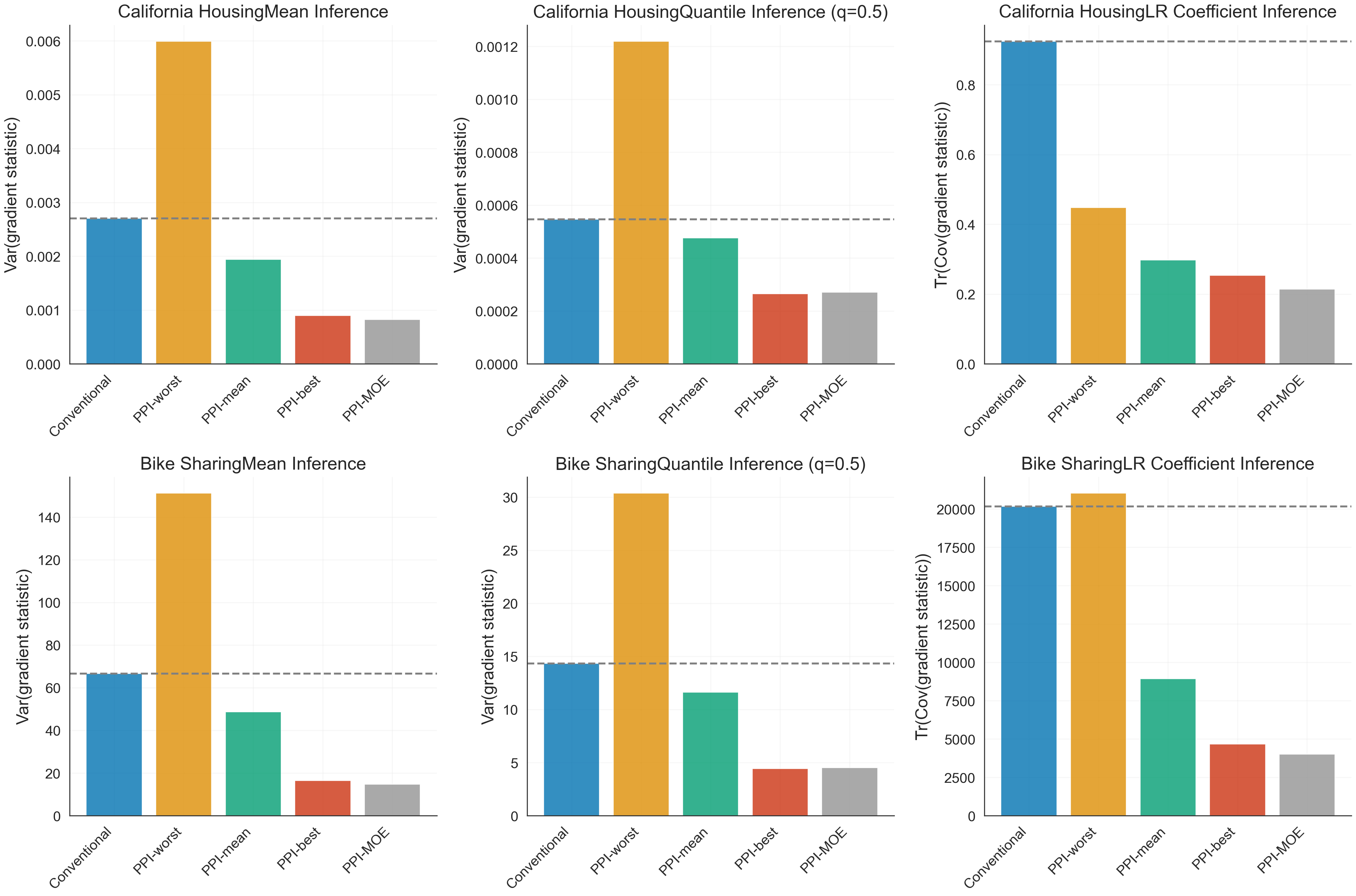}
\caption{Real-data variance comparison across inference tasks.}
\label{fig:realdata_variance}
\end{figure}

Figure~\ref{fig:realdata_variance} shows that on both datasets, PPI-MOE consistently attains the lowest variance (or is tied for the lowest) across tasks, while PPI-worst can substantially underperform due to poor predictor quality. This pattern reinforces the need for adaptive weighting rather than single-predictor selection.

\begin{figure}[ht]
\centering
\includegraphics[width=0.8\linewidth]{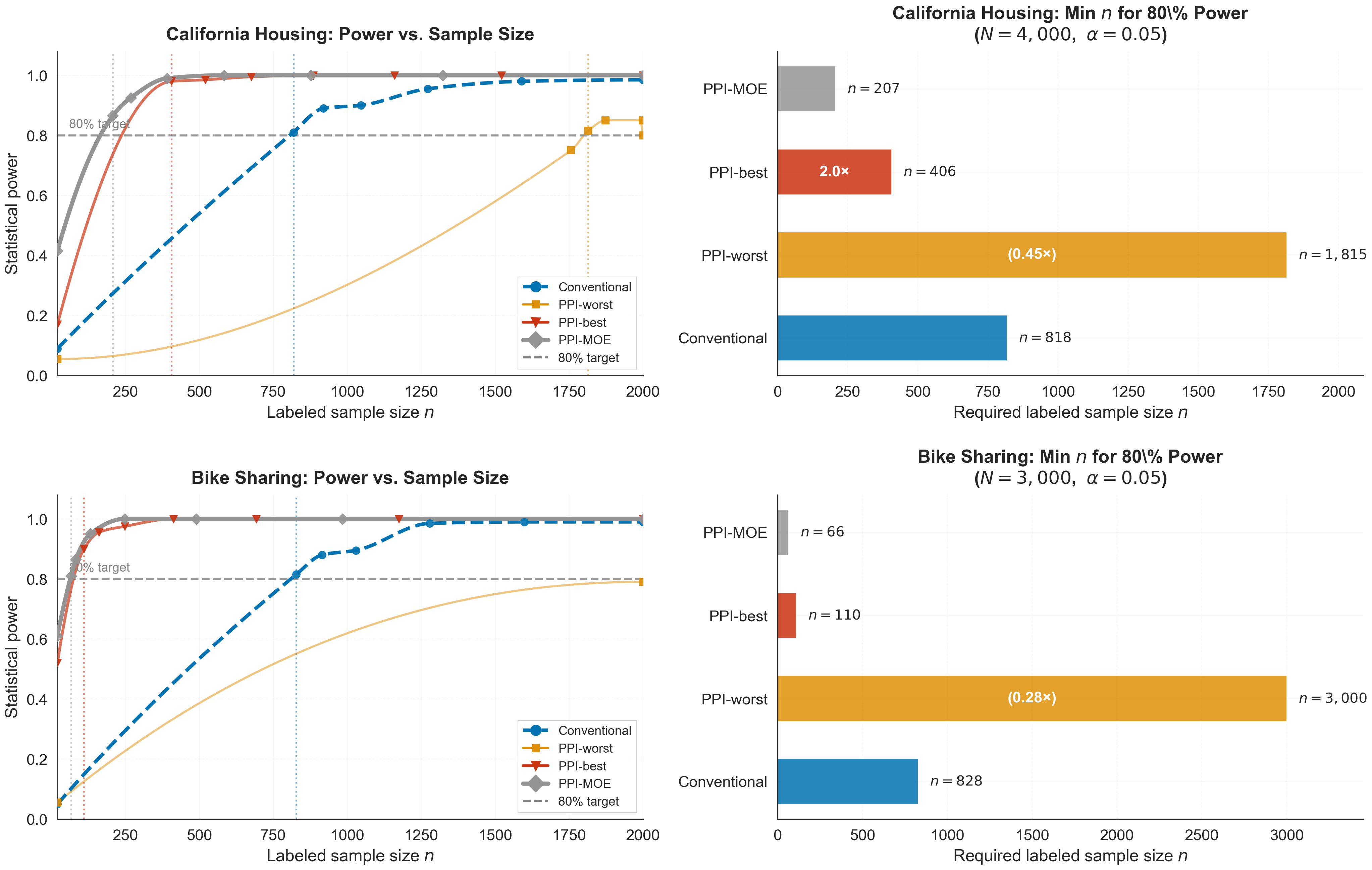}
\caption{Real-data power analysis and minimum labeled sample size for 80\% power.}
\label{fig:realdata_power}
\end{figure}

Power curves in Figure~\ref{fig:realdata_power} show that PPI-MOE and PPI-best require far fewer labeled samples than Conventional, with PPI-MOE frequently matching or exceeding PPI-best throughout the curve. These gains are operationally meaningful: the MOE strategy improves detection power while reducing annotation cost, without requiring oracle knowledge of the best single predictor. Note that PPI-worst performs considerably worse than the conventional approach under model misspecification. This suggests that blindly applying PPI may be detrimental, whereas the proposed MOE-based PPI provides a more reliable alternative.

\subsection{Empirical summary}

The numerical evidence strongly supports the central claims of this paper.
\begin{enumerate}
    \item \textit{Near-oracle adaptation without oracle selection.} Across tasks, PPI-MOE consistently achieves (and occasionally exceeds) PPI-best, whose predictor is oracle under well-specification, while avoiding infeasible oracle model selection.
    \item \textit{Robustness under misspecification.} When single predictors degrade under nonlinear misspecification, PPI-MOE remains stable and typically outperforms fixed single-model PPI baselines.
    \item \textit{Reliable uncertainty reduction.} For mean and quantile inference, PPI-MOE delivers substantial variance and interval-width reductions, with coverage close to nominal in most regimes.
    \item \textit{Large practical gains in labeling efficiency.} In the high-unlabeled regime, PPI-MOE attains target power with dramatically fewer labeled samples, yielding meaningful cost savings in data collection.
\end{enumerate}

Overall, these experiments show that inference-oriented expert aggregation is more effective than choosing a single predictor or averaging predictors uniformly. The MOE design provides a practical, adaptive, and robust route to high-quality prediction-powered inference.

\section{Discussions}
\label{sec:discuss}

We introduce a general MOE-powered inference framework that exploits unlabeled data to achieve more efficient statistical inference in the presence of multiple predictors with unknown predictive performance. Our framework is flexible and can robustly incorporate additional predictors whenever they become available. Compared with the PPI framework, it leverages the collective predictive power of multiple predictors and enjoys a best-expert guarantee relative to standard PPI. We establish non-asymptotic theory for the normal approximation of MOE-powered estimators and for the coverage properties of MOE-powered confidence sets. 

For simplicity, we focus primarily on global weighting, under which the mixture-of-experts framework reduces to classical model averaging in the statistical literature. The full power of MOE is typically realized by assigning covariate-dependent weights to the experts; for example, one may consider the linear mixture
$
F_{\beta}(x)=\sum_{k=1}^K \beta_k(x)f_k(x).
$
This reflects the belief that each expert has a domain-specific advantage over the others. However, adaptive weights pose significant challenges for estimation and inference. One possible approach is to parameterize the weight vector $\{\beta_k(x): k \in [K]\}$ through a softmax function and estimate it by maximizing variance reduction. Doing so would inevitably introduce additional bias and complicate the analysis of normal approximation and coverage properties. We leave these questions for future work.

\bibliographystyle{plainnat}
\bibliography{Bibliography-MM-MC}

\appendix

\section{Proof of Lemmas and Theorems}

\subsection{Proof of Lemma~\ref{lem:m-est-beta}}

By definition, we have $Q_{n}(\hat\beta_n)\leq Q_n(\beta_{\ast})$ which implies that 
\begin{align*}
Q(\hat\beta_n)-Q(\beta_\ast)\leq \big(Q(\hat\beta_n)-Q_n(\hat\beta_n)\big)-\big(Q(\beta_{\ast})-Q_n(\beta_{\ast})\big).
\end{align*}
For notational brevity, denote $\bmm(\beta)=\bmm_\theta(\beta, X, Y)$, $\bmm_i(\beta)=\bmm_\theta(\beta, X_i, Y_i)$ for $\forall i\in[n]$,  $\bar\bmm(\beta)=n^{-1}\sum_{i=1}^n \bmm_\theta(\beta, X_i, Y_i)$, and $\MM(\beta)=\EE \bmm_\theta(\beta, X, Y)$.   For any $\beta\in\calB$, we write 
\begin{align*}
Q_n(\beta)-Q(\beta) =&\frac{n}{n-1}\cdot\underbrace{\frac{1}{n}\sum_{i=1}^n \big[\tr\big(\bmm_i(\beta)\bmm_i^{\top}(\beta)\big)-\EE\tr\big(\bmm(\beta)\bmm^{\top}(\beta)\big)\big]}_{\calJ_1(\beta)}\\
+&\frac{n}{n-1}\underbrace{\Big[\tr\big(\MM(\beta)\MM^{\top}(\beta)\big)-\tr\big(\bar\bmm(\beta)\bar\bmm^{\top}(\beta)\big)\Big]}_{\calJ_2(\beta)}\\
+&\frac{1}{n-1}\underbrace{\Big[\EE\tr\big(\bmm(\beta)\bmm^{\top}(\beta)\big)-\tr\big(\MM(\beta)\MM^{\top}(\beta)\big)\Big]}_{=O(C_1^2)},
\end{align*}
where the last term is bounded by $O(U^2/n)$ because $\|\bmm(\beta)\|\leq U$ for all $\beta\in \calB$ and $(X,Y)\in\calX\times \calY$ under Assumption~\ref{assump:m-est}. 

Therefore, under the quadratic growth condition of Assumption~\ref{assump:m-est}, we get 
\begin{align}\label{eq:est-m-hat-err-bd1}
\tau_0\|\hat\beta_n-\beta_{\ast}\|^2\leq \big(\calJ_1(\beta_{\ast})-\calJ_1(\hat\beta_n)\big)+\frac{n}{n-1}\Big(\calJ_2(\beta_{\ast})-\calJ_2(\hat\beta_n)\Big)+O\bigg(\frac{U^2}{n}\bigg). 
\end{align}
Denote $g(\beta)=\tr\big(\bmm(\beta)\bmm^{\top}(\beta)\big)$ and $g_i(\beta)=\tr\big(\bmm_i(\beta)\bmm_i^{\top}(\beta)\big)$.  Then, 
\begin{align*}
\calJ_1(\beta_{\ast})-\calJ_1(\hat\beta_n)=\PP_n \big(g(\beta_{\ast})-g(\hat\beta_n)\big)-\PP\big(g(\beta_{\ast})-g(\hat\beta_n)\big),
\end{align*}
where $\PP\big(g(\beta_{\ast})-g(\hat\beta_n)\big):=\EE\big(g(\beta_{\ast})-g(\hat\beta_n)\big)$ and $\PP_n\big(g(\beta_{\ast})-g(\hat\beta_n)\big)=n^{-1}\sum_{i=1}^n \big(g_i(\beta_{\ast})-g_i(\hat\beta_n)\big)$.

For any $\delta>0$, define
$$
\gamma_n(\delta):=\underset{\beta\in \calB_{\ast}(\delta)}{\sup} \Big|\big(\PP_n-\PP\big)\big(g(\beta)-g(\beta_{\ast})\big) \Big|,
$$
which represents the supremum of an empirical process indexed by $\beta\in \calB_{\ast}(\delta)$ with $\calB_{\ast}(\delta):=\big\{\beta\in\calB: \|\beta-\beta_{\ast}\|\leq \delta\big\}$.  Then, $\big|\calJ_1(\hat\beta_n)-\calJ_1(\beta_{\ast}) \big|\leq \gamma_n\big(\|\hat\beta_n-\beta_{\ast}\|\big)$.  It remains to develop a uniform upper bound for $\gamma_n(\delta)$. 

\begin{lemma}\label{lem:gamma_n-bd}
Suppose Assumption~\ref{assump:m-est} holds.  Then,  there exist absolute constants $C_1>0$ such that,  for any $\delta\in\big[n^{-1}, D \big]$,  
$$
\gamma_n(\delta)\leq C_1\delta \tau_1 U\cdot \bigg(\sqrt{\frac{q}{n}}+\sqrt{\frac{\log (Dn)}{n}}+\frac{\log (Dn)}{n}\bigg),
$$
which holds with probability at least $1-n^{-10}$.
\end{lemma}

Denote the event in Lemma~\ref{lem:gamma_n-bd} by $\calE_1$. Therefore, on event $\calE_1$, 
$$
\big|\calJ_1(\hat\beta_n)-\calJ_1(\beta_{\ast})\big|\leq C_1\|\hat\beta_n-\beta_{\ast}\| \tau_1 U\cdot \bigg(\sqrt{\frac{q}{n}}+\sqrt{\frac{\log(Dn)}{n}}\bigg),
$$
assuming that $n\gg \log(Dn)$.

It remains to bound $\big|\calJ_2(\beta_{\ast})-\calJ_2(\hat\beta_n) \big|$. By definition, 
$$
\calJ_2(\beta)=-2\big<\MM(\beta), \bar\bmm(\beta)-\MM(\beta)\big>-\big\|\bar\bmm(\beta)-\MM(\beta) \big\|^2,
$$
and as a result, 
\begin{align}
\big|\calJ_2(\hat\beta_n)-\calJ_2(\beta_{\ast}) \big|\leq&\, 2\big|\big<\MM(\hat\beta_n), \bar\bmm(\hat\beta_n)-\MM(\hat\beta_n)\big>-\big<\MM(\beta_{\ast}), \bar\bmm(\beta_{\ast})-\MM(\beta_{\ast})\big> \big|\notag\\
&+\|\bar\bmm(\hat\beta_n)-\MM(\hat\beta_n)\|^2+\|\bar\bmm(\beta_{\ast})-\MM(\beta_{\ast})\|^2.\label{eq:m-est-J2-bd}
\end{align}
For any $\delta\in[n^{-1}, D]$, define
$$
\psi_n(\delta):=\sup_{\beta\in\calB_{\ast}(\delta)} \big\|(\PP_n-\PP\big)\big(\bmm(\beta)-\bmm(\beta_{\ast})\big)\big\|.
$$
The proof of the following lemma is almost identical to that of Lemma~\ref{lem:gamma_n-bd} and hence skipped. Recall that $\bmm(\beta)\in\RR^p$.  
\begin{lemma}\label{lem:psi_n-bd}
Suppose Assumption~\ref{assump:m-est} holds. Then, there exist absolute constant $C_1>0$ such that,  for any $\delta\in[n^{-1}, D]$, 
$$
\psi_n(\delta)\leq C_1\tau_1\delta \bigg(\sqrt{\frac{pq}{n}}+\sqrt{\frac{p\log(Dn)}{n}}+\frac{p\log(Dn)}{n}\bigg),
$$
which holds with probability at least $1-n^{-10}$.  
\end{lemma}

Denote the event in Lemma~\ref{lem:psi_n-bd} by $\calE_2$.  On event $\calE_2$,  we have 
$$
\big\| \big(\bar\bmm (\hat\beta_n)-\bar\bmm(\beta_{\ast})\big)-\big(\MM(\hat\beta_n)-\MM(\beta_{\ast})\big)\big\|\leq C_1\|\hat\beta_n-\beta_{\ast}\|\tau_1\cdot \bigg(\sqrt{\frac{pq}{n}}+\sqrt{\frac{p\log(Dn)}{n}}\bigg),
$$
assuming $n\gg p\log(Dn)$.  Moreover, by Bernstein inequality, there exists an event $\calE_3$ with $\PP(\calE_3)\geq 1-n^{-10}$ on which, 
$$
\big\|\bar\bmm(\beta_{\ast})-\MM(\beta_{\ast}) \big\|\leq C_2U\sqrt{\frac{\log n}{n}},
$$
for some constant $C_2>0$.  It implies, in the event $\calE_2\cap\calE_3$, that
$$
\big\|\bar\bmm(\hat\beta_n)-\MM(\hat\beta_n) \big\|\leq C_1\|\hat\beta_n-\beta_{\ast}\|\tau_1\cdot \bigg(\sqrt{\frac{pq}{n}}+\sqrt{\frac{p\log(Dn)}{n}}\bigg)+C_2U\sqrt{\frac{\log n}{n}}. 
$$
Continuing from (\ref{eq:m-est-J2-bd}),  under Assumption~\ref{assump:m-est} and the bound
\begin{align*}
    &2\big|\big<\MM(\hat\beta_n), \bar\bmm(\hat\beta_n)-\MM(\hat\beta_n)\big>-\big<\MM(\beta_{\ast}), \bar\bmm(\beta_{\ast})-\MM(\beta_{\ast})\big> \big|\\
    \leq &2\big|\big<\MM(\hat\beta_n)-\MM(\beta_\ast), \bar\bmm(\beta_\ast)-\MM(\beta_\ast)\big>\big|+2\big|\big<\MM(\hat\beta_n),(\PP_n-\PP)(\bmm(\hat\beta_n)-\bmm(\beta_\ast))\big>\big|\\
    \leq &2\big|\MM(\hat\beta_n)-\MM(\beta_\ast)\big|\cdot C_2U\sqrt{\frac{\log n}{n}} +2 C_1 U\|\hat\beta_n-\beta_\ast\|\tau_1\cdot\bigg(\sqrt{\frac{pq}{n}}+\sqrt{\frac{p\log(Dn)}{n}}\bigg)\\
    \lesssim &\tau_1U\|\hat\beta_n-\beta_\ast\|\sqrt{\frac{\log n}{n}} + U\|\hat\beta_n-\beta_\ast\|\tau_1\cdot\bigg(\sqrt{\frac{pq}{n}}+\sqrt{\frac{p\log(Dn)}{n}}\bigg),
\end{align*}
we get 
\begin{align*}
\big|\calJ_2(\hat\beta_n)-\calJ_2(\beta_{\ast})\big|\lesssim&\, U^2\frac{\log n}{n}+\|\hat\beta_n-\beta_{\ast}\|^2\tau_1^2\bigg(\frac{pq}{n}+\frac{p\log(Dn)}{n}\bigg)\\
&+\tau_1 U\|\hat\beta_n-\beta_{\ast}\|\bigg(\sqrt{\frac{pq}{n}}+\sqrt{\frac{p\log(Dn)}{n}}\bigg). 
\end{align*}

By (\ref{eq:est-m-hat-err-bd1}) and the bounds of $\calJ_1$ and $\calJ_2$, in the event $\calE_1\cap\calE_2\cap\calE_3$, we get 
$$
\|\hat\beta_n-\beta_{\ast}\|^2\lesssim U^2\bigg(\frac{\tau_0^{-1}\log n}{n}+\frac{\tau_1^2}{\tau_0^2}\cdot\frac{p(q+\log(Dn))}{n}\bigg),
$$
assuming $n\geq C_0 (\tau_1^2/\tau_0)p\big(q+\log(Dn)\big)$ for a large $C_0>1$,
which concludes the proof.

\subsection{Proof of Lemma~\ref{lem:gamma_n-bd}}
Let $\delta_0:=n^{-1}$ and define $\delta_j=2^{j-1} \delta_0$ for $j\in \big[\log (nD)\big]$.   Following the discretization procedure as in \cite[Lemma 2]{xia2019polynomial},   we derive a uniform upper bound for $\gamma_n(\delta_j)$ holding for all $j\in\big[\log(nD)\big]$,  which can be easily extended for $\gamma_n(\delta)$ for all $\delta\in[n^{-1}, D]$.  

Let us derive the upper bound for $\gamma_n(\delta_j)$ with any fixed $j$.   Note that
$$
\sup_{\beta\in\calB_{\ast}(\delta_j)} \big|g(\beta)-g(\beta_{\ast}) \big|\leq\sup_{\beta\in\calB_{\ast}(\delta_j)} \big\|\bmm(\beta)-\bmm(\beta_{\ast}) \big\|\big\|\bmm(\beta)+\bmm(\beta_{\ast}) \big\|\leq 2\tau_1 U\delta_j,
$$
where the last inequality is due to the Lipschitz and upper bound conditions from Assumption~\ref{assump:m-est}.  Similarly, 
$$
\sup_{\beta\in\calB_{\ast}(\delta_j)}\ \var\big(g(\beta)-g(\beta_{\ast})\big)\leq \sup_{\beta\in\calB_{\ast}(\delta_j)}\ \EE \big(g(\beta)-g(\beta_{\ast}) \big)^2\leq 4\tau_1^2 U^2\delta_j^2.
$$
Applying Bousquet's version of Talagrand's concentration inequality \citep{bousquet2002bennett},  with probability at least $1-e^{-t}$ for all $t>0$,  
$$
\gamma_n(\delta_j)\leq 2\EE \gamma_n(\delta_j)+2\tau_1 U\delta_j\bigg(\sqrt{\frac{t}{n}}+\frac{t}{n}\bigg).  
$$

It suffices to upper bound $\EE\gamma_n(\delta_j)$.  By the symmerization inequality \citep{koltchinskii2011oracle}, 
$$
\EE\gamma_n(\delta_j)\leq 2\EE\sup_{\beta\in\calB_{\ast}(\delta_j)}\bigg|\frac{1}{n}\sum_{i=1}^n \eps_i \big(g_i(\beta)-g_i(\beta_{\ast})\big) \bigg|,
$$
where $\eps_1,\cdots,\eps_n$ are i.i.d. Rademacher random variables.     Denote the function class $\calG_{\ast}(\delta_j):=\big\{g(\beta)-g(\beta_{\ast}): \beta\in\calB_{\ast}(\delta_j) \big\}$.  
Conditioned on $(X_1, Y_1),  \cdots,  (X_n, Y_n)$,  and for $h=g(\beta)-g(\beta_{\ast})\in \calG_{\ast}(\delta_j)$,  we denote $h(X_i, Y_i)=g_i(\beta)-g_i(\beta_{\ast})$ and define the distance $L_2(\PP_n)$ in $\calG_{\ast}(\delta_j)$ by 
$$
\|h_1-h_2\|^2_{L_2(\PP_n)}:=\frac{1}{n}\sum_{i=1}^n \big(h_1(X_i, Y_i)-h_2(X_i, Y_i)\big)^2,\quad \forall h_1, h_2\in\calG_{\ast}(\delta_j).
$$
By Dudley's entropy bound \citep[Theorem 3.10]{koltchinskii2011oracle},  we have 
$$
\EE\gamma_n(\delta_j)\leq 2\EE\sup_{h\in\calG_{\ast}(\delta_j)}\bigg|\frac{1}{n}\sum_{i=1}^n \eps_i h(X_i, Y_i) \bigg|\leq \frac{C_1}{\sqrt{n}}\EE \int_0^{\sqrt{2}\sigma_n}\sqrt{\log N\big(\calG_{\ast}(\delta_j);  L_2(\PP_n); \epsilon\big)}d\epsilon,
$$
where $\sigma_n^2:=\sup_{h\in\calG_{\ast}(\delta_j)}\PP_nh^2\leq \max_{i\in[n]}\sup_{\beta\in\calB_{\ast}(\delta_j)} |g_i(\beta)-g_i(\beta_{\ast})|^2\leq U^2\tau_1^2\delta_j^2$ and $C_1>0$ is an absolute constant.   Note that $N(\calF; d; \epsilon)$ represents the $\epsilon$-covering number of a set $\calF$ under the distance $d(\cdot, \cdot)$.  

If $h_1=g(\beta_1)-g(\beta_{\ast})$ and $h_2=g(\beta_2)-g(\beta_{\ast})$,  we have 
$$
\|h_1-h_2\|_{L_2(\PP_n)}^2=\frac{1}{n}\sum_{i=1}^n\big(g_i(\beta_1)-g_i(\beta_2)\big)^2\leq U^2\tau_1^2\|\beta_1-\beta_2\|^2,
$$
implying that $N\big(\calG_{\ast}(\delta_j);  L_2(\PP_n); \epsilon\big)\leq N\big(\calB_{\ast}(\delta_j),  \|\cdot\|,  \epsilon/(U\tau_1)\big)\leq \Big(\frac{2\delta_j U\tau_1}{\epsilon}\Big)^q$,  where the last inequality is due to the fact that $\calB_{\ast}(\delta_j)\subset \RR^q$ has a diameter at most $2\delta_j$.   Plugging it into the Dudley's entropy bound,  we get 
$$
\EE\gamma_n(\delta_j)\leq \frac{C_1}{\sqrt{n}} \EE\int_0^{2U\tau_1\delta_j}\sqrt{q}\sqrt{\log \frac{2\delta_j U\tau_1}{\epsilon}} d\epsilon=O\bigg(\delta_j U\tau_1\cdot \sqrt{\frac{q}{n}}\bigg).
$$
Therefore,  with probability at least $1-e^{-t}$,  
$$
\gamma_n(\delta_j)\leq C_1\tau_1 U\delta_j\bigg(\sqrt{\frac{q}{n}}+\sqrt{\frac{t}{n}}+\frac{t}{n}\bigg).  
$$
By setting $t=C_2\log(Dn)$ for a large $C_2>0$ and taking a union bound for all $j\in[\log(Dn)]$,  we get with probability at least $1-n^{-10}$,  
$$
\gamma_n(\delta_j)\leq C_1\tau_1 U\delta_j\bigg(\sqrt{\frac{q}{n}}+\sqrt{\frac{\log (Dn)}{n}}+\frac{\log (Dn)}{n}\bigg),\quad \textrm{ for all } j=1,2,\cdots, \log(Dn).
$$
By adjusting the constant $C_1$ and chaining through all the intervals $[C_j,  C_{j+1}]$,  we can extend the above bounds to all $\delta\in[n^{-1},  D]$,  which concludes the proof.

\subsection{Proof of Theorem~\ref{thm:m-est}}
By definition, we decompose the estimating function $\tilde \bg_{\theta,\hat\beta_n} + \hat\bDelta_{\theta,\hat\beta_n}$ as 
\begin{equation}\label{eq:m-est-decompose}
    \begin{split}
        &\tilde\bg_{\theta, \hat\beta_n} + \hat \bDelta_{\theta, \hat\beta_n} \\
        =\,& N^{-1}\sum_{i=1}^N\bg_\theta(\tilde X_i,F_{\hat\beta_n}(\tilde X_i))+n^{-1}\sum_{i=1}^n \Big(\bg_\theta(X_i,Y_i) - \bg_\theta(X_i, F_{\hat\beta_n}(X_i))\Big)\\
        =\,& \underbrace{N^{-1}\sum_{i=1}^N\Big[\bg_\theta(\tilde X_i,F_{\beta_\ast}(\tilde X_i))-\EE\bg_\theta(X,F_{\beta_\ast}(X))\Big]}_{N^{-1/2}\tilde \bZ_{N, \theta, \beta_\ast}} + \underbrace{N^{-1}\sum_{i=1}^N\Big[\bg_\theta(\tilde X_i,F_{\hat\beta_n}(\tilde X_i))-\bg_\theta(\tilde X_i,F_{\beta_\ast}(\tilde X_i))\Big]}_{\tilde \calJ_1}\\
        &+\underbrace{n^{-1}\sum_{i=1}^n \Big[\bg_\theta(X_i,Y_i) - \bg_\theta(X_i, F_{\beta_\ast}(X_i))-\big(\EE\bg_\theta(X,Y) - \EE\bg_\theta(X, F_{\beta_\ast}(X))\big)\Big]}_{n^{-1/2}\bZ_{n,\theta, \beta_\ast}}\\
        &-\underbrace{n^{-1}\sum_{i=1}^n \Big[\bg_\theta(X_i, F_{\hat\beta_n}(X_i))-\bg_\theta(X_i, F_{\beta_\ast}(X_i))\Big]}_{\calJ_1}+\EE\bg_\theta(X, Y).
    \end{split}
\end{equation}
By the definition of $\bZ_{n,\theta,\beta_\ast}$ and $\tilde \bZ_{N,\theta,\beta_\ast}$, the existence of finite $\cov\big(\bg_\theta(X,F_{\beta_\ast}(X)\big)$ and the boundedness of $\bmm_\theta(\beta,X,Y)$, the Central Limit Theorem indicates that
\begin{align}\label{eq:m-est-normality-base}
    \bZ_{n,\theta,\beta_\ast}\overset{d.}{\to}\calN\Big(0,\cov\big(\bmm_\theta(\beta_\ast, X, Y)\big)\Big)\quad \text{and} \quad\tilde \bZ_{N,\theta,\beta_\ast}\overset{d.}{\to}\calN\Big(0,\cov\big(\bg_\theta(X, F_{\beta_\ast}(X))\big)\Big)
\end{align}

\noindent{\it Step 1: bias}. 
First, we derive the bound of the bias of $\tilde\bg_{\theta, \hat\beta_n} + \hat \bDelta_{\theta, \hat\beta_n}$. Note that
\begin{align*}
    \EE \big(\tilde\bg_{\theta, \hat\beta_n} + \hat \bDelta_{\theta, \hat\beta_n}\big)-\EE\bg_\theta(X,Y) =\EE \tilde\calJ_1-\EE\calJ_1.
\end{align*}

By Lemma~\ref{lem:gamma_n-bd}, under the event in Lemma~\ref{lem:m-est-beta} with probability at least $1-n^{-9}$, we have
\begin{align}
    \big\|\tilde J_1 -J_1\big\| &\leq \big\|\big(\PP_n-\PP\big)\big(\bmm( \hat\beta_n) -  \bmm(\beta_\ast)\big)\big\| + \big\|\big(\PP_N-\PP\big)\big(\bmm( \hat\beta_n) -  \bmm(\beta_\ast)\big)\big\|\notag\\
    &\leq \psi_n(\delta_n) + \psi_N(\delta_n)\notag\\
    &\leq C_1\tau_1\delta_n \bigg(\sqrt{\frac{pq}{n}}+\sqrt{\frac{p\log(Dn)}{n}}+\frac{p\log(Dn)}{n}+\sqrt{\frac{pq}{N}}+\sqrt{\frac{p\log(DN)}{N}}+\frac{p\log(DN)}{N}\bigg)\notag\\
    &\lesssim \frac{p(q+\log(Dn))}{n} + \sqrt\frac{{p^2(q+\log(Dn))(q+\log(DN))}}{{nN}}.\label{eq:m-est-J1-diff}
\end{align}
where we set $\delta_n= C_2U\bigg(\sqrt\frac{\tau_0^{-1}\log n}{n}+\frac{\tau_1}{\tau_0}\cdot\sqrt\frac{p(q+\log(Dn))}{n}\bigg)$ for a sufficiently large $C_2>0$.

If $n\gtrsim p\big(q+\log(Dn)\big)$ and $N\gtrsim n$, then 
\begin{align}\label{eq:m-est-J1-bias}
    \big\|\EE\tilde \calJ_1-\EE\calJ_1\big\|\leq \EE\|\tilde \calJ_1-\calJ_1\big\|\lesssim \frac{p(q+\log(Dn))}{n}.
\end{align}

Take (\ref{eq:m-est-J1-bias}) back into the expectation form of (\ref{eq:m-est-decompose}), combining with (\ref{eq:m-est-normality-base}), we finally have the bias as
\begin{align}\label{eq:m-est-bias}
    \Big\|\EE \big(\tilde\bg_{\theta, \hat\beta_n} + \hat \bDelta_{\theta, \hat\beta_n}\big) - \EE\bg_\theta(X,Y)\Big\| = O\Big(\frac{p(q+\log(Dn))}{n}\Big).
\end{align}
Especially, if take $\theta=\theta_\ast$ so that $\EE\bg_{\theta_{\ast}}(X,Y)=0$, we have 
$$
\Big\|\EE \big(\tilde\bg_{\theta_\ast, \hat\beta_n} + \hat \bDelta_{\theta_\ast, \hat\beta_n}\big)\big\| = O\Big(\frac{p(q+\log(Dn))}{n}\Big).
$$

\noindent{\it Step 2: normal approximation}. 
By (\ref{eq:m-est-decompose}) and (\ref{eq:m-est-J1-diff}), we have
\begin{align*}
    \sqrt n\Big(\tilde\bg_{\theta, \hat\beta_n} + \hat \bDelta_{\theta, \hat\beta_n} -\EE\bg_\theta(X,Y)\Big)=\bZ_{n,\theta,\beta_\ast}+ \sqrt{\frac{n}{N}}\tilde \bZ_{N, \theta,\beta_\ast}+\tilde O_p\bigg(\frac{p\big(q+\log(Dn)\big)}{\sqrt{n}}\bigg).
\end{align*}
Bernstein Inequality implies there exists an event $\calE_1$ with $\PP(\calE_1)\geq 1-n^{-10}$, on which 
\begin{align*}
    \big\|\tilde \bZ_{N, \theta, \beta_\ast}\big\| \leq O\big(\sqrt{p\log n}\big),
\end{align*}
implying that
\begin{align*}
    \sqrt n\Big(\tilde\bg_{\theta, \hat\beta_n} + \hat \bDelta_{\theta, \hat\beta_n} -\EE\bg_\theta(X,Y)\Big)=\bZ_{n,\theta,\beta_\ast}+\tilde O_p\bigg({\frac{p(q+\log(Dn))}{\sqrt{n}}}+\sqrt{\frac{pn\log(n)}{N}}\bigg).
\end{align*}

By Berry-Esseen bound and the high probability bound inherited from $\tilde O_p(\cdot)$, we get 
\begin{align}\label{eq:m-est-moe-berry}
    \sup_{t\in\RR}\Bigg|\PP\bigg(\frac{\sqrt n\, \be_s^\top\big(\tilde\bg_{\theta_\ast, \hat\beta_n}+\hat\bDelta_{\theta_\ast,\hat\beta_n}\big)}{\sqrt{\be_s^\top\bW_{\theta_\ast, Y-F_{\beta_\ast}}\be_s}}\leq t\bigg)\Bigg|=O\bigg(\frac{p(q+\log(Dn))}{\sqrt{n}}+\sqrt{\frac{pn\log(n)}{N}}\bigg),
\end{align}
where $\bW_{\theta,Y-F_{\beta_\ast}}= \cov\big(\bg_{\theta}(X,Y)-\bg_{\theta}(X, F_{\beta_\ast}(X))\big)$. Therefore, for all $s\in[p]$,
\begin{align*}
    \PP\bigg(\bigg|\frac{\sqrt n\, \be_s^\top\big(\tilde\bg_{\theta_\ast, \hat\beta_n}+\hat\bDelta_{\theta_\ast,\hat\beta_n}\big)}{\sqrt{\be_s^\top\bW_{\theta_\ast, Y-F_{\beta_\ast}}\be_s}}\bigg|\leq z_{\alpha/(2p)}\bigg)\geq 1-\frac{\alpha}{p}+O\bigg(\frac{p(q+\log(Dn))}{\sqrt{n}}+\sqrt{\frac{pn\log(n)}{N}}\bigg).
\end{align*}

It remains to bound $\Big\|\hat\bW_{\theta_\ast, Y-F_{\hat\beta_n}}-\bW_{\theta_\ast, Y-F_{\beta_\ast}}\Big\|$:
\begin{align*}
    &\Big\|\hat\bW_{\theta_\ast, Y-F_{\hat\beta_n}}-\bW_{\theta_\ast, Y-F_{\beta_\ast}}\Big\|\\
    =\,&\Big\|\scov\big(\bmm_{\theta_\ast}(\hat\beta_n,X,Y) \big)-\cov\big(\bmm_{\theta_\ast}(\beta_\ast,X,Y) \big)\Big\|\\
    =\,&\Bigg\|n^{-1}\sum_{i=1}^n\bmm_{\theta_\ast}(\hat\beta_n,X_i,Y_i)^{\otimes2} - n^{-1}\sum_{i=1}^n\bmm_{\theta_\ast}(\beta_\ast,X_i,Y_i)^{\otimes2}\Bigg\|\\
    &+\Bigg\|\Big[n^{-1}\sum_{i=1}^n\bmm_{\theta_\ast}(\hat\beta_n,X_i,Y_i)\Big]^{\otimes2}-\Big[n^{-1}\sum_{i=1}^n\bmm_{\theta_\ast}(\beta_\ast,X_i,Y_i)\Big]^{\otimes2}\Bigg\|\\
    &+\Bigg\|n^{-1}\sum_{i=1}^n\bmm_{\theta_\ast}(\beta_\ast,X_i,Y_i)^{\otimes2} - \EE\bmm_{\theta_\ast}(\beta_\ast,X,Y)^{\otimes2}\Bigg\|\\
    &+\Bigg\|\Big[n^{-1}\sum_{i=1}^n\bmm_{\theta_\ast}(\beta_\ast,X_i,Y_i)\Big]^{\otimes2}-\Big[\EE\bmm_{\theta_\ast}(\beta_\ast,X,Y)\Big]^{\otimes2}\Bigg\|,
\end{align*}
where $\bx^{\otimes2}=\bx\bx^\top$ refers to the Kronecker product.

By the Lipschitz condition of $\bmm_\theta(\beta,X,Y)$ with respect to $\beta$, we have
\begin{equation}\label{eq:m-est-var-est-1}
    \begin{split}
        &\max \Bigg\{\Bigg\|n^{-1}\sum_{i=1}^n\bmm_{\theta_\ast}(\hat\beta_n,X_i,Y_i)^{\otimes2} - n^{-1}\sum_{i=1}^n\bmm_{\theta_\ast}(\beta_\ast,X_i,Y_i)^{\otimes2}\Bigg\|\\
        &\quad+\Bigg\|\Big[n^{-1}\sum_{i=1}^n\bmm_{\theta_\ast}(\hat\beta_n,X_i,Y_i)\Big]^{\otimes2}-   \Big[n^{-1}\sum_{i=1}^n\bmm_{\theta_\ast}(\beta_\ast,X_i,Y_i)\Big]^{\otimes2}\Bigg\|\Bigg\} \\
        \lesssim&\,  2\tau_1U\Big\|\hat\beta_n-\beta_\ast\Big\|=O\Big(\sqrt{\frac{p\log (Dn)}{n}}\Big),
    \end{split}
\end{equation}
under the event of Lemma~\ref{lem:m-est-beta}.
Since $\bmm_\theta(\beta,X,Y)$ is bounded, by matrix Bernstein inequality, there exist an event $\calE_2$ with $\PP(\calE_2)\geq 1-n^{-10}$, on which
\begin{equation}\label{eq:m-est-var-est-2}
    \begin{split}
        \max \Bigg\{&\Bigg\|n^{-1}\sum_{i=1}^n\bmm_{\theta_\ast}(\beta_\ast,X_i,Y_i)^{\otimes2} - \EE\bmm_{\theta_\ast}(\beta_\ast,X,Y)^{\otimes2}\Bigg\| \\
        &\Bigg\|\Big[n^{-1}\sum_{i=1}^n\bmm_{\theta_\ast}(\beta_\ast,X_i,Y_i)\Big]^{\otimes2}-\Big[\EE\bmm_{\theta_\ast}(\beta_\ast,X,Y)\Big]^{\otimes2}\Bigg\|\Bigg\} = O\Big(\sqrt{\frac{p\log n}{n}}\Big).
    \end{split}
\end{equation}

Combining (\ref{eq:m-est-var-est-1}) and (\ref{eq:m-est-var-est-2}), we conclude that
\begin{align*}
    \Big\|\hat\bW_{\theta_\ast, Y-F_{\hat\beta_n}}-\bW_{\theta_\ast, Y-F_{\beta_\ast}}\Big\|=O\Big(\sqrt{\frac{p\log (Dn)}{n}}\Big).
\end{align*}

For all $s\in[p]$,
\begin{align*}
  &\frac{\sqrt n\, \be_s^\top\big(\tilde\bg_{\theta_\ast, \hat\beta_n}+\hat\bDelta_{\theta_\ast,\hat\beta_n}\big)}{\sqrt{\be_s^\top\hat\bW_{\theta_\ast, Y-F_{\hat\beta_n}}\be_s}}\\
  = &\frac{\sqrt n\, \be_s^\top\big(\tilde\bg_{\theta_\ast, \hat\beta_n}+\hat\bDelta_{\theta_\ast,\hat\beta_n}\big)}{\sqrt{\be_s^\top\bW_{\theta_\ast, Y-F_{\beta_\ast}}\be_s}}\Bigg(1+\frac{{{\be_s^\top\Big(\bW_{\theta_\ast, Y-F_{\beta_\ast}}-\hat\bW_{\theta_\ast, Y-F_{\hat\beta_n}}\Big)\be_s}}}{\sqrt{\be_s^\top\hat\bW_{\theta_\ast, Y-F_{\hat\beta_n}}\be_s}\bigg(\sqrt{\be_s^\top\bW_{\theta_\ast, Y-F_{\beta_\ast}}\be_s}+\sqrt{\be_s^\top\hat\bW_{\theta_\ast, Y-F_{\hat\beta_n}}\be_s}\bigg)}\Bigg)\\
  = &\frac{\sqrt n\, \be_s^\top\big(\tilde\bg_{\theta_\ast, \hat\beta_n}+\hat\bDelta_{\theta_\ast,\hat\beta_n}\big)}{\sqrt{\be_s^\top\bW_{\theta_\ast, Y-F_{\beta_\ast}}\be_s}} + \tilde O_p\Big(\frac{p\log n}{\sqrt{n}}\Big).
\end{align*}

Plugging into (\ref{eq:m-est-moe-berry}), we have 
\begin{align}\label{eq:m-est-moe-berry-1}
    \sup_{t\in\RR}\Bigg|\PP\bigg(\frac{\sqrt n\, \be_s^\top\big(\tilde\bg_{\theta_\ast, \hat\beta_n}+\hat\bDelta_{\theta_\ast,\hat\beta_n}\big)}{\sqrt{\be_s^\top\hat\bW_{\theta_\ast, Y-F_{\hat\beta_n}}\be_s}}\leq t\bigg)\Bigg|=O\bigg(\frac{p(q+\log(Dn))}{\sqrt{n}}+\sqrt{\frac{pn\log(n)}{N}}\bigg),
\end{align}
implying
\begin{align*}
    \PP\bigg(\bigg|\frac{\sqrt n\, \be_s^\top\big(\tilde\bg_{\theta_\ast, \hat\beta_n}+\hat\bDelta_{\theta_\ast,\hat\beta_n}\big)}{\sqrt{\be_s^\top\hat\bW_{\theta_\ast, Y-F_{\hat\beta_n}}\be_s}}\bigg|\leq z_{\alpha/(2p)}\bigg)\geq 1-\frac{\alpha}{p}+O\bigg(\frac{p(q+\log(Dn))}{\sqrt{n}}+\sqrt{\frac{pn\log(n)}{N}}\bigg).
\end{align*}

The coverage probability satisfies that
\begin{align*}
    \PP\Big(\theta_\ast \in \calC_\alpha^\moe\Big) &= \PP\Bigg(\bigcap_{s=1}^p\Bigg\{\bigg|\frac{\sqrt n\, \be_s^\top\big(\tilde\bg_{\theta_\ast, \hat\beta_n}+\hat\bDelta_{\theta_\ast,\hat\beta_n}\big)}{\sqrt{\be_s^\top\hat\bW_{\theta_\ast, Y-F_{\hat\beta_n}}\be_s}}\bigg|\leq z_{\alpha/(2p)}\Bigg\}\Bigg)\\
    &\geq 1-\sum_{s=1}^p\PP\Bigg(\Bigg\{\bigg|\frac{\sqrt n\, \be_s^\top\big(\tilde\bg_{\theta_\ast, \hat\beta_n}+\hat\bDelta_{\theta_\ast,\hat\beta_n}\big)}{\sqrt{\be_s^\top\hat\bW_{\theta_\ast, Y-F_{\hat\beta_n}}\be_s}}\bigg|> z_{\alpha/(2p)}\Bigg\}\Bigg)\\
    &=1-\Bigg(\frac{\alpha}{p}+O\bigg(\frac{p(q+\log(Dn))}{\sqrt{n}}+\sqrt{\frac{pn\log(n)}{N}}\bigg)\Bigg)^p\\
    &=1-\alpha+O\bigg(\frac{p^2(q+\log(Dn))}{\sqrt{n}}+\sqrt{\frac{p^3n\log(n)}{N}}\bigg),
\end{align*}
if $n\gtrsim p^4\big(q^2+\log^2(Dn)\big)$ and $N\gtrsim p^3n\log n$, which concludes the proof.

\subsection{Proof of Lemma~\ref{lem:mean-moe-bias}}
By (\ref{eq:mean-moe-beta-1}) and the fact $\EE n^{-1}{\bf 1}_n^{\top}\by=\theta_{\ast}$,  we have 
\begin{align*}
\EE \hat\theta^{\moe}-\theta_{\ast}=&\EE \Big[\big(N^{-1}{\bf 1}_N^{\top}\tilde \bF-n^{-1}{\bf 1}_n^{\top}\bF\big)(\bF^{\top}\bP_n\bF)^{-1}\bF^{\top}\bP_n\by \Big]\\
=&\EE \Big[\big(N^{-1}{\bf 1}_N^{\top}\tilde \bF-n^{-1}{\bf 1}_n^{\top}\bF\big)(\bF^{\top}\bP_n\bF)^{-1}\bF^{\top}\bP_n\big(\by -\bF\beta_{\ast}\big)\Big],
\end{align*}
where the last equality holds since $\EE[N^{-1}{\bf 1}_N^{\top}\tilde\bF]=\EE[n^{-1}{\bf 1}_n^{\top}\bF]=\EE \bff_1^{\top}$.   Then,  
\begin{equation}\label{eq:mean-moe-bias-pf-1}
\Big|\EE\hat\theta^{\moe}-\theta_{\ast}  \Big|\leq \EE \big\|N^{-1}{\bf 1}_N^{\top}\tilde \bF-n^{-1}{\bf 1}_n^{\top}\bF \big\|\cdot \big\| (\bF^{\top}\bP_n\bF)^{-1}\bF^{\top}\bP_n\big(\by -\bF\beta_{\ast}\big)\big\|.
\end{equation}
By the matrix Bernstein inequality \citep{tropp2012user},  there is an event $\calE_0$ with $\PP(\calE_0)\geq 1-n^{-10}$ on which, 
$$
\max\bigg\{\bigg\|\frac{1}{n}\sum_{i=1}^n \bff_i\bff_i^{\top}-\EE \bff\bff^{\top} \bigg\|,\ \bigg\|\frac{1}{n}\bF^{\top}\by-\EE Y\bff\bigg\|,\ \big\|\bar\bff -\EE\bff\big\|\bigg\}\leq C_1\sqrt{\frac{K\log n}{n}},
$$
and $|\bar Y-\EE Y|\leq C_1/\sqrt{n}$ for some absolute constant $C_1>0$,  where $\bar Y=n^{-1}{\bf 1}_n^{\top}\by$. Therefore, conditioned on event $\calE_0$, we have 
$$
\bF^{\top}\bP_n\bF=n\cdot \bigg(\frac{1}{n}\sum_{i=1}^n \bff_i\bff_i^{\top}-\bar\bff \bar\bff^{\top}\bigg)
$$
implying that $\big\|\bF^{\top}\bP_n\bF -n\cdot \cov(\bff)\big\|_{\rm F}=O\big(K\sqrt{n\log n}\big)$. Similarly, conditioned on event $\calE_0$, we have 
\begin{align*}
\big\|\bF^{\top}\bP_n\by-n\cdot \cov(Y,\bff) \big\|=n\cdot \bigg\|\frac{1}{n}\sum_{i=1}^nY_i\bff_i -\bar Y \bar\bff-\cov(Y,\bff)\bigg\|=O\Big(\sqrt{Kn\log n}\Big). 
\end{align*}
As a result, conditioned on $\calE_0$ and if $n\geq C_2K^2\log n$ for a large constant $C_2>0$,  we have $\big\|(\bF^{\top}\bP_n\bF)^{-1}\big\|=O(n^{-1})$ and 
\begin{align*}
\big\|\bF^{\top}\bP_n(\by-\bF\beta_{\ast}) \big\|\leq & \big\|\bF^{\top}\bP_n\by-n\cdot\cov(Y,\bff)\big\|+\big\|\big(\bF^{\top}\bP_n\bF -n\cdot \cov(\bff)\big)\beta_{\ast}\big\|\\
\leq& C_1\sqrt{K^3n\log n}.
\end{align*}
By Bernstein inequality, there exist an event $\calE_1$ with $\PP(\calE_1)\geq 1-n^{-10}$ on which,  
$$
\big\|N^{-1}{\bf 1}_N\tilde\bF-n^{-1}{\bf 1}_n^{\top}\bF \big\|\leq C_1\sqrt{\frac{K\log n}{n}},
$$
we we assumed $N\gg n$. Finally, continuing from (\ref{eq:mean-moe-bias-pf-1}), we get 
\begin{align*}
\Big|\EE\hat\theta^{\moe}-\theta_{\ast}  \Big|\leq& \EE \big\|N^{-1}{\bf 1}_N^{\top}\tilde \bF-n^{-1}{\bf 1}_n^{\top}\bF \big\|\cdot \big\| (\bF^{\top}\bP_n\bF)^{-1}\bF^{\top}\bP_n\big(\by -\bF\beta_{\ast}\big)\big\|\II_{\calE_0\cap \calE_1}\\
+&\EE \big\|N^{-1}{\bf 1}_N^{\top}\tilde \bF-n^{-1}{\bf 1}_n^{\top}\bF \big\|\cdot \big\| (\bF^{\top}\bP_n\bF)^{-1}\bF^{\top}\bP_n\big(\by -\bF\beta_{\ast}\big)\big\|\II_{\calE_0^{\rm c}\cup \calE_1^{\rm c}}\leq C_1\cdot \frac{K^2\log n}{n},
\end{align*}
which concludes the proof.

\subsection{Proof of Theorem~\ref{thm:mean_estimation}}

Recall that $\hat\beta_n=(\bF^{\top}\bP_n \bF)^{-1}\bF^{\top}\bP_n\by$ and $\beta_{\ast}=\big(\cov(\bff)\big)^{-1}\cov(\bff, Y)$. We first derive the upper bound for $\|\hat\beta_n-\beta_{\ast}\|$. Note that
\begin{align*}
\hat\beta_n-\beta_{\ast}=\Big( (\bF^{\top}\bP_n \bF)^{-1}-\big(n\cov(\bff)\big)^{-1}\Big)\bF^{\top}\bP_n\by+\big(\cov(\bff)\big)^{-1}\Big(n^{-1}\bF^{\top}\bP_n\by-\cov(\bff, Y)\Big).
\end{align*}
As show in the proof of Lemma~\ref{lem:mean-moe-bias}, there exists an event $\calE_0$ with $\PP(\calE_0)\geq 1-n^{-10}$, on which the following bounds hold
\begin{align*}
\big\|\bF^{\top}\bP_n\bF-n\cov(\bff) \big\|_{\rm F}=O\Big(K\sqrt{n\log n}\Big)\qquad {\rm and}\qquad \big\|\bF^{\top}\bP_n\by-n\cov(\bff, Y) \big\|=O\Big(\sqrt{Kn\log n}\Big),
\end{align*}
and consequently $\|\bF^\top\bP_n\by\|=O(n)$.
Therefore, conditioned on event $\calE_0$, we get
\begin{align}
\big\|\hat\beta_n-\beta_{\ast} \big\|\leq & \|(\bF^{\top}\bP_n\bF)^{-1}\|\|\bF^{\top}\bP_n\bF-n\cov(\bff)\|  \big\|\big(n\cov(\bff)\big)^{-1}\big\|\|\bF^{\top}\bP_n\by\|\notag\\
+&\big\|\big(n\cov(\bff)\big)^{-1}\big\| \big\|\bF^{\top}\bP_n\by - n\cov(\bff, Y) \big\|=O\bigg(\sqrt{\frac{K^3\log n}{n}}\bigg),\label{eq:hatbeta_n-err-1}
\end{align}
where we assumed $n\gg K^2\log n$.

Denote the optimal MOE by $f_{\ast}(X)=\langle \beta_{\ast}, \bff\rangle=\sum_{k=1}^K \beta_{\ast, k}f_k(X)$.  We show that the difference between $\hat\theta^{\moe}$ and $\hat\theta^{\ppi}_{f_*}$ (recall its definition in (\ref{eq:mean-ppi})) is negligible. By definition,
\begin{align*}
\big\|\hat\theta^{\moe}-\hat\theta^{\ppi}_{f_*}\big\|
&=\big\|\big(N^{-1}{\bf 1}_N^{\top}\tilde \bF-n^{-1}{\bf 1}_n^{\top}\bF\big)(\hat\beta_n-\beta^*)\big\| \\
&\leq\big\|N^{-1}{\bf 1}_N^{\top}\tilde \bF-n^{-1}{\bf 1}_n^{\top}\bF \big\| \big\|\hat\beta_n-\beta_{\ast} \big\|.
\end{align*}
By Bernstein inequality, there is an event $\calE_1$ with $\PP(\calE_1)\geq 1-n^{-10}$, on which the following bounds hold
\begin{equation*}
\big\|N^{-1}{\bf 1}_N^{\top}\tilde\bF-\EE \bff\big\|=O\big(\sqrt{(K/N)\log n}\big),
\qquad {\rm and}\qquad
\big\|n^{-1}{\bf 1}_n^{\top}\bF-\EE \bff\big\|=O\big(\sqrt{(K/n)\log n}\big).
\end{equation*}
Together with (\ref{eq:hatbeta_n-err-1}), conditioned on event $\calE_0\cap \calE_1$, we get 
\begin{equation}\label{eq:moe-ppi-diff}
\Big\|\hat\theta^{\moe}-\hat\theta_{f_{\ast}}^{\ppi} \Big\|=O\bigg(\frac{K^2\log n}{n}\bigg).
\end{equation}
Thus it suffices to focus on the classical PPI-based estimator $\hat\theta_{f_{\ast}}^{\ppi}$ equipped with the optimal MOE as the predictor.  

Recall $\hat\theta_{f_{\ast}}^{\ppi}=N^{-1}{\bf 1}_N^{\top}\tilde\bF\beta_{\ast}- n^{-1}{\bf 1}_n^{\top}\big(\bF\beta_{\ast}-\by\big)$ implying that 
\begin{align}\label{eq:ppi-fstar-rep}
\sqrt{n}\big(\hat\theta_{f_{\ast}}^{\ppi}-\theta_{\ast}\big)=\sqrt{n}\bigg(\frac{1}{N}\sum_{i=1}^N f_{\ast}(\tilde X_i)-\EE f_{\ast}(X)\bigg)-\frac{1}{\sqrt{n}}\sum_{i=1}^n \Big(\big(f_{\ast}(X_i)-Y_i\big)-\EE\big(f_{\ast}(X)-Y\big)\Big)
\end{align}
The classical central limit theorem dictates that 
$$
Z_{n,f_{\ast}}:=-\frac{1}{\sqrt{n}}\sum_{i=1}^n \Big(\big(f_{\ast}(X_i)-Y_i\big)-\EE\big(f_{\ast}(X)-Y\big)\Big) \xrightarrow{d} \calN\Big(0, \var\big(f_{\ast}(X)-Y\big)\Big)
$$
and the Chebyshev's inequality gives that 
$$
\sqrt{n}\bigg(\frac{1}{N}\sum_{i=1}^N f_{\ast}(\tilde X_i)-\EE f_{\ast}(X)\bigg)=O_p\bigg(\sqrt{\frac{n}{N}\cdot \var\big(f(X)\big)}\bigg).
$$ 
Moreover, by Bernstein inequality, we get 
\begin{equation*}
\sqrt n\big(\hat\theta^{\mathrm{PPI}}_{f^*}-\theta_{\ast}\big)
=
Z_{n,f_{\ast}}
+\tilde O_p\bigg(\sqrt{\frac{n\log n}{N}}\bigg)
\end{equation*}
where $Z_{n,f_{\ast}}\to_d \calN\bigl(0,\var(Y-f_{\ast}(X))\bigr)$ as $n\to\infty$. Bound (\ref{eq:moe-ppi-diff}) shows that
$$
\sqrt{n}\Big(\hat\theta^{\moe}-\hat\theta^{\ppi}_{f_{\ast}}\Big)=\tilde O_p\bigg(\frac{K^2\log n}{\sqrt{n}}\bigg).
$$
Finally, we conclude that 
\begin{equation}\label{eq:sqrtn-moe-theta}
\sqrt n\big(\hat\theta^{\moe}-\theta_{\ast}\big)
=
Z_{n,f_{\ast}}
+\tilde O_p\bigg(\frac{K^2\log n}{\sqrt{n}}+\sqrt{\frac{n\log n}{N}}\bigg).
\end{equation}
By the Berry-Esseen bound and the high probability bound inherited from $\tilde O_p(\cdot)$, we get 
\begin{align}\label{eq:mean-moe-berry}
\sup_{t\in\RR}\bigg|\PP\bigg(\frac{\sqrt{n}(\hat\theta^{\moe}-\theta_{\ast})}{\sigma_{Y-f_{\ast}}}\leq t\bigg) -\Phi(t)\bigg|=O\bigg(\frac{K^2\log n}{\sqrt{n}}+\sqrt{\frac{n\log n}{N}}\bigg),
\end{align}
where $\sigma_{Y-f_{\ast}}^2:=\var(Y-f_{\ast}(X))$. Therefore,
\begin{align*}
\PP\Big(\theta_{\ast}\in \Big[\hat\theta^{\moe}-z_{\alpha/2}\frac{\sigma_{Y-f_{\ast}}}{\sqrt{n}},\  \hat\theta^{\moe}+z_{\alpha/2}\frac{\sigma_{Y-f_{\ast}}}{\sqrt{n}}\Big]\Big)=1-\alpha+O\bigg(\frac{K^2\log n}{\sqrt{n}}+\sqrt{\frac{n\log n}{N}}\bigg).
\end{align*}

It suffices to bound $\big|\hat\sigma^2_{Y-f_{\ast}}-\sigma^2_{Y-f_{\ast}}\big|$.  Define
\begin{equation*}
\hat\varepsilon_i:=Y_i-\langle \bff_i, \hat\beta_n\rangle
\qquad {\rm and}\qquad 
\varepsilon_i^*:=Y_i-\langle \bff_i, \beta_{\ast}\rangle,\qquad {\forall} i\in[n].
\end{equation*}
By definition, we write
\begin{equation*}
\hat\sigma_{Y-f_{\ast}}^2=\frac{1}{n}\sum_{i=1}^n \hat\varepsilon_i^2-\left(\frac{1}{n}\sum_{i=1}^n \hat\varepsilon_i\right)^2,
\end{equation*}
and as a result
\begin{align}
\hat\sigma_{Y-f_{\ast}}^2-\sigma_{Y-f_{\ast}}^2
        &=
        \left[ \frac{1}{n}\sum_{i=1}^n \hat\varepsilon_i^2 - \frac{1}{n}\sum_{i=1}^n (\varepsilon_i^*)^2 \right]
        +
        \left[ \frac{1}{n}\sum_{i=1}^n (\varepsilon_i^*)^2 - \EE (\varepsilon_1^*)^2 \right] \notag\\
        &\quad - \left[ \left(\frac{1}{n}\sum_{i=1}^n \hat\varepsilon_i\right)^2 - \left(\frac{1}{n}\sum_{i=1}^n \varepsilon_i^*\right)^2 \right]
        -
        \left[ \left(\frac{1}{n}\sum_{i=1}^n \varepsilon_i^*\right)^2 - (\EE \varepsilon_1^*)^2 \right]. \label{equ:conv_mean_var}
\end{align}
By (\ref{eq:hatbeta_n-err-1}), conditioned on event $\calE_0$, we get 
\begin{equation*}
\big|\hat\eps_i-\eps_i^*\big|=\big|\langle \bff_i, \hat\beta_n-\beta_{\ast}\rangle\big|=O\bigg(\sqrt{\frac{K^4\log n}{n}}\bigg). 
\end{equation*}
By Bernstein inequality, there exists an event $\calE_2$ with $\PP(\calE_2)\geq 1-n^{-10}$, on which the following bounds hold
$$
\bigg|\frac{1}{n}\sum_{i=1}^n (\eps_i^{\ast})^2-\EE (\eps_1^{\ast})^2\bigg|=O\bigg(\sqrt{\frac{K^4\log n}{n}}\bigg)\qquad{\rm and}\qquad \bigg|\frac{1}{n}\sum_{i=1}^n \eps_i^{\ast}-\EE \eps_1^{\ast}\bigg|=O\bigg(\sqrt{\frac{K^2\log n}{n}}\bigg),
$$
where we used the fact $|\eps_i^{\ast}|=O(K)$ under Assumption~\ref{assump:fK}.

These bounds imply that 
\begin{equation*}
\big|\hat\sigma_{Y-f_{\ast}}^2-\sigma_{Y-f_{\ast}}^2\big|=\tilde O_p\bigg(\sqrt{\frac{K^4\log n}{n}}\bigg)\quad{\rm and}\quad \frac{\big|\hat\sigma_{Y-f_{\ast}}-\sigma_{Y-f_{\ast}}\big|}{\hat\sigma_{Y-f_{\ast}}}=\tilde O_p\bigg(\sqrt{\frac{K^4\log n}{n}}\bigg),
\end{equation*}
where the second inequality holds assuming $\sigma_{Y-f_{\ast}}^2>0$ and $n\gg K^4\log n$. 

Observe that 
\begin{align}\label{eq:proof-mean-value-ineq2}
\frac{\sqrt{n}(\hat\theta^{\moe}-\theta_{\ast})}{\hat\sigma_{Y-f_{\ast}}}=\frac{\sqrt{n}(\hat\theta^{\moe}-\theta_{\ast})}{\sigma_{Y-f_{\ast}}}\cdot \bigg(1+\frac{\sigma_{Y-f_{\ast}}-\hat\sigma_{Y-f_{\ast}}}{\hat\sigma_{Y-f_{\ast}}}\bigg).
\end{align}
Applying Bernstein inequality to $Z_{n,f_{\ast}}$ in (\ref{eq:sqrtn-moe-theta}),  we get that there exists an event $\calE_3$ with $\PP(\calE_3)\geq 1-n^{-10}$,  on which the following bound holds
$$
\bigg|\frac{\sqrt{n}(\hat\theta^{\moe}-\theta_{\ast})}{\sigma_{Y-f_{\ast}}} \bigg|=O\big(\sqrt{\log n}\big),
$$
assuming that $n\gg K^4\log^2 n$ and $N\gg n\log n$.  Together with (\ref{eq:proof-mean-value-ineq2}),  we get 
$$
\frac{\sqrt{n}(\hat\theta^{\moe}-\theta_{\ast})}{\hat\sigma_{Y-f_{\ast}}}=\frac{\sqrt{n}(\hat\theta^{\moe}-\theta_{\ast})}{\sigma_{Y-f_{\ast}}}+\tilde O_p\bigg(\frac{K^2\log n}{\sqrt{n}}\bigg). 
$$
Plugging into (\ref{eq:mean-moe-berry}),  we conclude that 
\begin{align}\label{eq:mean-moe-berry-1}
\sup_{t\in\RR}\bigg|\PP\bigg(\frac{\sqrt{n}(\hat\theta^{\moe}-\theta_{\ast})}{\hat \sigma_{Y-f_{\ast}}}\leq t\bigg) -\Phi(t)\bigg|=O\bigg(\frac{K^2\log n}{\sqrt{n}}+\sqrt{\frac{n\log n}{N}}\bigg),
\end{align}
implying that 
$$
\PP\Big(\theta_{\ast}\in \Big[\hat\theta^{\moe}-z_{\alpha/2}\frac{\hat\sigma_{Y-f_{\ast}}}{\sqrt{n}},\  \hat\theta^{\moe}+z_{\alpha/2}\frac{\hat\sigma_{Y-f_{\ast}}}{\sqrt{n}}\Big]\Big)=1-\alpha+O\bigg(\frac{K^2\log n}{\sqrt{n}}+\sqrt{\frac{n\log n}{N}}\bigg),
$$
which concludes the proof.

\subsection{Proof of Theorem~\ref{thm:mean-value+}}
From (\ref{eq:moe-ppi-diff}) and (\ref{eq:ppi-fstar-rep}) in the Proof of Theorem~\ref{thm:mean_estimation},  we get 
\begin{align*}
\sqrt{n}\big(\hat\theta^{\moe}-\theta_{\ast}\big)=-\frac{1}{\sqrt{n}}\sum_{i=1}^n& \Big(\big(f_{\ast}(X_i)-Y_i\big)-\EE\big(f_{\ast}(X)-Y\big)\Big)+\frac{\sqrt{n}}{N}\sum_{i=1}^N\big(f(\tilde X_i)-\EE f(\tilde X)\big)\\
&+\tilde O_p\Big(\frac{K^2\log n}{\sqrt{n}}\Big).
\end{align*}
Applying the Berry-Esseen bound to both $Z_{n,f_{\ast}}$ and $\tilde Z_N$,  we get 
\begin{align*}
\sup_t\bigg|F_{\frac{Z_{n,f_{\ast}}}{\sigma_{Y-f_{\ast}}}}(t) -\Phi(t) \bigg|=O\bigg(\frac{1}{\sqrt{n}}\bigg)\qquad {\rm and}\qquad
\sup_t\bigg|F_{\frac{\tilde Z_{N}}{\sqrt{n/N}\sigma_{f_{\ast}}}}(t)-\Phi(t) \bigg|=O\bigg(\frac{1}{\sqrt{N}}\bigg),
\end{align*}
where $\sigma_{f_{\ast}}^2:=\var\big(f_{\ast}(X)\big)$ and $F_{Y}(\cdot)$ denotes the c.d.f.  of the random variable $Y$.    Moreover,  by the convergence rate for the sum of independent non-identically distributed random variables \citep{petrov2012sums},  we get 
\begin{align*}
\sup_t \bigg|\PP\bigg(\frac{\sqrt{n}(\hat\theta^{\moe}-\theta_{\ast})}{\sqrt{\sigma^2_{Y-f_{\ast}}+(n/N)\sigma^2_{f_{\ast}}}}\leq t\bigg)-\Phi(t) \bigg|=O\bigg(\frac{K^2\log n}{\sqrt{n}}+\frac{1}{\sqrt{N}}\bigg).
\end{align*}
The rest of the proof follows the same as the proof of Theorem~\ref{thm:mean_estimation}.

\subsection{Proof of Lemma~\ref{lem:quantile-est-beta}}

Recall that $\beta_\ast=\beta_\ast(\theta_\ast)$ and $\hat\beta_n=\hat\beta_n(\theta_\ast)$. The local strongly convexity and the separation condition ({\it A2})  in Assumption~\ref{assump:quantile} ensures that $\beta_{\ast}$ is the unique minimizer of $Q(\theta_\ast, \beta)$.

\noindent{\it Step 1: $\hat\beta_n$ is close to $\beta_{\ast}$.}

By Bernstein inequality, for a given $\beta\in\calB$, there exists an event $\calE_0$ with $\PP(\calE_0)\geq 1-n^{-10}$, on which
\begin{align*}
    \max\bigg\{&\bigg|\frac1n\sum_{i=1}^n \Big(S_h(\theta_\ast-Y_i)-S_h(\theta_\ast-\bff_i^\top\beta)\Big)^2-\EE\big[\big(S_h(\theta_\ast-Y)-S_h(\theta_\ast-\bff^\top \beta)\big)^2\big]\bigg|,\\
    &\bigg|\frac1n\sum_{i=1}^n \Big(S_h(\theta_\ast-Y_i)-S_h(\theta_\ast-\bff_i^\top\beta)\Big)-\EE\big[S_h(\theta_\ast-Y)-S_h(\theta_\ast-\bff^\top \beta)\big]\bigg|\Bigg\}\leq C_0\sqrt{\frac{\log n }{n}}.
\end{align*}
Moreover, due to the Lipschitz continuity of $S_h(\theta_{\ast}-\bff^{\top}\beta)$ with respect to $\beta$, we can apply Talagrand's concentration inequality and Dudley's entropy bound as in the proof of Lemma~\ref{lem:m-est-beta} to bound the following empirical process (see a sharper bound in {\it Step~2})
\begin{align*}
    \sup_{\beta\in\calB}|Q_n(\theta_\ast,\beta)-Q(\theta_\ast,\beta)|\leq C_1\sqrt{\frac{K+\log n }{n}},
\end{align*} 
which holds with probability at least $1-n^{-10}$. We denote this event by $\tilde \calE_0$. 

On the event $\tilde\calE_0$, we have 
\begin{align*}
    &\inf_{\beta\in\calB: \|\beta-\beta_{\ast}\|\geq \delta_0} \big\{Q_n(\theta_\ast,\beta)-Q_n(\theta_\ast,\beta_\ast)\big\}\\
    \geq& 
    \inf_{\beta\in\calB: \|\beta-\beta_{\ast}\|\geq \delta_0} \big\{Q(\theta_\ast,\beta)-Q(\theta_\ast,\beta_\ast)\big\} - 2\sup_{\beta\in\calB}|Q_n(\theta_\ast,\beta)-Q(\theta_\ast,\beta)|\geq \eta_0-C_1\sqrt{\frac{K+\log n}{n}},
\end{align*}
which is positive as long as $n\gtrsim \eta_0^{-2}(K+\log n)$. Since $Q_n(\theta_{\ast}, \hat\beta_n)\leq Q_n(\theta_{\ast}, \beta_{\ast})$ by definition, we conclude that $\|\hat\beta_n-\beta_{\ast}\|< \delta_0$ on event $\tilde\calE_0$, according to condition ({\it A2}) of Assumption~\ref{assump:quantile}.

\noindent {\it Step 2: upper bound for $\big|Q(\theta_\ast,\hat\beta_n)-Q(\theta_\ast,\beta_\ast) \big|$}. 

By definition, $Q_{n}(\theta_{\ast}, \hat\beta_n)\leq Q_n(\theta_{\ast}, \beta_{\ast})$ implying that
\begin{align*}
&Q(\theta_{\ast}, \hat\beta_n)-Q(\theta_{\ast}, \beta_{\ast})\leq \big|\big(Q_n(\theta_{\ast}, \hat\beta_n)-Q_n(\theta_{\ast},\beta_{\ast})\big)-\big(Q(\theta_{\ast},\hat\beta_n)-Q(\theta_{\ast},\beta_{\ast})\big) \big|\\
\leq\,& 2\underbrace{\Big|(\PP_n-\PP)\Big(S_h(\theta_{\ast}-Y)\big[S_h(\theta_{\ast}-\bff^{\top}\hat\beta_n)-S_h(\theta_{\ast}-\bff^{\top}\beta_{\ast})\big]\Big) \Big|}_{\calJ_1}\\
&+\underbrace{\Big|(\PP_n-\PP)\Big(S_h^2(\theta_{\ast}-\bff^{\top}\hat\beta_n)-S_h^2(\theta_{\ast}-\bff^{\top}\beta_{\ast})\Big) \Big|}_{\calJ_2}\\
&+\bigg|\Big(\PP_n\big(S_h(\theta_{\ast}-Y)-S_h(\theta_{\ast}-\bff^\top \hat\beta_n)\big)\Big)^2-\Big(\PP_n\big(S_h(\theta_{\ast}-Y)-S_h(\theta_{\ast}-\bff^\top \beta_{\ast})\big)\Big)^2\\
&-\Big(\PP\big(S_h(\theta_{\ast}-Y)-S_h(\theta_{\ast}-\bff^\top \hat\beta_n)\big)\Big)^2+\Big(\PP\big(S_h(\theta_{\ast}-Y)-S_h(\theta_{\ast}-\bff^\top \beta_{\ast})\big)\Big)^2\bigg|,
\end{align*}
where we denote the third term on RHS by $\calJ_3$. 
Note that the above LHS is lower bounded by $0.5h^{-1}\|\hat\beta_n-\beta_{\ast}\|^2\cdot \inf_{\beta\in\calB_{\ast}(\delta_0)}\lambda_{\min}\big(h\nabla^2 Q(\theta_{\ast}, \beta)\big)$,  on the event $\tilde\calE_0$, under the condition ({\it A2}) of Assumption~\ref{assump:quantile}. 

All the three terms $\calJ_1, \calJ_2, \calJ_3$ can be bounded in a similar fashion. Take the term $\calJ_1$ for example.  For any $\delta\in (n^{-1}, \delta_0)$, we study the supremum of an empirical processed indexed by $\beta\in\calB_{\ast}(\delta)$:
$$
\gamma_n(\delta):=\underset{\beta\in\calB_{\ast}(\delta)}{\sup} \Big| (\PP_n-\PP)\Big(S_h(\theta_{\ast}-Y)\big[S_h(\theta_{\ast}-\bff^{\top}\beta)-S_h(\theta_{\ast}-\bff^{\top}\beta_{\ast})\big]\Big)\Big|.
$$
Then, $|\calJ_1|\leq \gamma_n\big(\|\hat\beta_n-\beta_{\ast}\|\big)$. It suffices to develop an upper bound of $\gamma_n(\delta)$ uniformly for all $\delta\in(n^{-1}, \delta_0)$. 

Fix a $\delta\in(n^{-1}, \delta_0)$. Under Assumption~\ref{assump:quantile}, we have 
$$
\sup_{\beta\in\calB_{\ast}(\delta)}\big|S_h(\theta_{\ast}-Y)\big[S_h(\theta_{\ast}-\bff^{\top}\beta)-S_h(\theta_{\ast}-\bff^{\top}\beta_{\ast})\big]\big|\leq h^{-1}U\sup_{\beta\in\calB_{\ast}(\delta)}\|\beta-\beta_{\ast}\|\leq h^{-1}U\delta.
$$

Moreover, for any $\beta\in\calB_{\ast}(\delta)$, we have
\begin{align*}
    &\,\EE S_h^2(\theta_{\ast}-Y)\big[S_h(\theta_{\ast}-\bff^{\top}\beta)-S_h(\theta_{\ast}-\bff^{\top}\beta_{\ast})\big]^2\\
    \leq&\,\EE\big[S_h(\theta_{\ast}-\bff^{\top}\beta)-S_h(\theta_{\ast}-\bff^{\top}\beta_{\ast})\big]^2&(S_h(t)\leq 1)\\
    =&\,\EE\bigg[\int_0^1h^{-1}S'\Big(\frac{\theta_\ast-\bff^\top\beta_t}{h}\Big)\bff^\top(\beta-\beta_\ast)dt\bigg]^2&(\text{Fundamental theorem of calculus})\\
    \leq&\,\EE\bigg[\int_0^1\frac{\|\bff\|\,\|\beta-\beta_\ast\|}{h}S'\Big(\frac{\theta_\ast-\bff^\top\beta_t}{h}\Big)dt\bigg]^2\\
    \leq&\,\EE\int_0^1\frac{\|\bff\|^2\,\|\beta-\beta_\ast\|^2}{h^2}\bigg[S'\Big(\frac{\theta_\ast-\bff^\top\beta_t}{h}\Big)\bigg]^2dt&(\text{Cauchy-Schwarz inequality})\\
    = &\,\frac{U^2\,\|\beta-\beta_\ast\|^2}{h^2}\int_0^1\EE\Big[S'\Big(\frac{\theta_\ast-\bff^\top\beta_t}{h}\Big)\Big]^2dt &(\text{Fubini's theorem})\\
    \leq &\,\frac{U^2\|\beta-\beta_\ast\|^2}{h^2}\int_0^1\EE  S'\Big(\frac{\theta_\ast-\bff^\top\beta_t}{h}\Big)dt&(S'(t)<1)
\end{align*}
where $\beta_t:=\beta_\ast + t(\beta-\beta_\ast)$ and $S(t):=\big(1+\exp(-t)\big)^{-1}$. It's sufficient to control $\EE  S'\Big(\frac{\theta-\bff^\top\beta}{h}\Big)$ for all $\beta\in\calB_\ast(\delta)$.

By Assumption~\ref{assump:quantile}, $Z_{\beta_t}:=\bff^\top\beta_t$ has a bounded density function $p_\beta(z)$ for all $\beta\in\calB_\ast(\delta)$. Thus, 
\begin{align*}
    \EE  S'\bigg(\frac{\theta_\ast-\bff^\top\beta_t}{h}\bigg) &=\int S'\bigg(\frac{\theta_\ast-\bff^\top\beta_t}{h}\bigg)p_{\beta_t}(z)dz\\
    &=h\int S'(u)p_{\beta_t}(\theta_\ast-hu)du\leq C_0 h \int S'(u)du\leq C_2 h,
\end{align*}
where $C_0$ is the upper bound of the density in Assumption~\ref{assump:quantile}.

Consequently, we have $\EE S_h^2(\theta_{\ast}-Y)\big[S_h(\theta_{\ast}-\bff^{\top}\beta)-S_h(\theta_{\ast}-\bff^{\top}\beta_{\ast})\big]^2\lesssim h^{-1}U^2\|\beta-\beta_\ast\|^2$. This indicates that there exists an constant $C_1>0$ such that
\begin{align}\label{equ:quantile_var_bound_J1}
    \var\Big(S_h(\theta_\ast-Y)\big[S_h(\theta_\ast-\bff^\top\beta)-S_h(\theta_\ast-\bff^\top\beta_\ast)\big]\Big)\leq \frac{C_1U^2}{h}\|\beta-\beta_\ast\|^2,\quad \forall\beta\in\calB_\ast(\delta).
\end{align}

Applying Bousquet's version of Talagrand's concentration inequality \citep{bousquet2002bennett},  with probability at least $1-e^{-t}$ for all $t>0$,  
$$
\gamma_n(\delta)\leq 2\EE \gamma_n(\delta)+2U\delta\bigg(\sqrt{\frac{t}{nh}}+\frac{t}{nh}\bigg).  
$$
Similarly, we need to bound $\EE\gamma_n(\delta)$. By the symmerization inequality \citep{koltchinskii2011oracle}, 
\begin{align*}
    \EE \gamma_n(\delta) \leq 2\EE\sup_{\beta\in\calB_\ast(\delta)}\Big|\frac1n \sum_{i=1}^n\varepsilon_i\big(S_h(\theta_\ast-Y_i)\big[S_h(\theta_\ast-\bff_i^\top\beta)-S_h(\theta_\ast-\bff_i^\top\beta_\ast)\big]\big)\Big|,
\end{align*}
where $\varepsilon_1, ..,\varepsilon_n$ are i.i.d. Rademacher random variables. Denote the function class $\calG_\ast(\delta) := \big\{S_h(\theta_\ast-Y)\big[S_h(\theta_\ast-\bff^\top\beta)-S_h(\theta_\ast-\bff^\top\beta_\ast)\big]:\beta\in\calB_\ast(\delta)\big\}$. Conditioned on $(X_1,Y_1),\dots, (X_n,Y_n)$, and for $g=S_h(\theta_\ast-Y)\big[S_h(\theta_\ast-\bff^\top\beta)-S_h(\theta_\ast-\bff^\top\beta_\ast)\big]\in\calG_\ast(\delta)$, we denote $g(X_i,Y_i)=S_h(\theta_\ast-Y_i)\big[S_h(\theta_\ast-\bff_i^\top\beta)-S_h(\theta_\ast-\bff^\top_i\beta_\ast)\big]$ and define the distance $L_2(\PP_n)$ in $\calG_\ast(\delta)$ by
$$
\|g_1-g_2\|^2_{L_2}:=\frac1n\sum_{i=1}^n\big(g_1(X_i,Y_i)-g_2(X_i,Y_i)\big)^2,\quad \forall g_1, g_2\in \calG_\ast(\delta).
$$
By Dudley's entropy bound \citep[Theorem 3.11]{koltchinskii2011oracle}, we have
$$
\EE\gamma_n(\delta)\leq 2\EE\sup_{g\in\calG_\ast(\delta)}\Big|\frac1n\sum_{i=1}^n\varepsilon_i g(X_i,Y_i)\Big|\leq \frac{C_2}{\sqrt n}\EE\int_0^{2\sigma_n}\sqrt{\log N\big(\calG_\ast(\delta);L_2(\PP_n);\epsilon\big)}d\epsilon,
$$
where $C_2>0$ is an absolute constant and $\sigma^2_n:=\sup_{g\in\calG_\ast(\delta)}\PP_ng^2\lesssim h^{-2}U^2\delta^2$. Note that $N(\calF;d;\epsilon)$ represents the $\epsilon$-covering number of a set $\calF$ under the distance $d(\cdot,\cdot)$. 

If $g_1=S_h(\theta_\ast-Y)\big[S_h(\theta_\ast-\bff^\top\beta_1)-S_h(\theta_\ast-\bff^\top\beta_\ast)\big]$ and $g_2=S_h(\theta_\ast-Y)\big[S_h(\theta_\ast-\bff^\top\beta_2)-S_h(\theta_\ast-\bff^\top\beta_\ast)\big]$, we have an constant $C_3>0$ such that  $\sigma_n^2\leq C_3h^{-2}U^2\delta^2$ with probability at least $1-n^{-11}$, and
$$
\|g_1-g_2\|_{L_2(\PP_n)}^2=\frac1n\sum_{i=1}^nS_h^2(\theta_\ast-Y_i)\big[S_h(\theta_\ast-\bff_i^\top\beta_1)-S_h(\theta_\ast-\bff_i^\top\beta_2)\big]^2\leq \frac{C_3 U^2}{h^2}\|\beta_1-\beta_2\|^2,
$$
implying that 
$$
N\big(\calG_\ast(\delta);L_2(\PP_n);\epsilon\big)\leq N\big(\calB_\ast(\delta);\|\cdot\|,\frac{h\epsilon}{\sqrt{C_3}U}\big)\leq \Big(\frac{3\sqrt{C_3}\delta U}{h\varepsilon }\Big)^K,
$$
where the last inequality is due to the fact that $\calB_\ast(\delta)\subset\RR^K$ has a diameter at most $2\delta$.

Plugging them into the Dudley's entropy bound, we get
\begin{align*}
    \EE\gamma_n(\delta) &\leq \frac{C_2}{\sqrt n}\EE\int_0^{2\sqrt{C_3}\cdot U\delta/h}\sqrt{K}\sqrt{\log \frac{3\sqrt{C_3}\delta U}{h\epsilon}}d\epsilon\\
    &=\frac{2C_2\sqrt{KC_3}U\delta}{h\sqrt n}\int_0^1\sqrt{\log\Big(\frac{3}{2x}\Big)}dx\\
    &\lesssim \sqrt{\frac{K}{n}}\frac{U\delta}{h}.
\end{align*}
Therefore, there exists an constant $C_2>0$, with probability at least $1-e^{-t}$, 
$$
\gamma_n(\delta)\leq \frac{C_{2}U\delta}{h}\Big(\sqrt{\frac{K}{n}}+\sqrt{\frac tn}+\frac tn\Big).
$$
By setting $t=C\log(n)$ for a sufficiently large constant $C>0$, we get with probability at least $1-n^{-11}$,
$$
\gamma_n(\delta)\leq \frac{CU\delta}{h}\sqrt{\frac{K+\log n}{n}},
$$
which holds for any fixed $\delta>0$. Following the discretization argument as in the proof of Lemma~\ref{lem:gamma_n-bd}, we can extend it to all $\delta\in(n^{-1}, \delta_0)$ and with a union bound of probabilities.

Consequently, there exists an event $\calE_1$ with $\PP(\calE_1)\geq 1-n^{-10}$, on which
\begin{align}\label{eq:quantile_Q_J1}
    \calJ_1\leq \gamma_n(\|\hat\beta_n-\beta_\ast\|)\leq \frac{C_{2}U}{h}\sqrt{\frac {K+\log n}{n}}\|\hat\beta_n-\beta_\ast\|.
\end{align}

Similarly, we bound $\sup_{\beta\in\calB_\ast(\delta)} \big|S_h(\theta_\ast-\bff^\top \beta)-S_h(\theta_\ast-\bff^\top \beta_\ast)\big|$. First,
\begin{align}
    \sup_{\beta\in\calB_\ast(\delta)} \big|S_h(\theta_\ast-\bff^\top \beta)-S_h(\theta_\ast-\bff^\top \beta_\ast)\big|
    \leq \,2h^{-1}U\sup_{\beta\in\calB_\ast(\delta)}\|\beta-\beta_\ast\|\leq 2h^{-1}U\delta,\label{eq:diff_S_bound}
\end{align}
and
\begin{align}
    \sup_{\beta\in\calB_\ast(\delta)} \var\big(S_h(\theta_\ast-\bff^\top \beta)-S_h(\theta_\ast-\bff^\top \beta_\ast)\big)
    \leq &\,\sup_{\beta\in\calB_\ast(\delta)} \EE\big[S_h(\theta_\ast-\bff^\top \beta)-S_h(\theta_\ast-\bff^\top \beta_\ast)\big]^2\notag\\
    \leq &\,h^{-1}U^2\sup_{\beta\in\calB_\ast(\delta)}\|\beta-\beta_\ast\|^2\lesssim h^{-1}U^2\delta^2.\label{eq:diff_S2_bound}
\end{align}
These bounds indicate that
\begin{align*}
    &\sup_{\beta\in\calB_\ast(\delta)} \Big|(\PP_n-\PP)\big(S_h(\theta_\ast-\bff^\top \beta)-S_h(\theta_\ast-\bff^\top \beta_\ast)\big)\Big|\\
    \leq\, &\EE\sup_{\beta\in\calB_\ast(\delta)} \Big|(\PP_n-\PP)\big(S_h(\theta_\ast-\bff^\top \beta)-S_h(\theta_\ast-\bff^\top \beta_\ast)\big)\Big|+2U\delta\Big(\sqrt{\frac{t}{nh}}+\frac{t}{nh}\Big).
\end{align*}
Then, $\sup_{\beta\in\calB_\ast(\delta)}\PP_n\big(S_h(\theta_\ast-\bff^\top \beta)-S_h(\theta_\ast-\bff^\top \beta_\ast)\big)^2$ shares the same order with $\sigma_n^2$ in the proof of $\calJ_1$, that is $h^{-2}U^2\delta^2$. The same procedure concludes that there exist a constant $C_{2}>0$, with probability that at least $1-n^{-10}$, 
\begin{align}\label{eq:quantile_Q_J2_sub}
    \sup_{\beta\in\calB_\ast(\|\hat\beta_n-\beta_\ast\|)} \Big|(\PP_n-\PP)\big(S_h(\theta_\ast-\bff^\top \beta)-S_h(\theta_\ast-\bff^\top \beta_\ast)\big)\Big|\leq \frac{C_{2}U}{h}\sqrt{\frac{K+\log n}{n}}\|\hat\beta_n-\beta_\ast\|.
\end{align}
While $\calJ_2\leq 2\Big|(\PP_n-\PP)\big(S_h(\theta_\ast-\bff^\top \hat\beta_n)-S_h(\theta_\ast-\bff^\top \beta_\ast)\big)\Big|$, 
\begin{align}\label{eq:quantile_Q_J2}
    \calJ_2\leq \frac{2C_{2}U}{h}\sqrt{\frac{K+\log n}{n}}\|\hat\beta_n-\beta_\ast\|.
\end{align}


For $\calJ_3$, we rewrite it as
\begin{align*}
    &\bigg|\Big(\PP_n\big(S_h(\theta_{\ast}-Y)-S_h(\theta_{\ast}-\bff^\top \hat\beta_n)\big)\Big)^2-\Big(\PP_n\big(S_h(\theta_{\ast}-Y)-S_h(\theta_{\ast}-\bff^\top \beta_{\ast})\big)\Big)^2\\
    &-\Big(\PP\big(S_h(\theta_{\ast}-Y)-S_h(\theta_{\ast}-\bff^\top \hat\beta_n)\big)\Big)^2+\Big(\PP\big(S_h(\theta_{\ast}-Y)-S_h(\theta_{\ast}-\bff^\top \beta_{\ast})\big)\Big)^2\bigg|\\
    =\, &\bigg|(\PP_n-\PP)\big(S_h(\theta_{\ast}-Y)-S_h(\theta_{\ast}-\bff^\top \hat\beta_n)\big) \cdot (\PP_n+\PP)\big(S_h(\theta_{\ast}-Y)-S_h(\theta_{\ast}-\bff^\top \hat\beta_n)\big)\\
    &-(\PP_n-\PP)\big(S_h(\theta_{\ast}-Y)-S_h(\theta_{\ast}-\bff^\top \beta_{\ast})\big)\cdot (\PP_n+\PP)\big(S_h(\theta_{\ast}-Y)-S_h(\theta_{\ast}-\bff^\top \beta_{\ast})\big)\bigg|\\
    \leq  \, &\Big|(\PP_n-\PP)\big(2S_h(\theta_\ast-Y)-S_h(\theta_{\ast}-\bff^\top \hat\beta_n)-S_h(\theta_{\ast}-\bff^\top \beta_\ast)\big)\Big|\Big|\PP_n\big(S_h(\theta_{\ast}-\bff^\top \hat\beta_n)-S_h(\theta_{\ast}-\bff^\top \beta_\ast)\big)\Big|\\
    &+\Big|\PP\big(2S_h(\theta_\ast-Y)-S_h(\theta_{\ast}-\bff^\top \hat\beta_n)-S_h(\theta_{\ast}-\bff^\top \beta_\ast)\big)\Big|\Big|(\PP_n-\PP)\big(S_h(\theta_{\ast}-\bff^\top \hat\beta_n)-S_h(\theta_{\ast}-\bff^\top \beta_\ast)\big)\Big|.
\end{align*}

Since $\big\{2S_h(\theta_\ast-Y)-S_h(\theta_{\ast}-\bff^\top \beta)-S_h(\theta_{\ast}-\bff^\top \beta_\ast):\beta\in\calB_\ast(\delta_0)\big\}$ is uniformly bounded and indexed by a finite-dimensional parameter, a standard empirical process bound yields 
\begin{align}
    \sup_{\beta\in\calB_\ast(\delta_0)}\Big|(\PP_n-\PP)\big(2S_h(\theta_\ast-Y)-S_h(\theta_{\ast}-\bff^\top \beta)-S_h(\theta_{\ast}-\bff^\top \beta_\ast)\big)\Big|&\leq C_4\sqrt{\frac{K+\log n}n},\label{eq:quantile_Q_J3_sub1}    
\end{align}
and 
\begin{align}
    \Big|\PP\big(2S_h(\theta_\ast-Y)-S_h(\theta_{\ast}-\bff^\top \beta)-S_h(\theta_{\ast}-\bff^\top \beta_\ast)\big)\Big|&\leq 2,\label{eq:quantile_Q_J3_sub2}
\end{align}
where $C_4>0$ is a constant. Here, $\PP(\cdot)$ represents the expectation operator $\EE(\cdot)$.

With the bound 
\begin{align}
    &\Big|\PP\big(S_h(\theta_{\ast}-\bff^\top \hat\beta_n)-S_h(\theta_{\ast}-\bff^\top \beta_\ast)\big)\Big|\notag\\
    \leq \, &\sup_{\beta\in\calB_\ast(\|\hat\beta_n-\beta_\ast\|)}\Big|\EE\int_0^1h^{-1}S'\Big(\frac{\theta_\ast-\bff^\top\beta_t}{h}\Big)\bff^\top(\beta-\beta_\ast)dt\Big|\notag\\
    \leq &\, \sup_{\beta\in\calB_\ast(\|\hat\beta_n-\beta_\ast\|)}\frac{U\|\beta-\beta_\ast\|}{h}\Big|\int_0^1\EE S'\Big(\frac{\theta_\ast-\bff^\top\beta_t}{h}\Big)dt\Big|\notag\\
    \leq &\, \sup_{\beta\in\calB_\ast(\|\hat\beta_n-\beta_\ast\|)}\frac{U\|\beta-\beta_\ast\|}{h}\cdot C_0h=UC_0\|\hat\beta_n-\beta_\ast\|,\label{eq:quantile_Q_J3_sub3}
\end{align}
there exists a constant $C_5>0$ such that
\begin{align}
    &\Big|\PP_n\big(S_h(\theta_{\ast}-\bff^\top \hat\beta_n)-S_h(\theta_{\ast}-\bff^\top \beta_\ast)\big)\Big|\notag\\
    \leq\,&\Big|(\PP_n-\PP)\big(S_h(\theta_{\ast}-\bff^\top \hat\beta_n)-S_h(\theta_{\ast}-\bff^\top \beta_\ast)\big)\Big|+\Big|\PP\big(S_h(\theta_{\ast}-\bff^\top \hat\beta_n)-S_h(\theta_{\ast}-\bff^\top \beta_\ast)\big)\Big|\notag\\
    \leq \,&\frac{C_{2}U}{h}\sqrt{\frac {K+\log n}{n}}\cdot \|\hat\beta_n-\beta_\ast\|+UC_0\|\hat\beta_n-\beta_\ast\|,\label{eq:quantile_Q_J3_sub4}
\end{align}
holding with probability at least $1-n^{-11}$.

Combining (\ref{eq:quantile_Q_J2_sub}), (\ref{eq:quantile_Q_J3_sub1}), (\ref{eq:quantile_Q_J3_sub2}), and (\ref{eq:quantile_Q_J3_sub4}), there exist a constant $C_{\calJ_3}>0$ and an event $\calE_3$ with probability $\PP(\calE_3)\geq 1-3n^{-11}\geq 1-n^{-10}$, on which $\calJ_3$ has the bound
\begin{align}\label{eq:quantile_Q_J3}
    \calJ_3 &\leq C_4\sqrt{\frac{K+\log n}{n}}\cdot \bigg(\frac{C_{2}U}{h}\sqrt{\frac {K+\log n}{n}}\cdot \|\hat\beta_n-\beta_\ast\|+UC_0\|\hat\beta_n-\beta_\ast\|\bigg)+2\cdot \frac{C_{2}U}{h}\sqrt{\frac {K+\log n}{n}}\|\hat\beta_n-\beta_\ast\|\notag\\
    &\leq \frac{C_{_3}U}{h}\sqrt{\frac{K+\log n}{n}}\|\hat\beta_n-\beta_\ast\|,
\end{align}
for $n\gtrsim K+\log n$.

Combining (\ref{eq:quantile_Q_J1}), (\ref{eq:quantile_Q_J2}) and (\ref{eq:quantile_Q_J3}), conditioned on $\calE_1\cap\calE_2\cap\calE_3$ with probability at least $1-3n^{-10}$, there exist a constant $C$ such that
\begin{align}\label{eq:quantile_Q_diff_bound}
    Q(\theta_\ast,\hat\beta_n) - Q(\theta_\ast,\beta_\ast)\leq \frac{C}{h}\sqrt{\frac{K+\log n}{n}}\|\hat\beta_n-\beta_\ast\|,
\end{align}
since $n\gg \log n$.

\noindent {\it Step 3: upper bound for $\|\hat\beta_n-\beta_\ast\|$}. 

Combining with the locally strong convexity of $Q(\theta_\ast,\beta)$ in Assumption~\ref{assump:quantile}, we have
\begin{align*}
    &\frac1{2h}c_0\|\hat\beta_n-\beta_\ast\|^2\leq 
    \frac1{2h}\|\hat\beta_n-\beta_\ast\|^2\inf_{\beta\in\calB(\delta_0)}\lambda_{\min}\big(h\nabla^2Q(\theta_\ast,\beta)\big)\\
    \leq \,&Q(\theta_\ast,\hat\beta_n)-Q(\theta_\ast,\beta_\ast)\leq \frac{C}{h}\sqrt{\frac{K+\log n}{n}}\|\hat\beta_n-\beta_\ast\|.
\end{align*}

Overall, there exist a constant $C$ and an event $\calE$ with $\PP(\calE)\geq 1-n^{-9}$, on which
\begin{align*}
    \|\hat\beta_n-\beta_\ast\|\leq C\sqrt{\frac{K+\log n}{n}}.
\end{align*}

This completes the proof.

\subsection{Proof of Lemma~\ref{lem:quantile-moe-bias}}
Observe that
\begin{align*}
    &\,\Big|\EE \hat m^\moe_{\hat\beta_n}(\theta_{\ast})\Big|\\
    =& \, \bigg|\frac{1}{n}\sum_{i=1}^n\EE\big[S_h(\theta_\ast-Y_i)-S_h(\theta_\ast-\bff_i^\top \hat\beta_n)\big] +\frac{1}{N}\sum_{i=1}^N\EE S_h(\theta_\ast-\tilde{\bff}_i^\top\hat \beta_n)-q\bigg|\\
    \leq & \, \big|\underbrace{\EE S_h(\theta_\ast-Y)- \PP(Y\leq\theta_\ast)}_{\calJ_1}\big|\\
    &+ \bigg |\underbrace{\EE \frac1n\sum_{i=1}^n\big[S_h(\theta_\ast-\bff_i^\top \hat\beta_n)- S_h(\theta_\ast-\bff_i^\top \beta_\ast)\big]-\EE \frac{1}{N}\sum_{i=1}^N\big[ S_h(\theta_\ast-\tilde\bff_i^\top \hat\beta_n)- S_h(\theta_\ast-\tilde \bff_i^\top \beta_\ast)\big ]}_{\calJ_2}\bigg |.
\end{align*}

For the first term, by the smoothness of the probability density $f_Y$ in a neighborhood of $\theta_\ast$, we fix a small $\delta>0$ and write
\begin{align*}
    \mathcal J_1 = \int_{\RR} \big[ S_h(\theta_*-y)-\II\{y\le \theta_*\} \big]\,dF_Y(y)
    =:T_{1n}+T_{2n}+T_{3n},
\end{align*}
where
\begin{align*}
    T_{1n} &:= \int_{(-\infty,\theta_\ast-\delta]} \big[ S_h({\theta_\ast-y})-1 \big]dF_Y(y),\\
    T_{2n} &:= \int_{(\theta_\ast-\delta,\theta_\ast+\delta]} \big[ S_h(\theta_\ast-y)-\II\{y\leq \theta_\ast\} \big]dF_Y(y),\\
    T_{3n} &:= \int_{(\theta_\ast+\delta,\infty)} S_h({\theta_\ast-y})dF_Y(y).
\end{align*}

For the local term $T_{2n}$, using the change of variables $y=\theta_\ast+hu$, we obtain 
\begin{align*}
    T_{2n} = h \int_{-\delta/h}^{\delta/h} \psi(u)\,f_Y(\theta_\ast+hu)du, \qquad \psi(u):=S(-u)-\II\{u\leq 0\},
\end{align*}
where $S(u)=\big(1+\exp(-u)\big)^{-1}$. Since $\psi$ is odd, a Taylor expansion of $f_Y(\theta_\ast+hu)$ around $\theta_\ast$ yields
\begin{align*}
    T_{2n}
    =&\, h f_Y(\theta_\ast)\int_{\RR}\psi(u)\,du
    + h^2 f_Y'(\theta_\ast)\int_{\RR}u\psi(u)\,du
    + O\bigg(h^3\int_{\RR}|u|^2|\psi(u)|\,du\bigg)\\
    =&\, h^2 f_Y'(\theta_\ast)\int_{\RR}u\big[S(u)-\II\{u\ge 0\}\big]\,du + o(h^2)
    = O(h^2).
\end{align*}

For the left tail term $T_{1n}$, if $y\le \theta_\ast-\delta$, then
\begin{align*}
    \frac{\theta_*-y}{h}\ge \frac{\delta}{h}.
\end{align*}
Since $S$ is a sigmoid function,
\begin{align*}
    \sup_{y\le \theta_*-\delta} \big| S_h(\theta_*-y)-1 \big| =1-\frac1{1+e^{-\delta/h}}\leq \exp\bigg\{-\frac{\delta}{h}\bigg\},
\end{align*}
and consequently $|T_{1n}|\leq q\exp\{-\delta/h\}$.
Similarly, for the right tail term $T_{3n}$ follows $|T_{3n}|\leq (1-q)\exp\{-\delta/h\}$.

Combining the above bounds, we obtain
\begin{align}\label{eq:quantile-ee-bd1}
    |\calJ_1| \lesssim h^2 + e^{-\delta/h}=O(h^2).
\end{align}

For $\calJ_2$, we write $|\calJ_2|$ as
\begin{align*}
     &\bigg|\EE \frac1n\sum_{i=1}^n\Big[S_h(\theta_\ast-\bff^\top_i\hat\beta_n)-S_h(\theta_\ast-\bff^\top_i\beta_\ast)\Big]-\EE\frac1N\sum_{i=1}^N\Big[S_h(\theta_\ast-\tilde\bff^\top_i\hat\beta_n)-S_h(\theta_\ast-\tilde\bff^\top_i\beta_\ast)\Big]\bigg|\\
    \leq &\EE_{\mathcal L}  \bigg|\frac1n\sum_{i=1}^n\Big[S_h(\theta_\ast-\bff^\top_i\hat\beta_n)-S_h(\theta_\ast-\bff^\top_i\beta_\ast)\Big]-\mathbb E_{\mathcal U}\frac1N\sum_{i=1}^N\Big[S_h(\theta_\ast-\tilde \bff^\top_i\hat\beta_n)-S_h(\theta_\ast-\tilde \bff^\top_i\beta_\ast)\Big]\bigg|\\
    \leq &\EE_{\mathcal L} \sup_{\beta\in\mathcal{B}_{\ast}(\|\hat\beta_n-\beta_\ast\|)} \bigg|\frac1n\sum_{i=1}^n\Big[S_h(\theta_\ast-\bff^\top_i\beta)-S_h(\theta_\ast-\bff^\top_i\beta_\ast)\Big]-\mathbb E_{\mathcal U}\frac1N\sum_{i=1}^N\Big[S_h(\theta_\ast-\tilde \bff^\top_i\beta)-S_h(\theta_\ast-\tilde \bff^\top_i\beta_\ast)\Big]\bigg|\\
    \leq &\EE \sup_{\beta\in\mathcal{B}_{\ast}(\|\hat\beta_n-\beta_\ast\|)} \bigg|(\PP_n-\PP)\Big[S_h(\theta_\ast-\bff^\top_i\beta)-S_h(\theta_\ast-\bff^\top_i\beta_\ast)\Big]\bigg|.
\end{align*}
By \eqref{eq:quantile_Q_J2_sub}, we get
\begin{align}
    |\calJ_2|&\leq \frac{C_{2}U}{h}\sqrt{\frac{K+\log n}{n}}\EE\|\hat\beta_n-\beta_\ast\|\notag\\
    &\leq \frac{C_{2}U}{h}\sqrt{\frac{K+\log n}{n}}\Big[C\sqrt{\frac{K+\log n}{n}}\PP({\calE})+\text{diam}(\calB) \PP({\calE^c})\Big]\notag\\
    &\lesssim \frac{K+\log n}{nh},\label{eq:quantile-ee-bd2}
\end{align}
where the event $\calE$ refers to the conclusion of Lemma~\ref{lem:quantile-est-beta} holds with $\PP(\calE)\geq 1-n^{-10}$.

Together with \eqref{eq:quantile-ee-bd1} and \eqref{eq:quantile-ee-bd2}, the bias satisfies
$$
\Big|\EE\hat m^\moe_{\hat\beta_n}(\theta_\ast)\Big|
=
O\bigg(
h^2+\frac{K+\log n}{nh}
\bigg),
$$
which concludes the proof.

\subsection{Proof of Theorem~\ref{thm:quantile_estimation}}
By Lemma~\ref{lem:quantile-est-beta}, $\hat\beta_n$ is a good estimator of $\beta_\ast$ at $\theta=\theta_\ast$, we then prove that $\hat m^\moe_{\hat\beta_n}(\theta_\ast)$ is also a good estimator of $\hat m^\moe_{\beta_\ast}(\theta_\ast)$.

\begin{align*}
&\hat m^\moe_{\hat\beta_n}(\theta_\ast) - \hat m^\moe_{\beta_\ast}(\theta_\ast) \\
=&\frac1n\sum_{i=1}^n \Big[S_h(\theta_\ast - \bff_i^\top\beta_\ast)-S_h(\theta_\ast - \bff_i^\top\hat\beta_n)\Big]-\frac1N\sum_{i=1}^N \Big[S_h(\theta_\ast - \tilde \bff_i^\top\beta_\ast)-S_h(\theta_\ast - \tilde \bff_i^\top\hat\beta_n)\Big]\\
=&\big(\PP_n-\PP\big)\Big[S_h(\theta_\ast - \bff^\top\beta_\ast)-S_h(\theta_\ast - \bff^\top\hat\beta_n)\Big]-\big(\PP_N-\PP\big)\Big[S_h(\theta_\ast - \bff^\top\beta_\ast)-S_h(\theta_\ast - \bff^\top\hat\beta_n)\Big].
\end{align*}
Then, by \eqref{eq:diff_S_bound}, \eqref{eq:diff_S2_bound}, \eqref{eq:quantile_Q_J2_sub} and Bernstein inequality, we have
\begin{align}
    &\sqrt n\Big|\hat m^\moe_{\hat\beta_n}(\theta_\ast) - \hat m^\moe_{\beta_\ast}(\theta_\ast)\Big|\notag\\
    \leq\, &\sqrt n\sup_{\beta\in\mathcal{B}_{\ast}(\|\hat\beta_n-\beta_\ast\|)}\Bigg|\big(\PP_n-\PP\big)\Big[S_h(\theta_\ast - \bff^\top\beta)-S_h(\theta_\ast - \bff^\top\beta_\ast)\Big]\Bigg|\notag\\
    &+\sqrt n\Bigg|\big(\PP_N-\PP\big)\Big[S_h(\theta_\ast - \bff^\top\hat\beta_n)-S_h(\theta_\ast - \bff^\top\beta_\ast)\Big]\Bigg|\notag\\
    \leq\, &\sqrt{n}\frac{C_2U}{h}\sqrt{\frac{K+\log n}{n}}\|\hat\beta_n-\beta_\ast\|+\tilde O_p\Big(\sqrt{\frac{nU^2\|\hat\beta_n-\beta_\ast\|^2}{Nh}}+\frac{\sqrt{n}U\|\hat\beta_n-\beta_\ast\|}{Nh}\Big)\notag\\
    \leq\, &\tilde O_p\Big(\frac{K+\log n}{\sqrt{n}h}+\sqrt{\frac{K+\log n}{Nh}}\Big),\label{equ:quantile-m-error}
\end{align}

For a fixed $\beta$, let
\begin{align*}
    m_\beta^\moe(\theta_\ast) =\EE \hat m_\beta^\moe(\theta_\ast) = \EE\big[S_h(\theta_{\ast}-Y)\big]-q=O(h^2),
\end{align*}
which is irrelevant to $\beta$.

We then decompose $\sqrt n\,\hat m^\moe_{\hat\beta_n}(\theta_\ast)$:
\begin{align*}
    \sqrt n\,\hat m^\moe_{\hat\beta_n}(\theta_\ast) =& 
    \sqrt n\,{m^\moe_{\beta_\ast }(\theta_\ast)} 
    + \sqrt n\Big[{\hat m^\moe_{\hat\beta_n}(\theta_\ast) - \hat m^\moe_{\beta_\ast}(\theta_\ast)} \Big]
    + \sqrt n\Big[{\hat m^\moe_{\beta_\ast}(\theta_\ast)-m^\moe_{\beta_\ast}(\theta_\ast)}\Big]\\
    =&O(\sqrt{n}h^2)+\tilde O_p\Big(\frac{K+\log n}{\sqrt{n}h}+\sqrt{\frac{K+\log n}{Nh}}\Big)+\sqrt n\Big[{\hat m^\moe_{\beta_\ast}(\theta_\ast)-m^\moe_{\beta_\ast}(\theta_\ast)}\Big].
\end{align*}
It suffices to study the last term. Define
\begin{align*}
    Z_{i}&=S_h({\theta_\ast-Y_i})-S_h({\theta_\ast-\bff_i^\top \beta_\ast})-\big[\EE S_h({\theta_\ast-Y})-\EE S_h({\theta_\ast-\bff^\top \beta_\ast})\big],\\
    \tilde Z_i &= S_h({\theta_\ast-\tilde \bff_i^\top \beta_\ast}) - \EE S_h({\theta_\ast-\bff^\top \beta_\ast}),
\end{align*}
here, we suppress the dependence on $\beta_\ast$ for simplicity.
Then 
\begin{align*}
    \hat m^\moe_{\beta_\ast}(\theta_\ast) - m^\moe_{\beta_\ast}(\theta_\ast) = \frac1n \sum_{i=1}^n Z_i + \frac1N \sum_{i=1}^N \tilde Z_i,
\end{align*}
and $\{Z_i\}_{i=1}^n$ and $\{\tilde Z_i\}_{i=1}^N$ are independent with mean zero. 
Since $S_h(\cdot)\in(0,1)$, we have $|Z_i|\leq 2$ and $|\tilde Z_i|\leq 1$. 
Moreover, 
\begin{align*}
    \var\big(Z_i\big) = \var\big(S_h({\theta_\ast- Y})-S_h({\theta_\ast- \bff^\top \beta_\ast})\big) \quad \text{and} \quad \var\big(\tilde Z_i\big)=\var\big(S_h({\theta_\ast-\tilde \bff^\top \beta_\ast})\big)
\end{align*}
are bounded as well. 

By central limit theorem, we get
\begin{align*}
    \frac1{\sqrt n} \sum_{i=1}^n Z_i
    &\overset{d}{\to} \calN\Big(0, \var\big(S_h({\theta_\ast- Y})-S_h({\theta_\ast- \bff^\top \beta_\ast})\big)\Big).
\end{align*}
By Bernstein inequality, we get
\begin{align*}
    \frac{\sqrt n}{N} \sum_{i=1}^N \tilde Z_i
    =\tilde O_p\Bigg(\sqrt{\frac{n\log n}{N}\cdot \var\big(S_h({\theta_\ast- \bff^\top \beta_\ast})\big)}\Bigg).
\end{align*}
Therefore, we write
\begin{align}\label{eq:ppi-fstar-rep-quantile}
    \sqrt n\Big(\hat m^\moe_{\beta_\ast}(\theta_\ast) - m^\moe_{\beta_\ast}(\theta_\ast)\Big)=Z_{n, f_\ast} + \frac{\sqrt{n}}{N}\sum_{i=1}^N\tilde Z_i,
\end{align}
where $Z_{n, f_\ast}\to_d\calN\Big(0, \var\big(S_h(\theta_\ast- Y)-S_h({\theta_\ast- \bff^\top \beta_\ast})\big)\Big)$ as $n\to \infty$.

Overall, combing (\ref{eq:quantile-ee-bd1}), (\ref{equ:quantile-m-error}) and (\ref{eq:ppi-fstar-rep-quantile}), we have
\begin{align}
     \sqrt n\,\hat m^\moe_{\hat\beta_n}(\theta) &= \tilde O_p\Big(\sqrt nh^2\Big)+\tilde O_p\Big(\frac{K+\log n}{\sqrt{n}h}+\sqrt{\frac{K+\log n}{Nh}}\Big) +Z_{n, f_\ast} + \tilde O_p\Bigg(\sqrt{\frac{n\log n}{N}}\Bigg)\notag\\
     &= Z_{n,f_\ast}+\tilde O_p\Bigg(\sqrt{n}h^2+\frac{K+\log n}{\sqrt {n}h}+\sqrt{\frac{K+\log n}{Nh}}+\sqrt{\frac{n\log n}{N}}\Bigg)\label{eq:sqrtn-moe-quantile-theta}.
\end{align}
By the Berry-Esseen bound and the high probability bound inherited from $\tilde O_p(\cdot)$, we get
\begin{align}\label{eq:quantile-moe-bery}
    \sup_{t\in\RR}\Bigg|\PP\bigg(\frac{\sqrt{n}\, \hat m_{\hat\beta_n}^\moe(\theta_\ast)}{\sqrt{Q(\theta_\ast,\beta_\ast)}}\leq t\bigg)-\Phi(t)\Bigg| =  O\Bigg(\sqrt{n}h^2+\frac{K+\log n}{\sqrt {n}h}+\sqrt{\frac{K+\log n}{Nh}}+\sqrt{\frac{n\log n}{N}}\Bigg),
\end{align}
where $Q(\theta_\ast,\beta_\ast)=\var\big(S_h({\theta_\ast- Y})-S_h({\theta_\ast- \bff^\top \beta_\ast})\big)$. 

Therefore,
\begin{align*}
    \PP\Bigg( \bigg|\frac{\sqrt{n}\, \hat m_{\hat\beta_n}^\moe(\theta_\ast)}{\sqrt{Q(\theta_\ast,\beta_\ast)}}\bigg|\leq z_{\alpha/2}\Bigg) =1-\alpha+ O\Bigg(\sqrt{n}h^2+\frac{K+\log n}{\sqrt {n}h}+\sqrt{\frac{K+\log n}{Nh}}+\sqrt{\frac{n\log n}{N}}\Bigg).
\end{align*}
It remains to bound $\Big|Q_n(\theta_\ast,\hat\beta_n)-Q(\theta_\ast, \beta_\ast)\Big|$. By \eqref{eq:quantile_Q_J2_sub}, \eqref{eq:quantile_Q_diff_bound} and Lemma~\ref{lem:quantile-est-beta},
\begin{align*}
    &\Big|Q_n(\theta_\ast,\hat\beta_n)-Q(\theta_\ast, \beta_\ast)\Big|\\
    \leq \,&\sup_{\beta\in\calB_\ast(\|\hat\beta_n-\beta_\ast\|)}\Big|Q_n(\theta_\ast,\beta)-Q(\theta_\ast, \beta)\Big| + \Big|Q(\theta_\ast,\hat\beta_n)-Q(\theta_\ast, \beta_\ast)\Big|\\
    \leq \,&\Big|Q(\theta_\ast,\hat\beta_n)-Q(\theta_\ast, \beta_\ast)\Big|+\sup_{\beta\in\calB_\ast(\|\hat\beta_n-\beta_\ast\|)}\Big|(\PP_n-\PP)\big(S_h(\theta_\ast-\bff^\top \beta_\ast)-S_h(\theta_\ast-\bff^\top\beta)\big)^2\Big|\\
    &+\sup_{\beta\in\calB_\ast(\|\hat\beta_n-\beta_\ast\|)}\Big|(\PP_n-\PP)\big(S_h(\theta_\ast-\bff^\top \beta_\ast)-S_h(\theta_\ast-\bff^\top\beta)\big)\Big|\cdot \Big|(\PP_n+\PP)\big(S_h(\theta_\ast-\bff^\top \beta_\ast)-S_h(\theta_\ast-\bff^\top\beta)\big)\Big|\\
    \leq \,&\Big|Q(\theta_\ast,\hat\beta_n)-Q(\theta_\ast, \beta_\ast)\Big|+4\sup_{\beta\in\calB_\ast(\|\hat\beta_n-\beta_\ast\|)}\Big|(\PP_n-\PP)\big(S_h(\theta_\ast-\bff^\top \beta_\ast)-S_h(\theta_\ast-\bff^\top\beta)\big)\Big|\\
    \leq\, & \tilde O_p\Big(\frac{K+\log n}{nh}\Big).
\end{align*}

This bound implies
\begin{align*}
    &\frac{\Big|\sqrt{Q_n(\theta_\ast,\hat\beta_n)}-\sqrt{Q(\theta_\ast, \beta_\ast)}\Big|}{\sqrt{Q_n(\theta_\ast,\hat\beta_n)}}=\tilde O_p\Big(\frac{K+\log n}{nh}\Big),
\end{align*}
where the second inequality holds assuming $Q(\theta_\ast, \beta_\ast)>0$ and $nh\gtrsim K+\log n$.

Since 
\begin{align}\label{eq:proof-quantile-value-ineq2}
    \frac{\sqrt{n}\, \hat m_{\hat\beta_n}^\moe(\theta_\ast)}{\sqrt{Q_n(\theta_\ast,\hat\beta_n)}} = \frac{\sqrt{n}\, \hat m_{\hat\beta_n}^\moe(\theta_\ast)}{\sqrt{Q(\theta_\ast,\beta_\ast)}}\cdot \bigg(1+\frac{\sqrt{Q(\theta_\ast,\beta_\ast)}-\sqrt{Q_n(\theta_\ast, \beta_\ast)}}{\sqrt{Q_n(\theta_\ast,\beta_\ast)}}\bigg).
\end{align}
Applying Bernstein inequality to $Z_{n,f_\ast}$ in (\ref{eq:sqrtn-moe-quantile-theta}), we get 
\begin{align*}
    \Bigg|\frac{\sqrt n\, \hat m_{\hat \beta_n}^\moe(\theta_\ast)}{\sqrt{Q(\theta_\ast, \beta_\ast)}}\Bigg| = \tilde O_p(\sqrt{\log n}),
\end{align*}
assuming $n\to\infty$, $nh^2\to\infty$, $nh^4\to0$ and $N\gtrsim n$. Together with (\ref{eq:proof-quantile-value-ineq2}), we get
\begin{align*}
    \frac{\sqrt{n}\, \hat m_{\hat\beta_n}^\moe(\theta_\ast)}{\sqrt{Q_n(\theta_\ast,\hat\beta_n)}} = \frac{\sqrt{n}\, \hat m_{\hat\beta_n}^\moe(\theta_\ast)}{\sqrt{Q(\theta_\ast,\beta_\ast)}} + \tilde O_p\bigg(\frac{\log n}{\sqrt n}+\frac{(K+\log n)\sqrt{\log n}}{nh}\bigg).
\end{align*}
Plugging into (\ref{eq:quantile-moe-bery}), we conclude that
\begin{align}\label{eq:quantile-moe-berry-1}
    \sup_{t\in\RR}\Bigg|\PP\bigg(\frac{\sqrt{n}\, \hat m_{\hat\beta_n}^\moe(\theta_\ast)}{\sqrt{Q_n(\theta_\ast,\hat\beta_n)}}\leq t\bigg)-\Phi(t)\Bigg| = O\Bigg(\sqrt{n}h^2+\frac{K+\log n}{\sqrt {n}h}+\sqrt{\frac{K+\log n}{Nh}}+\sqrt{\frac{n\log n}{N}}\Bigg),
\end{align}
implying that
\begin{align*}
    \PP(\theta_\ast\in\calC^\moe_\alpha)=\PP\Bigg( \frac{\big|\sqrt{n}\, \hat m_{\hat\beta_n}^\moe(\theta_\ast)\big|}{\sqrt{Q_n(\theta_\ast,\hat\beta_n)}}\leq z_{\alpha/2}\Bigg) =1-\alpha+ O\Bigg(\sqrt{n}h^2+\frac{K+\log n}{\sqrt {n}h}+\sqrt{\frac{K+\log n}{Nh}}+\sqrt{\frac{n\log n}{N}}\Bigg),
\end{align*}
which concludes the proof.

\subsection{Proof of Theorem \ref{thm:quantile_estimation+}}
By combining (\ref{eq:quantile-ee-bd1}), (\ref{equ:quantile-m-error}) and (\ref{eq:ppi-fstar-rep-quantile}) in the proof of Theorem \ref{thm:quantile_estimation}, we have
\begin{align*}
    \sqrt n\, \hat m_{\hat\beta_n}^\moe =Z_{n, f_\ast}+\tilde Z_{N, f_\ast} + \tilde O_p\Big(\sqrt nh^2+\frac{K+\log n}{\sqrt {n}h}\Big).
\end{align*}

Applying the Berry-Esseen bound to both $Z_{n, f_\ast}$ and $\tilde Z_{N, f_\ast}$, we get 
\begin{align*}
\sup_t\bigg|F_{\frac{Z_{n,f_{\ast}}}{\sqrt{Q(\theta_\ast,\beta_\ast)}}}(t) -\Phi(t) \bigg| =O\bigg(\frac{1}{\sqrt{n}}\bigg)
\qquad {\rm and} \qquad
\sup_t\bigg|F_{\frac{\tilde Z_{N,f_\ast}}{\sqrt{n/N}\sqrt{\tilde Q(\theta_\ast,\beta_\ast)}}}(t)-\Phi(t) \bigg|=O\bigg(\frac{1}{\sqrt{N}}\bigg),
\end{align*}
$F_{Y}(\cdot)$ denotes the c.d.f.  of the random variable $Y$. Moreover,  by the convergence rate for the sum of independent non-identically distributed random variables \citep{petrov2012sums},  we get 
\begin{align*}
\sup_t \Bigg|\PP\bigg(\frac{\sqrt{n}\,\hat m^{\moe}_{\hat\beta_n}(\theta_{\ast})}{\sqrt{Q(\theta_\ast, \beta_\ast)+(n/N)\tilde Q(\theta_\ast, \beta_\ast)}}\leq t\bigg)-\Phi(t) \Bigg|=O\bigg(\sqrt nh^2+\frac{K+\log n}{\sqrt {n}h}+\frac{1}{\sqrt{N}}\bigg).
\end{align*}
The rest of the proof follows the same as the proof of Theorem~\ref{thm:quantile_estimation}.

\subsection{Proof of Lemma~\ref{lem:mlr-moe-bias}}
Let us derive the explicit form of $\beta_{\ast}$ and $\hat\beta_n$, respectively. Recall $\bff:=\bff(X)=\big(f_1(X),\cdots,f_K(X)\big)^{\top}$ so that $F_{\beta}(X)=\bff^{\top}\beta$. 
By definition,  we write
\begin{align*}
\tr\big(\bW_{Y-F_{\beta}}&\big)=\big<\bSigma^{-2}, \EE\big[\big(Y-F_{\beta}(X)-X^{\top}\bdelta_{F_{\beta}}\big)^2XX^{\top}\big]\big>\\
=&\big<\bSigma^{-2}, \EE\big[\big(\big(X^{\top}\Sigma^{-1}\EE(X\bff^{\top})-\bff^{\top}\big)\beta\big)^2XX^{\top}\big]\\
+&2\EE\big[\underbrace{(Y-X^{\top}\bSigma^{-1}\EE(XY))}_{u(Y)}\underbrace{\big(X^{\top}\bSigma^{-1}\EE(X\bff^{\top})-\bff^{\top}\big)}_{\bv^{\top}(X)}\beta XX^{\top}\big]\big>+{\rm const.}
\end{align*}

\bigskip

\noindent\textit{Step 1: boundedness of $\|\hat\bH-\bH\|$ and $\|\hat \br-\br\|$.}

Denote
$$
\bH=\EE\Big[\big(X^{\top}\bSigma^{-2}X\big) \bv(X)\bv^{\top}(X)\Big]\quad  {\rm and}\quad \br=-\EE\Big[\big(X^{\top}\bSigma^{-2}X\big)u(Y)\bv(X)\Big]. 
$$
and their estimators
$$
\hat\bH=\frac{1}{n}\sum_{i=1}^n (X_i^{\top}\hat\bSigma^{-2}X_i)\hat\bv_i(X)\hat\bv_i^{\top}(X)\quad {\rm and}\quad \hat\br=-\frac{1}{n}\sum_{i=1}^n(X_i^{\top}\hat\bSigma^{-2}X_i)\hat u_i(Y)\hat \bv_i(X).
$$

Note that $\hat\bSigma=n^{-1}\bX^{\top}\bX=n^{-1}\sum_{i=1}^n X_iX_i^{\top}$.   Similarly,  we denote $\tilde\bSigma=N^{-1}\tilde \bX^{\top} \tilde \bX$.  

Suppose that Assumption~\ref{assump:mlr} holds and $n\gtrsim d\log n$.  Bernstein inequality dictates that there exists an event $\calE_1$ with $\PP(\calE_1)\geq 1-n^{-10}$,  under which the following bounds hold:
\begin{align}\label{eq:XX-con-bd}
\big\|n^{-1}\bX^{\top}\bX- \bSigma\big\|=O\bigg(\sqrt{\frac{d\log n}{n}}\bigg)\quad {\rm and}\quad \big\|N^{-1}\tilde \bX^{\top}\tilde \bX- \bSigma\big\|=O\bigg(\sqrt{\frac{d\log n}{N}}\bigg)
\end{align}
Under event $\calE_1$,  by Neumann series,  we have 
\begin{align*}
\big\|\hat \bSigma^{-1}-\bSigma^{-1}+\bSigma^{-1}(\hat\bSigma-\bSigma)\bSigma^{-1}-\bSigma^{-1}(\hat\bSigma-\bSigma)\bSigma^{-1}(\hat\bSigma-\bSigma)\bSigma^{-1}\big\|=O\big(\|\hat\bSigma-\bSigma\|^3\big).
\end{align*}
Therefore,  by handling $\EE \hat\bSigma^{-1}\II_{\calE_1}$ and $\EE \hat\bSigma^{-1}\II_{\calE_1^{\rm c}}$ separately,  we get 
\begin{align}\label{eq:EE-XXinv-bd}
\big\|\EE\hat\bSigma^{-1} -\bSigma^{-1}\big\|=O\bigg(\frac{d\log n}{n}\bigg)\quad {\rm and}\quad \big\|\EE \tilde \bSigma^{-1}-\bSigma^{-1} \big\|=O\bigg(\frac{d\log n}{N}\bigg),
\end{align}
where we used the fact $\EE \hat\bSigma=\EE\tilde\bSigma=\bSigma$.  

Similarly, there is an event $\calE_2$ with $\PP(\calE_2)\geq 1-n^{-10}$, under which the following bounds hold
$$
\Big\|\frac{\bX^{\top}\by}{n} -\EE YX\Big\|+\Big\|\frac{\bX^{\top}f_{\ast}(\bX)}{n}-\EE Xf_{\ast}(X) \Big\|=O\bigg(\sqrt{\frac{\log n}{n}}\bigg)\quad {\rm and}\quad \Big\|\frac{\tilde\bX^{\top}f_{\ast}(\tilde\bX)}{N}-\EE Xf_{\ast}(X) \Big\|=O\bigg(\sqrt{\frac{\log n}{N}}\bigg).
$$

Note that $\hat\bv_i^{\top}(X)-\bv_i^{\top}(X)=X_i^{\top}\big(\hat\bSigma^{-1}-\bSigma^{-1}\big)\big(n^{-1}\bX^{\top}\bF\big)+X_i^{\top}\bSigma^{-1}\big(n^{-1}\bX^{\top}\bF-\EE X\bff^{\top}\big)$. Therefore, on events $\calE_1\cap \calE_2$, we have 
\begin{align*}
\big|u_i(Y)-\hat u_i(Y)\big|=O\bigg(\sqrt{\frac{d\log n}{n}}\bigg)\quad {\rm and}\quad \big|\bv_i(X)-\hat\bv_i(X)\big|=O\bigg(\sqrt{\frac{Kd\log n}{n}}\bigg),\quad \forall i\in[n].  
\end{align*}
As a result, 
\begin{align*}
\big\|\hat\bH-\bH\big\|\leq& \bigg\|\frac{1}{n}\sum_{i=1}^n\Big(\big(X_i^{\top}\hat\bSigma^{-2}X_i\big)\hat\bv_i(X)\hat\bv_i^{\top}(X)-\big(X_i^{\top}\bSigma^{-2}X_i\big)\bv_i(X)\bv_i^{\top}(X)\Big) \bigg\|\\
&+\bigg\|\frac{1}{n}\sum_{i=1}^n \big(X_i^{\top}\bSigma^{-2}X_i\big)\bv_i(X)\bv_i^{\top}(X) -\EE\Big[\big(X^{\top}\bSigma^{-2}X\big)\bv(X)\bv^{\top}(X)\Big] \bigg\|,
\end{align*}
where the second term is bounded by $O\big(\sqrt{K^2\log(n)/n}\big)$ on an event $\calE_3 $ satisfying $\PP(\calE_3)\geq 1-n^{-10}$.  

Therefore,  on event $\calE_1\cap \calE_2\cap \calE_3$,  we get 
$$
\|\hat\bH-\bH\|=O\bigg(\sqrt{\frac{dK^2\log n}{n}}\bigg).
$$
We can bound $\|\br-\hat\br\|$ in a similar fashion and conclude that there exists an event $\calE_4$ with $\PP(\calE_4)\geq 1-n^{-10}$ such that the following bound holds 
$$
\|\hat\br-\br\|=O\bigg(\sqrt{\frac{Kd\log n}{n}}\bigg),
$$
under the event $\calE_1\cap \calE_2\cap\calE_3$. 

Moreover, there exists an event $\calE_0$ with $\PP(\calE_0)\geq 1-e^{-c_1n/(dK^2)}$ such that $\lambda_{\min}(\hat\bH)>0.01 c_0$. On the event $\calE_0^{\rm c}$, we simply have $\hat\beta_n=0$ by the design of Algorithm~\ref{algo:linear_regression}. 

\noindent\textit{Step 2: closed-form solution for $\beta_{\ast}$ and $\hat\beta_n$. } 

Since $\bH \succeq c_0\bI$, we have $\lambda_{\min}(\bH)\geq c_0$. By Weyl's inequality,
$$
\lambda_{\min}(\widehat \bH)
\ge
\lambda_{\min}(H)-\|\hat\bH-\bH\| \ge c_0-\|\hat\bH-\bH\|.
$$

Thus, for a sufficiently large $n$ such that $\|\hat\bH-\bH\|<c_0$, we have $\lambda_{\min}(\hat \bH)>0$ and consequently $\hat \bH$ is nonsingular.

By solving $\partial \tr\big(\bW_{Y-F\beta}\big)/\partial \beta=0$, we get $\beta_{\ast}=\bH^{-1}\br$. 

Note that $\EE Xu(Y)=0$ and $\EE X\bv^{\top}(X)={\bf 0}$. Similarly, by (\ref{eq:hat-cov-ppi}), we get
\begin{align*}
\tr\big(\hat\bW_{Y-F_{\beta}}&\big)=\Big<\hat\bSigma^{-2}, \frac{1}{n}\sum_{i=1}^n\Big(\underbrace{\big(Y_i-X_i^{\top}(\bX^{\top}\bX)^{-1}\bX^{\top}\by\big)}_{\hat u_i(Y)} +\underbrace{\big(X_i^{\top}(\bX^{\top}\bX)^{-1}\bX^{\top}\bF-\bff_i^{\top}\big)}_{\hat\bv_i^{\top}(X)}\beta\Big)X_iX_i^{\top}\Big>.
\end{align*}
Therefore, we have $\hat\beta_n=\hat\bH^{-1}\hat \br$.

\bigskip

\noindent\textit{Step 3: bounding $\hat\beta_n-\beta_{\ast}$. }  Note that $\hat\beta_n-\beta_{\ast}=\hat\bH^{-1}\hat\br-\bH^{-1}\br$.  To this end, define
$$
u_i(Y)=Y_i-X_i^{\top}\bSigma^{-1}\EE(XY)\quad {\rm and}\quad \bv_i^{\top}(X)=X_i^{\top}\bSigma^{-1}\EE(X\bff^{\top})-\bff_i^{\top},\quad \forall i\in[n].
$$
 It is then straightforward to show that 
\begin{align}\label{eq:mlr-hatbeta-bd}
\EE^{1/2}\|\hat\beta_n-\beta_{\ast}\|^2=O\bigg(\sqrt{\frac{dK^3\log n}{n}}\bigg),
\end{align}
if $n\gtrsim dK^2\log n$.  Note that we used the simple fact that $\|\hat\beta_n\|$ is uniformly bounded. Indeed, on the event $\calE_0$, we have $\lambda_{\min}(\hat\bH)>0.01c_0$,  and $\|\hat\br\|$ is uniformly bounded under Assumption~\ref{assump:mlr}; on the event $\calE_0^{\rm c}$, we have $\hat\beta_n=0$ by the design of Algorithm~\ref{algo:linear_regression}.

\bigskip

\noindent\textit{Step 4: bias of PPI-based estimator.} 
The bias of the oracle PPI-based estimator is 
\begin{align}
\big\|\EE\hat\theta^{\ppi}_{f_{\ast}}-\theta_{\ast} \big\|&\leq \big\|\EE (\bX^{\top}\bX)^{-1}\bX^{\top}\by-\theta_{\ast}\big\|\notag\\
&+\big\|\EE(\tilde\bX^{\top} \tilde\bX)^{-1}\tilde\bX^{\top}f_{\ast}(\tilde\bX) -\EE (\bX^{\top}\bX)^{-1}\bX^{\top}f_{\ast}(\bX)\big\|\label{eq:ee-ppi-bd1}
\end{align}
Observe that 
\begin{align*}
\big\|\EE (\bX^{\top}\bX)^{-1}&\bX^{\top}\by-\theta_{\ast}\big\|=\big\|\EE\big[\big(\hat\bSigma^{-1}-\bSigma^{-1}\big)n^{-1}\bX^{\top}\by\big]\big\|\\
\leq&\big\|\EE\big[\big(\hat\bSigma^{-1}-\bSigma^{-1}\big)(\EE YX)\big]\big\|+\big\|\EE\big[\big(\hat\bSigma^{-1}-\bSigma^{-1}\big)(n^{-1}\bX^{\top}\by-\EE YX)\big]\big\|\\
=&O\bigg(\frac{d\log n}{n}\bigg),
\end{align*}
where the last inequality is due to (\ref{eq:EE-XXinv-bd}) and the definition of events $\calE_1\cap \calE_2$, assuming that $N\gg n\gtrsim d\log n$. The second term in (\ref{eq:ee-ppi-bd1}) can be handled in a similar fashion. As a result, we conclude with
\begin{align}\label{eq:ee-ppi-bd2}
\big\|\EE\hat\theta_{f_{\ast}}^{\ppi}-\theta_{\ast} \big\|=O\bigg(\frac{d\log n}{n}\bigg).
\end{align}  

\bigskip

\noindent\textit{Step 5: bias of MOE-powered estimator.} 
Recall that $\hat\theta^{\moe}=\hat\theta^{\ppi}_{\hat\beta_n}$ and $f_{\ast}=F_{\beta_{\ast}}$.  Therefore,
\begin{align*}
\big\|\EE\hat\theta^{\moe}-&\theta_{\ast}\big\|= \Big\|\EE\big(\hat\theta^{\ppi}_{F_{\hat{\beta}_n}}-\hat\theta^{\ppi}_{F_{\beta_{\ast}}} \big)\Big\|+O\Big(\frac{d\log n}{n}\Big)\\
=&\Big\|\EE\Big[\big((\tilde\bX^{\top}\tilde\bX)^{-1}\tilde\bX^{\top}\tilde\bF-(\bX^{\top}\bX)^{-1}\bX^{\top}\bF\big)(\hat\beta_n-\beta_{\ast})\Big] \Big\|+O\bigg(\frac{d\log n}{n}\bigg)\\
\leq& \EE^{1/2}\Big\| \big((\tilde\bX^{\top}\tilde\bX)^{-1}\tilde\bX^{\top}\tilde\bF-(\bX^{\top}\bX)^{-1}\bX^{\top}\bF\Big\|^2\EE^{1/2}\|\hat\beta_n-\beta_{\ast}\|^2+O\bigg(\frac{d\log n}{n}\bigg).
\end{align*}
By the definition of events $\calE_1$ and $\calE_2$,  if $n\gtrsim (d+K)\log n$,  we have 
\begin{align}\label{eq:mlr-proof-XX-bd1}
 \EE^{1/2}\big\| \big((\tilde\bX^{\top}\tilde\bX)^{-1}\tilde\bX^{\top}\tilde\bF-(\bX^{\top}\bX)^{-1}\bX^{\top}\bF\big\|^2=O\bigg(\sqrt{\frac{K\log n}{n}}\bigg).
\end{align}
Together with (\ref{eq:mlr-hatbeta-bd}),  we conclude that 
$$
\big\|\EE\hat\theta^{\moe}-\theta_{\ast}\big\|=O\bigg(\frac{(K^2\sqrt{d}+d)\log n}{n}\bigg),
$$
which concludes the proof.

\subsection{Proof of Theorem~\ref{thm:mlr-MOE}}
Recall from the proof of Lemma~\ref{lem:mlr-moe-bias} that $\hat\beta_n=\hat\bH^{-1}\hat\br$ and $\beta_{\ast}=\bH^{-1}\br$. 

\noindent\textit{Step 1: connecting $\hat\theta^{\moe}$ to $\hat\theta^{\ppi}_{f_{\ast}}$}.  We first show that
\begin{equation}\label{eq:mlr-moe-ppi-df1}
    \sqrt n(\hat\theta^{\moe}-\theta_{\ast})=\sqrt{n}(\hat\theta^{\text{PPI}}_{f_{\ast}}-\theta_{\ast})+ \tilde O_p\Big(\frac{\sqrt{d}K^2\log n}{\sqrt{n}}\Big)
\end{equation}
where $f_{\ast}=F_{\beta_{\ast}}=\sum_{k=1}^K \beta_{\ast, k}f_k$.  

As shown in the Step 3 of the proof of Lemma~\ref{lem:mlr-moe-bias},  we have 
\begin{align*}
\|\hat\theta^{\moe}-\hat\theta^{\ppi}_{f_{\ast}}\|=\|\hat\theta_{F_{\hat\beta_n}}^{\ppi}-\hat\theta^{\ppi}_{F_{\beta_{\ast}}}\|\leq \big\| \big((\tilde\bX^{\top}\tilde\bX)^{-1}\tilde\bX^{\top}\tilde\bF-(\bX^{\top}\bX)^{-1}\bX^{\top}\bF\big\|\cdot \|\hat\beta_n-\beta_{\ast}\|
\end{align*}
Under the event $\calE_1\cap \calE_2\cap \calE_3$ defined in the proof of Lemma~\ref{lem:mlr-moe-bias},  we have 
$$
\big\| \big((\tilde\bX^{\top}\tilde\bX)^{-1}\tilde\bX^{\top}\tilde\bF-(\bX^{\top}\bX)^{-1}\bX^{\top}\bF\big\|=O\bigg(\sqrt{\frac{K\log n}{n}}\bigg)\quad {\rm and}\quad \|\hat\beta_n-\beta_{\ast}\|=O\bigg(\sqrt{\frac{dK^2\log n}{n}}\bigg),
$$
if $n\gtrsim dK^2\log n$,  which immediately leads to (\ref{eq:mlr-moe-ppi-df1}).

\bigskip

\noindent\textit{Step 2: normal approximation of $\hat\theta^{\ppi}_{f_{\ast}}$}.  It suffices to establish the convergence rate of $\sqrt{n}(\hat\theta^{\ppi}_{f_{\ast}}-\theta_{\ast})$ to a normal distribution.  By decomposing $\bSigma^{-1}\EE(XY)=\bSigma^{-1}\EE(Xf_{\ast}(X))+\bSigma^{-1}\EE\big(X(Y-f_{\ast}(X))\big)$ and the definition of $\hat\theta^{\ppi}_{f_{\ast}}$,  we get
\begin{align*}
\hat\theta^{\ppi}_{f_{\ast}}-\theta_{\ast}=&\underbrace{\big(\tilde\bX^{\top}\tilde\bX\big)^{-1}\tilde\bX^{\top}f_{\ast}(\tilde\bX)-\bSigma^{-1}\EE\big(Xf_{\ast}(X)\big)}_{\calJ_1}\\
+&\underbrace{(\bX^{\top}\bX)^{-1}\bX^{\top}\big(\by-f_{\ast}(\bX)\big)-\bSigma^{-1}\EE\big(X(Y-f_{\ast}(X))\big)}_{\calJ_2}.
\end{align*}
We first bound the term $\calJ_1$.  Note that
\begin{align*}
\calJ_1=\Big(\big(N^{-1}\tilde\bX^{\top}\tilde\bX\big)^{-1}-\bSigma^{-1}\Big)\big(N^{-1}\tilde\bX^{\top}f_{\ast}(\tilde\bX)\big)+\bSigma^{-1}\big(N^{-1}\tilde\bX^{\top}f_{\ast}(\tilde\bX)-\EE Xf_{\ast}(X)\big).
\end{align*}
By the definition of the event $\calE_1\cap \calE_2\cap \calE_3$ and assuming $N\gtrsim d\log n$,  we get 
\begin{align}\label{eq:mlr-moe-j1-bd}
\calJ_1=\tilde O_p\bigg(\sqrt{\frac{d\log n}{N}}\bigg),
\end{align}
where we used the fact $\|\beta_{\ast}\|=O(1)$ and $|f_{\ast}(X)|=O(1)$ almost surely.  This also implies that $\|\bdelta_{f_{\ast}}\|\lesssim\big(\big\|\EE\big(X(Y-f_{\ast}(X))\big) \big\|=O(1)$ under Assumption~\ref{assump:mlr}.

For $\calJ_2$ term,  we write
\begin{align*}
\calJ_2=&\big(\bX^{\top}\bX\big)^{-1}\bX^{\top}\big(\by-f_{\ast}(\bX)-\bX\bdelta_{f_{\ast}})\big)\\
=&\bSigma^{-1}n^{-1}\bX^{\top}\big(\by-f_{\ast}(\bX)-\bX\bdelta_{f_{\ast}})\big)+\big((\bX^{\top}\bX/n)^{-1}-\bSigma^{-1}\big)\cdot\big(n^{-1}\bX^{\top}\big(\by-f_{\ast}(\bX)-\bX\bdelta_{f_{\ast}})\big).
\end{align*}
By Bernstein inequality,  there exists an event $\calE_4$ with $\PP(\calE_4)\geq 1-n^{-10}$ such that 
\begin{align}\label{eq:mlr-moe-j2-bd}
\Big|\frac{1}{n}\bX^{\top}(\by-f_{\ast}(\bX)-\bX\bdelta_{f_{\ast}})\Big|=O\bigg(\sqrt{\frac{\log n}{n}}\bigg),
\end{align}
where we used the fact $\EE\big(X(Y-f_{\ast}(X)-X^{\top}\bdelta_{f_{\ast}})\big)=0$.  Then,  on event $\calE_1\cap \calE_4$,  we get 
$$
\sqrt{n}\calJ_2=\bSigma^{-1}\cdot \bigg(\frac{1}{\sqrt{n}}\sum_{i=1}^n\big(Y_i-f_{\ast}(X_i)-X_i^{\top}\bdelta_{f_{\ast}}\big)X_i\bigg)+\tilde O_p\bigg(\sqrt{\frac{d\log^2n}{n}}\bigg)
$$

Based on (\ref{eq:mlr-moe-j1-bd}) and (\ref{eq:mlr-moe-j2-bd}),  we get 
\begin{equation*}
    \sqrt n(\hat\theta^{\ppi}_{f_{\ast}}-\theta_{\ast})=\bSigma^{-1}\cdot \bigg(\frac{1}{\sqrt{n}}\sum_{i=1}^n\big(Y_i-f_{\ast}(X_i)-X_i^{\top}\bdelta_{f_{\ast}}\big)X_i\bigg)+\tilde O_p\bigg(\sqrt{\frac{dn\log n}{N}}+\sqrt{\frac{d\log^2 n}{n}}\bigg).
\end{equation*}
By the multivariate Berry-Esseen theorem \citep{raivc2019multivariate}, we get 
\begin{align}\label{eq:mlr-ppi-be-bd}
\sup_{\calU}\Big|\PP\Big(\sqrt{n}\big(\hat\theta^{\ppi}_{f_{\ast}}-\theta_{\ast}\big)\in\calU\Big)-\PP\big(T_{f_{\ast}}\in\calU\big) \Big|=O\bigg(\sqrt{\frac{d^2n\log n}{N}}+\sqrt{\frac{d^2\log^2 n}{n}}\bigg),
\end{align}
where $T_{f_{\ast}}\sim \calN\big(0, \bSigma^{-1}\big(\EE(Y-f_{\ast}(X)-X^{\top}\bdelta_{f_{\ast}})^2XX^{\top}\big)\bSigma^{-1}\big)$ and $\calU$ is taken supremum over all convex sets in $\RR^d$. Note that we also used the Nazarov-type inequality \citep{nazarov2004maximal} concerning the Gaussian surface area of convex sets. 

Moreover, for a fixed index $s\in[d]$, we can also get 
\begin{align}\label{eq:mlr-ppi-be-bd1}
\sup_{t\in\RR}\bigg|\PP\bigg(\frac{\sqrt{n}\big(\hat\theta^{\ppi}_{f_{\ast},s}-\theta_{\ast,s}\big)}{V_{\ast,s}}\leq t\bigg)-\Phi(t) \bigg|=O\bigg(\sqrt{\frac{dn\log n}{N}}+\sqrt{\frac{d\log^2 n}{n}}\bigg),
\end{align}
where $V_{\ast,s}^2:=\be_s^{\top}\bSigma^{-1}\big(\EE(Y-f_{\ast}(X)-X^{\top}\bdelta_{f_{\ast}})^2XX^{\top}\big)\bSigma^{-1}\be_s$. 

\bigskip

\noindent\textit{Step 3: Berry-Esseen bound for $\sqrt{n}\big(\hat\theta^{\moe}-\theta_{\ast}\big)$.} Combining (\ref{eq:mlr-moe-ppi-df1}) and (\ref{eq:mlr-ppi-be-bd1}),  we get 
\begin{align}\label{eq:mlr-moe-be-bd}
\sup_{t\in\RR}\bigg|\PP\bigg(\frac{\sqrt{n}\big(\hat\theta^{\moe}_{s}-\theta_{\ast,s}\big)}{V_{\ast,s}}\leq t\bigg)-\Phi(t) \bigg|=O\bigg(\sqrt{\frac{dn\log n}{N}}+\frac{\sqrt{d}K^2\log n}{\sqrt{n}}\bigg),
\end{align}
for any fixed index $s\in[d]$.  

Denote 
$$
\hat V_{\ast, s}^2=\be_s^{\top}\hat\bSigma^{-1}\cdot \bigg(\frac{1}{n}\sum_{i=1}^n\big(Y_i-F_{\hat\beta_n}(X_i) - X_i^{\top}\hat\bdelta_{F_{\hat\beta_n}}\big)^2X_iX_i^{\top}\bigg)\hat\bSigma^{-1}\be_s,
$$
where $F_{\hat\beta_n}(X_i)=\bff_i^{\top}\hat\beta_n$ and $\hat\bdelta_{F_{\hat\beta_n}}=\hat\bSigma^{-1}\cdot n^{-1}\sum_{i=1}^n\big(Y_i-F_{\hat\beta_n}(X_i)\big)X_i$.  
Under the event $\calE_1\cap \calE_2\cap\calE_3$,  we have 
$$
\big|F_{\hat\beta_n}(X_i)-f_{\ast}(X_i)\big|=\big|\bff_i^{\top}(\hat\beta_n-\beta_{\ast})\big|\lesssim \sqrt{K}\|\hat\beta_n-\beta_{\ast}\|=O\bigg(\sqrt{\frac{dK^3\log n}{n}}\bigg),\quad \forall i\in[n],  
$$
and as a result
\begin{align*}
\big\|\hat\bdelta_{F_{\hat\beta_n}}-\hat\bdelta_{f_{\ast}} \big\|\lesssim \frac{1}{n}\sum_{i=1}^n \Big|F_{\hat\beta_n}(X_i)-f_{\ast}(X_i) \Big|\|X_i\|=O\bigg(\sqrt{\frac{dK^3\log n}{n}}\bigg). 
\end{align*}
Therefore,  on event $\calE_1\cap\calE_2\cap \calE_3$,  
\begin{align*}
\big\|\hat\bdelta_{F_{\hat\beta_n}}-\bdelta_{f_{\ast}} \big\|\leq& \big\| \hat\bdelta_{f_{\ast}}-\bdelta_{f_{\ast}}\big\|+O\bigg(\sqrt{\frac{dK^3\log n}{n}}\bigg)\\
\leq&\big\|\hat\bSigma^{-1} \big\|\cdot\Big\|\frac{1}{n}\sum_{i=1}^{n}\big(Y_i-f_{\ast}(X_i)\big)X_i-\EE\big(Y-f_{\ast}(X)\big)X \Big\|+O\bigg(\sqrt{\frac{dK^3\log n}{n}}\bigg)\\
&+\big\|\hat\bSigma^{-1}-\bSigma^{-1} \big\|\cdot \big\|\EE\big(Y-f_{\ast}(X)\big)X \big\|\\
=&\tilde O_p\bigg(\sqrt{\frac{dK^3\log n}{n}}\bigg),
\end{align*}
where the last bound holds due to the definition of the event $\calE_1\cap \calE_2\cap \calE_3$.   This also implies that 
$$
\Big|\big(Y_i-F_{\hat\beta_n}(X_i) - X_i^{\top}\hat\bdelta_{F_{\hat\beta_n}}\big)^2-\big(Y_i-f_{\ast}(X_i) - X_i^{\top}\bdelta_{f_{\ast}}\big)^2 \Big|=O\bigg(\sqrt{\frac{dK^3\log n}{n}}\bigg),\quad \forall i\in[n],
$$
on the same event.  As a result,  on the event $\calE_1\cap\calE_2\cap \calE_3$,  we have 
\begin{align*}
\hat V_{\ast,  s}^2= &\be_s^{\top}\hat\bSigma^{-1}\cdot \bigg(\frac{1}{n}\sum_{i=1}^n\big(Y_i-f_{\ast}(X_i) - X_i^{\top}\bdelta_{f_{\ast}}\big)^2X_iX_i^{\top}\bigg)\cdot \hat\bSigma^{-1}\be_s+O\bigg(\sqrt{\frac{dK^3\log n}{n}}\bigg).
\end{align*}
By Bernstein inequality \citep{tropp2012user},  there exists an event $\calE_5$ with $\PP(\calE_5)\geq 1-n^{-10}$ on which,
$$
\bigg\|\frac{1}{n}\sum_{i=1}^n\big(Y_i-f_{\ast}(X_i) - X_i^{\top}\bdelta_{f_{\ast}}\big)^2X_iX_i^{\top}-\EE \big(Y-f_{\ast}(X)-X^{\top}\bdelta_{f_{\ast}}\big)XX^{\top} \bigg\|=O\bigg(\sqrt{\frac{d\log n}{n}}\bigg).  
$$
Therefore,  on event $\cap_{k=1}^5 \calE_k$,  we have 
$$
\hat V_{\ast,s}^2=V_{\ast,s}^2+O\bigg(\sqrt{\frac{dK^3\log n}{n}}\bigg) \textrm{ implying that } \frac{\hat V_{\ast ,s}}{V_{\ast ,s}}=1+O\bigg(\sqrt{\frac{dK^3\log n}{n}}\bigg).
$$
Following the same argument as that in the proof of Theorem~\ref{thm:mean_estimation},  we conclude that 
\begin{align*}
\sup_{t\in\RR}\bigg|\PP\bigg(\frac{\sqrt{n}\big(\hat\theta^{\moe}_{s}-\theta_{\ast,s}\big)}{\hat V_{\ast,s}}\leq t\bigg)-\Phi(t) \bigg|=O\bigg(\sqrt{\frac{dn\log n}{N}}+\frac{\sqrt{dK^3}\log n}{\sqrt{n}}\bigg).  
\end{align*}
The rest of the proof is straightforward and hence omitted.  

\subsection{Proof of Lemma~\ref{lem:logistic-moe-bias}}

{\it Step 1: formulas for $\beta_{\ast}$ and $\hat\beta_n$.} Let us derive the explicit form and difference bound of $\beta_\ast$ and $\hat\beta_n$. Note that 
\begin{align*}
    \tr\big(\bW_{Y-F_{\beta}}\big) &= \tr\Big(\EE \big[(\bff^\top\beta-Y)^2XX^\top\big] -\EE\big[(\bff^\top\beta-Y)X\big]\EE\big[(\bff^\top\beta-Y)X^\top\big]\Big)\\
    &=\EE \big[(\bff^\top\beta-Y)^2X^\top X\big]-\EE\big[(\bff^\top\beta-Y)X^\top\big]\EE\big[(\bff^\top\beta-Y)X\big]\\
    &=\beta^\top \underbrace{\Big(\EE \bff X^\top X\bff^\top-(\EE \bff X^\top)( \EE X\bff^\top)\Big)}_{\bH}\beta-2\underbrace{\Big(\EE YX^\top X\bff^\top-(\EE Y X^\top)(\EE X\bff^\top)\Big)}_{\br}\beta+\text{constant};
\end{align*}

\begin{align*}
    \tr\big(\hat\bW_{Y-F_{\beta}}\big) &= \tr\bigg(n^{-1}\sum_{i=1}^n \big[(\bff_i^\top\beta-Y_i)^2 X_iX_i^\top\big]-n^{-2}\sum_{i=1}^n \big[(\bff_i^\top\beta-Y_i) X_i\big]\sum_{i=1}^n \big[(\bff_i^\top\beta-Y_i) X_i^\top\big]\bigg)\\
    &= n^{-1}\sum_{i=1}^n \big[(\bff_i^\top\beta-Y_i)^2 X_i^\top X_i\big] - n^{-2}\sum_{i=1}^n \big[(\bff_i^\top\beta-Y_i) X_i^\top\big]\sum_{i=1}^n \big[(\bff_i^\top\beta-Y_i) X_i\big]\\
    &= \beta^\top\underbrace{\bigg(n^{-1}\sum_{i=1}^n \bff_iX_i^\top X_i\bff_i^\top-n^{-2}\sum_{i=1}^n \bff_i X_i^\top\sum_{i=1}^n  X_i\bff_i^\top\bigg)}_{\hat\bH}\beta\\
    &\quad-2\underbrace{\bigg(n^{-1}\sum_{i=1}^n Y_iX_i^\top X_i\bff_i^\top-n^{-2}\sum_{i=1}^n Y_iX_i^\top\sum_{i=1}^n  X_i\bff_i^\top\bigg)}_{\hat\br}\beta+\text{constant}.
\end{align*}


Under Assumption~\ref{assump:logistic}, $\bH$ is invertible so that $\beta_{\ast}=\bH^{-1}\br$ exists and is unique.

By Bernstein Inequality, there exist an event $\calE_1$ with $\PP(\calE_1)\geq 1-n^{-11}$ such that 
\begin{align*}
    \max\bigg\{&\Big\|n^{-1}\sum_{i=1}^n \bff_iX_i^\top X_i\bff_i^\top-\EE \bff X^\top X\bff^\top\Big\|,
    \Big\|n^{-1}\sum_{i=1}^n \bff_i X_i^\top-\EE \bff X^\top\Big\|,\\
    &\Big\|n^{-1}\sum_{i=1}^n Y_iX_i^\top X_i\bff_i^\top-\EE YX^\top X\bff^\top\Big\|,
    \Big\|n^{-1}\sum_{i=1}^n Y_iX_i^\top-\EE Y X^\top \EE X\bff^\top\Big\|\bigg\}= O\bigg(\sqrt{\frac{(d+K)\log n}{n}}\bigg).
\end{align*}
Conditioned on $\calE_1$, we have
\begin{align*}
    &\max\Big\{\big\|\hat \bH-\bH\big\|,\big\|\hat \br-\br\big\|\Big\} = O\bigg(\sqrt{\frac{(d+K)\log n}{n}}\bigg).
\end{align*}
If $n\geq C_1(d+K)\log n$ for a large enough $C_1>0$, we have $\big\|\hat\bH-\bH\big\| \le {c_0}/{2}$ on the event $\calE_1$.  Then Weyl's inequality yields
\begin{align*}
    \lambda_{\min}(\hat \bH) \ge \lambda_{\min}(\bH)-\big\|\hat \bH-\bH\big\| \ge  \frac{c_0}{2},
\end{align*}
implying that $\hat\beta_n=\hat\bH^{-1}\hat\br$. 

Moreover, there exists an event $\calE_0$ with $\PP(\calE_0)\geq 1-e^{c_1n/(d+K)}$ such that $\lambda_{\min}(\hat \bH)\geq 0.01c_0$. On the event $\calE_0$, $\|\hat\beta_n\|$ is uniformly bounded under Assumption~\ref{assump:logistic}; on the event $\calE_0^{\rm c}$, $\hat\beta_n=0$ by Algorithm~\ref{algo:logistic_regression}. Therefore, we conclude that $\|\hat\beta_n\|$ is uniformly bounded. 

\noindent{\it Step 2: upper bound for $\|\hat\beta_n-\beta_{\ast}\|$.} 
Recall that on event $\calE_1$,  
\begin{align}\label{eq:logistic-moe-beta-1}
    \beta_\ast =\bH^{-1}\br\quad \text{and} \quad \hat\beta_n =\hat\bH^{-1}\hat \br.
\end{align}
Then,
\begin{align*}
    \big\|\hat \beta_n-\beta_\ast\big\|=\big\|\hat \bH^{-1}\hat \br-\bH^{-1}\br\big\|
    \leq \big\|\hat \bH^{-1}\big\|\big\|\hat \bH-\bH\big\|\big\|\bH^{-1}\big\|\cdot \|\hat \br\|+\|\bH^{-1}\|\cdot \big\|\hat \br -\br\big\|.
\end{align*}
Under Assumption~\ref{assump:logistic}, we have $\big\|\bH^{-1}\big\| \le c_0^{-1}$. Putting together the bounds above, we have conditional on $\calE_1$ that
\begin{align}\label{eq:logistic-moe-beta-err-rate}
    \big\|\hat \beta_n-\beta_\ast\big\|=O\bigg(\sqrt{\frac{(d+K)\log n}{n}}\bigg).
\end{align}

\noindent {\it Step 3: upper bound for $\big\|\EE\hat \bmm^{\moe}_{\hat\beta_n}(\theta)-\bmm(\theta))\big\|$.} 
For any fixed $\theta\in\Theta$, by definition, we get 
\begin{align}
    &\Big\|\EE\hat \bmm_{\hat\beta_n}^\moe(\theta)-\EE X(Y-S(X^\top\theta))\Big\| \\
    = &\Big\|N^{-1}\EE\tilde \bX^\top\Big(\tilde \bF\hat\beta_n-S(\tilde \bX\theta)\Big) 
    - n^{-1}\EE \bX^\top\Big( \bF\hat\beta_n-\by)\Big)\Big\|\notag\\
    \leq &\Big\|\EE\Big[\big(N^{-1}\tilde \bX^\top\tilde\bF - n^{-1}\bX^\top\bF\big) (\hat\beta_n-\beta_\ast)\Big]\Big\| 
    + \Big\|\big(N^{-1}\EE\tilde \bX^\top\tilde\bF - n^{-1}\EE \bX^\top\bF\big) \beta_\ast\Big\| \notag\\
    & + \Big\|n^{-1} \EE\bX^\top \by -\EE XY\Big\|+\Big\|N^{-1}\EE\tilde \bX S(\tilde \bX^\top\theta)-\EE XS(X^\top\theta))\Big\|.\label{eq:moe-logistic-bias-bd1}
\end{align}
Since $N^{-1}\EE\tilde \bX^\top\tilde\bF = n^{-1}\EE \bX^\top\bF=\EE X\bff^{\top}$, $n^{-1}\EE \bX^\top \by=\EE XY$ and $N^{-1}\EE \tilde\bX S(\tilde\bX^\top \theta)=\EE XS(X^\top\theta)$, the latter three terms of (\ref{eq:moe-logistic-bias-bd1}) vanish. It suffices to bound the first term. 
\begin{align*}
    &\Big\|\EE\Big[\big(N^{-1}\tilde \bX^\top\tilde\bF - n^{-1}\bX^\top\bF\big) (\hat\beta_n-\beta_\ast)\Big]\Big\|\\
    \leq &\,\EE\Big\|\big(N^{-1}\tilde \bX^\top\tilde\bF - n^{-1}\bX^\top\bF\big) (\hat\beta_n-\beta_\ast)\Big\|\\
    \leq &\,\EE\Big[\big(\big\|n^{-1} \bX^\top\bF-\EE X  \bff \big\|+\big\|N^{-1} \tilde \bX^\top\tilde\bF-\EE X \bff \big\|\big)\cdot\|\hat\beta_n-\beta_\ast\|\Big]
\end{align*}
By Bernstein inequality, there exists an event $\calE_2$ with $\PP(\calE_2)\geq 1-n^{-11}$ such that
\begin{align*}
    \Big\|n^{-1} \bX^\top\bF-\EE X \bff \Big\| = O\bigg(\sqrt{\frac{(d+K)\log n}{n}}\bigg)\quad \text{and}\quad \Big\|N^{-1} \tilde \bX^\top\tilde\bF-\EE X \bff \Big\| = O\bigg(\sqrt{\frac{(d+K)\log n}{N}}\bigg).
\end{align*}
Combining with (\ref{eq:logistic-moe-beta-err-rate}), we then have 
\begin{align*}
    &\,\Big\|\EE\hat \bmm_{\hat\beta_n}^\moe(\theta)-\EE X(Y-S(\theta^\top X))\Big\|\\
    \leq&\, \EE\Big[\big(\big\|n^{-1} \bX^\top\bF-\EE X  \bff \big\|+\big\|N^{-1} \tilde \bX^\top\tilde\bF-\EE X \bff \big\|\big)\cdot\|\hat\beta_n-\beta_\ast\|\II_{\calE_1\cap\calE_2}\Big]\\
    &+\EE\Big[\big(\big\|n^{-1} \bX^\top\bF-\EE X  \bff \big\|+\big\|N^{-1} \tilde \bX^\top\tilde\bF-\EE X \bff \big\|\big)\cdot\|\hat\beta_n-\beta_\ast\|\II_{\calE_1^{\rm c}\cup\calE_2^{\rm c}}\Big]\\
    =&\,O\bigg(\frac{{(d+K)\log n}}{n}\bigg),
\end{align*}
where the last inequality holds due to the boundedness of $X$, $\bff$, and $\beta$. 
Note that we need an upper bound on $\EE^{1/2}\|\hat\beta_n-\beta_{\ast}\|^2$, which was established at the end of {\it Step 1}.

\subsection{Proof of Theorem~\ref{thm:logistic_estimation}}
\noindent{\it Step 1: normal approximation of $\hat\bmm^{\moe}_{\hat\beta_n}(\theta)$}. 
By definition, for any $\theta\in\Theta$, we have
\begin{align}\label{eq:moe-logistic-thm-eq1}
    \sqrt{n}\,\big(\hat \bmm_{\hat\beta_n}^\moe(\theta)-\bmm(\theta)\big) &= \sqrt{n}\Big({\hat \bmm_{\hat\beta_n}^\moe(\theta)-\hat \bmm_{f_\ast}^\ppi(\theta)}\Big) + \sqrt n\,  \big(\hat \bmm_{f_\ast}^\ppi(\theta)-\bmm(\theta)\big),
\end{align}
where $f_\ast=F_{\beta_\ast}$ and $\bmm(\theta) = \EE\hat \bmm_{f_\ast}^\ppi(\theta) =\EE X(Y-S(X^\top\theta))$.

Observe that
\begin{align*}
    \Big\|\hat \bmm_{\hat\beta_n}^\moe(\theta)-\hat \bmm_{f_\ast}^\ppi(\theta)\Big\|
    =&\,\Big\|\big(N^{-1}\tilde \bX^\top \tilde\bF-n^{-1} \bX^\top \bF \big) \big(\hat\beta_n-\beta_\ast\big)\Big\|\\
    \leq &\,\Big(\Big\| N^{-1}\tilde \bX^\top \tilde\bF-\EE X\bff \Big\|+\Big\|n^{-1} \bX^\top\bF-\EE X\bff \Big\|\Big)\Big\|\hat\beta_n-\beta_\ast\Big\|.
\end{align*}
On the event $\calE_1\cap \calE_2$ defined in the proof of Lemma~\ref{lem:logistic-moe-bias}, we have
\begin{align}\label{eq:moe-ppi-diff-logistic}
    \sqrt{n}\Big\|\hat \bmm_{\hat\beta_n}^\moe(\theta)-\hat \bmm_{f_\ast}^\ppi(\theta)\Big\| = O\bigg({\frac{{(d+K)\log n}}{\sqrt n}}\bigg).
\end{align}
For second term in the RHS of (\ref{eq:moe-logistic-thm-eq1}) can be written as 
\begin{equation}\label{eq:logistic-normality}
    \begin{split}
        \sqrt n\,  \big(\hat \bmm_{f_\ast}^\ppi(\theta)-\bmm(\theta)\big) &= -\frac{1}{\sqrt{n}}\sum_{i=1}^n \Big[X_i\big(F_{\beta_{\ast}}(X_i)-Y_i\big)-\EE X(F_{\beta_{\ast}}(X) - Y)\Big] \\
        &\quad +\frac{\sqrt{n}}{N}\sum_{i=1}^N \Big[\tilde X_i\Big(F_{\beta_{\ast}}(\tilde X_i)-S\big(\tilde X_i^\top \theta\big)\Big)-\EE X(F_{\beta_\ast}(X)-S(X^\top\theta))\Big]\\
        &=\, -\bZ_{n,f_\ast} + \sqrt{\frac nN}\tilde \bZ_{N, f_\ast},
    \end{split}
\end{equation}
where $\tilde \bZ_{N, f_\ast} := N^{-1/2}\sum_{i=1}^N\Big[ \tilde X_i\Big(F_{\beta_\ast}(\tilde X_i)-S\big(\tilde X_i^\top \theta\big)\Big)-\EE X\big(F_{\beta_\ast}(X)-S(X^\top\theta)\big)\Big]$.

Since $\|X\|, |Y|$ and $\|\bff\|$ are bounded under Assumption~\ref{assump:logistic}, by Bernstein inequality, there exists an event $\calE_3$ with $\PP(\calE_3)\geq1-n^{-10}$ such that the following equality holds
\begin{align*}
    \sqrt n\,  \big(\hat \bmm_{\beta_\ast}^\ppi(\theta) - \bmm(\theta))\big) = -\bZ_{n,f_\ast} + O\Big(\sqrt{\frac{{(d+K)n\log n}}{N}}\Big)
\end{align*}
on the event $\calE_1\cap\calE_2\cap\calE_3$.  This proves the first claim that
\begin{align}\label{eq:logistic-normality-1}
    \sqrt n\,  \big(\hat \bmm_{\hat\beta_n}^\moe(\theta)- \bmm(\theta))\big) = -\bZ_{n,f_\ast} + \tilde O_p\Big({\frac{{(d+K)\log n}}{\sqrt n}}+\sqrt{\frac{{(d+K)n\log n}}{N}}\Big).
\end{align}

\noindent{\it Step 2: coverage probability.}  We continue from (\ref{eq:logistic-normality-1}). Note that $\bmm(\theta_{\ast})=0$.
By the Berry-Esseen bound and the high probability bound inherited from $\tilde O_p(\cdot)$, we get
\begin{align}\label{eq:logistic-moe-berry}
    \sup_{t\in\RR}\Bigg|\PP\bigg(\frac{\sqrt{n}\, \be_s^\top\hat \bmm_{\hat\beta_n}^\moe (\theta_\ast)}{\sqrt{\be_s^\top\bW_{Y-F_{\beta_{\ast}}}\be_s}}\leq t\bigg)-\Phi(t)\Bigg|=O\bigg({\frac{{(d+K)\log n}}{\sqrt n}}+\sqrt{\frac{{(d+K)n\log n}}{N}}\bigg),\forall s\in[d].
\end{align}
Therefore, 
\begin{align*}
    \PP\bigg( \Bigg|\frac{{\sqrt{n}\, \be_s^\top\hat \bmm_{\hat\beta_n}^\moe (\theta_\ast)}}{\sqrt{\be_s^\top\bW_{Y-F_{\beta_\ast}}\be_s}}\Bigg|\leq z_{\alpha/(2d)} \bigg)=1-\alpha/d + O\bigg({\frac{{(d+K)\log n}}{\sqrt n}}+\sqrt{\frac{{(d+K)n\log n}}{N}}\bigg),\forall s\in[d].
\end{align*}

Then it's sufficient to bound $\big\|\hat\bW_{Y-F_{\hat\beta_n}}-\bW_{Y-F_{\beta_{\ast}}}\big\|$. By the derivation of $\hat\bW_{Y-F_{\beta}}$ from the proof of Lemma~\ref{lem:logistic-moe-bias}, we have
\begin{align*}
    &\big\|\hat\bW_{Y-F_{\hat\beta_n}}-\bW_{Y-F_{\beta_{\ast}}}\big\|\\
    \leq &\big\|\hat\bW_{Y-F_{\hat\beta_n}}-\hat\bW_{Y-F_{\beta_{\ast}}}\big\| +\big\|\hat\bW_{Y-F_{\beta_{\ast}}}-\bW_{Y-F_{\beta_{\ast}}}\big\|\\
    \leq&\Big\|\big(\hat\beta_n+\beta_\ast\big)^\top \scov\big(\bff^\top X, \bff^\top X \big)\Big\|\cdot \big\|\hat\beta_n-\beta_\ast \big\| + \Big\| \scov\big(Y, \bff^\top X \big)\Big\|\cdot \big\|\hat\beta_n-\beta_\ast\big\|\\
    &+ \bigg\|n^{-1}\sum_{i=1}^n \big(F_{\beta_\ast}(X_i)-Y_i\big)X_i-\EE(F_{\beta_\ast}(X)-Y)X\bigg\|\cdot \bigg\|n^{-1}\sum_{i=1}^n \big(F_{\beta_\ast}(X_i)-Y_i\big)X_i+\EE(F_{\beta_\ast}(X)-Y)X\bigg\|\\
    &+\bigg\|n^{-1}\sum_{i=1}^n \big(\bff_i^\top\beta_\ast-Y_i\big)^2X_iX_i^\top-\EE(\bff^\top \beta_\ast-Y)^2XX^\top\bigg\|
\end{align*}
By Bernstein inequality, there exists an event $\calE_4$ with $\PP(\calE_4)\geq 1-n^{-10}$ such that  
\begin{align*}
    \max\Bigg\{&\bigg\|n^{-1}\sum_{i=1}^n \big(\bff_i^\top\beta_\ast-Y_i\big)X_i-\EE(\bff^\top \beta_\ast-Y)X\bigg\|,\\
    &\bigg\|n^{-1}\sum_{i=1}^n \big(\bff_i^\top\beta_\ast-Y_i\big)^2X_iX_i^\top-\EE(\bff^\top \beta_\ast-Y)^2XX^\top\bigg\|\Bigg\} = O\Big(\sqrt{\frac{d\log n}{n}}\Big).
\end{align*}
Due to the bound (\ref{eq:logistic-moe-beta-err-rate}) and the uniform boundedness of $\|X\|, |Y|$ and $|f_k(X)|$, conditional on $\bigcap_{i=1}^4\calE_i$, we get
\begin{align*}
    \big\|\hat\bW_{Y-F_{\hat\beta_n}}-\bW_{Y-F_{\beta_{\ast}}}\big\| = O\Big(\sqrt{\frac{ (d+K) \log n}{n}}\Big).
\end{align*}

Observe that 
\begin{align*}
    \frac{\sqrt{n}\, \be_s^\top\hat \bmm_{\hat\beta_n}^\moe (\theta_\ast)}{\sqrt{\be_s^\top\hat\bW_{Y-F_{\hat\beta_n}}\be_s^\top}} = 
    \frac{\sqrt{n}\, \be_s^\top\hat \bmm_{\hat\beta_n}^\moe (\theta_\ast)}{\sqrt{\be_s^\top\bW_{Y-F_{\beta_{\ast}}}\be_s}}\Bigg(1+\frac{\be_s^\top\big[\hat\bW_{Y-F_{\hat\beta_n}}-\bW_{Y-F_{\beta_{\ast}}}\big]\be_s}{\sqrt{\be_s^\top\bW_{Y-F_{\beta_{\ast}}}\be_s} \Big(\sqrt{\be_s^\top\hat\bW_{Y-F_{\hat\beta_n}}\be_s} + \sqrt{\be_s^\top\bW_{Y-F_{\beta_{\ast}}}\be_s}\Big)}\Bigg)
\end{align*}
By the normal approximation in (\ref{eq:logistic-normality-1}) and the boundedness of covariance matrix, there exists an event $\calE_5$ with $\PP(\calE_5)\geq 1-n^{-10}$ such that
\begin{align*}
    \frac{\sqrt{n}\, \be_s^\top\hat \bmm_{\hat\beta_n}^\moe (\theta_\ast)}{\sqrt{\be_s^\top\hat\bW_{Y-F_{\hat\beta_n}}\be_{s}}} = \frac{\sqrt{n}\, \be_s^\top \hat \bmm_{\hat\beta_n}^\moe (\theta_\ast)}{\sqrt{\be_s^\top\bW_{Y-F_{\beta_{\ast}}}\be_{s}}}+O\bigg(\sqrt{\frac{d+K}{n}}\log n\bigg)
\end{align*}
when $n\gtrsim (d+K)^2\log^2 n$ and $N\gtrsim (d+K)n\log n$. 

Together with (\ref{eq:logistic-moe-berry}), we conclude that 
\begin{align}\label{eq:logistic-moe-berry-1}
    \sup_{t\in\RR}\Bigg|\PP\bigg(\frac{\sqrt{n}\, \be_s^\top\hat \bmm_{\hat\beta_n}^\moe (\theta_\ast)}{\sqrt{\be_s^\top\hat\bW_{Y-F_{\hat\beta_n}}\be_{s}}}\leq t\bigg)-\Phi(t)\Bigg|=O\bigg({\frac{{(d+K)\log n}}{\sqrt n}}+\sqrt{\frac{{(d+K)n\log n}}{N}}\bigg),
\end{align}
implying that
\begin{align*}
    \PP\Big(\theta_\ast \in\calC_\alpha^\moe\Big) &= \PP\Bigg(\bigcap_{s=1}^d\Bigg\{\Bigg|\frac{\sqrt{n}\, \be_s^\top\hat \bmm_{\hat\beta_n}^\moe (\theta_\ast)}{\sqrt{\be_s^\top\hat\bW_{Y-F_{\hat\beta_n}}\be_{s}}}\Bigg|\leq z_{\alpha/(2d)}\Bigg\}\Bigg)\\
    &\geq1-\prod_{s=1}^d\PP\Bigg(\Bigg|\frac{\sqrt{n}\, \be_s^\top\hat \bmm_{\hat\beta_n}^\moe (\theta_\ast)}{\sqrt{\be_s^\top\hat\bW_{Y-F_{\hat\beta_n}}\be_{s}}}\Bigg|> z_{\alpha/(2d)}\Bigg)\\
    &=1-\alpha + O\bigg({\frac{{d(d+K)\log n}}{\sqrt n}}+\sqrt{\frac{{d^2(d+K)n\log n}}{N}}\bigg),
\end{align*}
which concludes the proof.

\subsection{Proof of Theorem~\ref{thm:logistic+}}
For any fixed $\theta\in\Theta$, by (\ref{eq:moe-ppi-diff-logistic}) and (\ref{eq:logistic-normality}) in the Proof of Theorem~\ref{thm:logistic_estimation},  we get 
\begin{align*}
\sqrt{n}\,\big(\hat \bmm_{\hat\beta_n}^{\moe}(\theta)-\bmm(\theta)\big) = -\bZ_{n, f_\ast}+\sqrt{\frac{n}{N}}\tilde \bZ_{N,f_\ast}+\tilde O_p\Big(\frac{(d+K)\log n}{\sqrt{n}}\Big).
\end{align*}
Applying the Berry-Esseen bound to both $\bZ_{n,f_{\ast}}$ and $\tilde \bZ_{N,f_\ast}$,  we get 
\begin{align*}
\sup_{t\in\RR}&\bigg|\PP\bigg({\frac{\be_s^\top \bZ_{n,f_{\ast}}}{\sqrt{\be_s^\top\bW_{Y-f_{\ast}}\be_s}}}\leq t\bigg) -\Phi(t) \bigg|=O\bigg(\sqrt{\frac{d+K}{n}}\bigg)\\
\sup_{t\in\RR}&\bigg|\PP\bigg(\frac{\be_s^\top\tilde \bZ_{N,f_\ast}}{\be_s^\top\bW_{f_{\ast}-X^{\top}\theta}\be_{s}}\leq t\bigg)-\Phi(t) \bigg|=O\bigg(\sqrt{\frac{d+K}{N}}\bigg),
\end{align*}
for any $s\in[d]$.    

By the convergence rate for the sum of independent non-identically distributed random variables,  we get, for each $s\in[d]$, that 
\begin{align*}
\sup_t \bigg|\PP\bigg(\frac{\sqrt{n}\,\big(\hat \bmm_{\hat\beta_n}^{\moe}(\theta)\big)_{s}}{\sqrt{\be_s^\top\bW_{Y-f_{\ast}}\be_{s}+(n/N)\be_s^\top\bW_{f_{\ast}-X^{\top}\theta}\be_{s}}}\leq t\bigg)-\Phi(t) \bigg|=O\bigg(\frac{(d+K)\log n}{\sqrt{n}}+\sqrt{\frac{d+K}{N}}\bigg).
\end{align*}
The rest of the proof follows the same as the proof of Theorem~\ref{thm:logistic_estimation}.

\section{Experiments Result}

\subsection{Result of coverage and interval width}\label{sec:result_coverage}

Table~\ref{tab:full_results} reports the empirical coverage, interval width, width ratio, and coverage-agreement code for all methods across tasks and settings. Overall, for mean and quantile inference, the proposed PPI-MOE method achieves coverage close to the nominal 95\% level while producing intervals that are substantially shorter than those of the conventional estimator. This gain is especially pronounced in both linear and nonlinear settings, where the width ratio of PPI-MOE is often around 0.2--0.35 relative to the conventional baseline, indicating a large efficiency improvement without a noticeable loss in coverage.

For linear regression inference with adequate samples $n=500$, all methods behave reasonably well, and PPI-MOE remains competitive in terms of both coverage and interval width. However, when the samples size is limited to $200$, coverage deteriorates substantially for all methods, including the conventional estimator, with several entries falling well below the nominal level and receiving weak or empty coverage-agreement codes. This suggests that the issue is not specific to one particular prediction-powered construction, but rather reflects a broader instability under small samples and high dimensions.

For logistic regression inference, the results are more variable. In some settings, especially under the nonlinear design, PPI-MOE still attains near-nominal coverage with noticeably shorter intervals than the conventional method. In contrast, some alternative PPI procedures yield extremely large interval widths in the linear setting, suggesting numerical instability or near-separation phenomena in the underlying logistic regression fit. Relative to these unstable alternatives, PPI-MOE appears considerably more stable and practically usable.

Taken together, the results indicate that PPI-MOE delivers the clearest gains for mean and quantile inference, where it consistently maintains coverage while substantially reducing interval width. For regression coefficient inference, the benefit of MOE-powered prediction-powered methods depends more strongly on whether the inferential model is well specified. Nevertheless, PPI-MOE still adaptively chooses a suitable weight, moving toward the variance-minimizing choice when prediction-powered correction is beneficial and degenerating to the conventional estimator when it is not.

\small
\begin{longtable}{cccccccc}
\caption{Full Coverage and Width results across tasks. The width ratio is computed relative to the Conventional estimator within the same task, setting, and sample size. Column Code is defined in Table~\ref{tab:moes_summary_n500}.}
\label{tab:full_results}\\
\toprule
Task & Data Mode & \(n\) & Method & Cov. & Width & Ratio & Code \\
\midrule
\endfirsthead

\caption[]{Full Coverage and Width results across tasks (continued).}\\
\toprule
Task & Data Mode & \(n\) & Method & Cov. & Width & Ratio & Code \\
\midrule
\endhead

\midrule
\multicolumn{8}{r}{Continued on next page} \\
\endfoot

\bottomrule
\endlastfoot

\multicolumn{8}{l}{\textbf{Mean inference}} \\
\midrule
Mean & Linear & 200 & Conventional      & 0.934 & 45.9838 & 1.000 & ** \\
Mean & Linear & 200 & PPI-best          & 0.956 &  9.6054 & 0.209 & *** \\
Mean & Linear & 200 & PPI-mean          & 0.953 & 11.5188 & 0.250 & *** \\
Mean & Linear & 200 & PPI-worst         & 0.954 & 16.7861 & 0.365 & *** \\
Mean & Linear & 200 & \textbf{PPI-MOE}  & 0.952 &  9.4184 & 0.205 & *** \\

Mean & Linear & 500 & Conventional      & 0.960 & 29.1178 & 1.000 & ** \\
Mean & Linear & 500 & PPI-best          & 0.946 &  6.0713 & 0.209 & *** \\
Mean & Linear & 500 & PPI-mean          & 0.944 &  7.2855 & 0.250 & ** \\
Mean & Linear & 500 & PPI-worst         & 0.944 & 10.6361 & 0.365 & ** \\
Mean & Linear & 500 & \textbf{PPI-MOE}  & 0.940 &  5.9590 & 0.205 & ** \\

Mean & Nonlinear & 200 & Conventional     & 0.940 & 55.1793 & 1.000 & ** \\
Mean & Nonlinear & 200 & PPI-best         & 0.954 & 11.9046 & 0.216 & *** \\
Mean & Nonlinear & 200 & PPI-mean         & 0.949 & 14.3381 & 0.260 & *** \\
Mean & Nonlinear & 200 & PPI-worst        & 0.954 & 20.8877 & 0.379 & *** \\
Mean & Nonlinear & 200 & \textbf{PPI-MOE} & 0.944 & 11.5958 & 0.210 & ** \\

Mean & Nonlinear & 500 & Conventional     & 0.934 & 34.8191 & 1.000 & ** \\
Mean & Nonlinear & 500 & PPI-best         & 0.958 &  7.5351 & 0.216 & *** \\
Mean & Nonlinear & 500 & PPI-mean         & 0.948 &  9.0691 & 0.260 & *** \\
Mean & Nonlinear & 500 & PPI-worst        & 0.934 & 13.1799 & 0.379 & ** \\
Mean & Nonlinear & 500 & \textbf{PPI-MOE} & 0.940 &  7.3539 & 0.211 & ** \\

\midrule
\multicolumn{8}{l}{\textbf{Quantile inference (\(q=0.5\))}} \\
\midrule
Quantile & Linear & 200 & Conventional      & 0.954 & 61.5012 & 1.000 & *** \\
Quantile & Linear & 200 & PPI-best          & 0.954 & 18.4687 & 0.300 & *** \\
Quantile & Linear & 200 & PPI-mean          & 0.955 & 23.9068 & 0.389 & *** \\
Quantile & Linear & 200 & PPI-worst         & 0.954 & 34.3709 & 0.559 & *** \\
Quantile & Linear & 200 & \textbf{PPI-MOE}  & 0.954 & 18.7296 & 0.305 & *** \\

Quantile & Linear & 500 & Conventional      & 0.936 & 37.4121 & 1.000 & ** \\
Quantile & Linear & 500 & PPI-best          & 0.962 & 11.8119 & 0.316 & ** \\
Quantile & Linear & 500 & PPI-mean          & 0.949 & 15.1306 & 0.404 & *** \\
Quantile & Linear & 500 & PPI-worst         & 0.950 & 21.6758 & 0.579 & *** \\
Quantile & Linear & 500 & \textbf{PPI-MOE}  & 0.950 & 11.8886 & 0.318 & *** \\

Quantile & Nonlinear & 200 & Conventional     & 0.938 & 70.3177 & 1.000 & ** \\
Quantile & Nonlinear & 200 & PPI-best         & 0.950 & 24.2521 & 0.345 & *** \\
Quantile & Nonlinear & 200 & PPI-mean         & 0.940 & 30.2135 & 0.430 & ** \\
Quantile & Nonlinear & 200 & PPI-worst        & 0.926 & 42.4006 & 0.603 & . \\
Quantile & Nonlinear & 200 & \textbf{PPI-MOE} & 0.942 & 23.7573 & 0.338 & *** \\

Quantile & Nonlinear & 500 & Conventional     & 0.934 & 43.1535 & 1.000 & ** \\
Quantile & Nonlinear & 500 & PPI-best         & 0.952 & 15.2533 & 0.353 & *** \\
Quantile & Nonlinear & 500 & PPI-mean         & 0.938 & 19.1010 & 0.443 & ** \\
Quantile & Nonlinear & 500 & PPI-worst        & 0.922 & 26.7733 & 0.620 &  \\
Quantile & Nonlinear & 500 & \textbf{PPI-MOE} & 0.950 & 14.9006 & 0.345 & *** \\

\midrule
\multicolumn{8}{l}{\textbf{Linear regression coefficient (\(\hat{\beta}_1\))}} \\
\midrule
Lin.\ Reg. & Linear & 200 & Conventional      & 0.920 & 2.7534 & 1.000 &  \\
Lin.\ Reg. & Linear & 200 & PPI-best          & 0.920 & 2.7535 & 1.000 &  \\
Lin.\ Reg. & Linear & 200 & PPI-mean          & 0.918 & 5.3552 & 1.945 &  \\
Lin.\ Reg. & Linear & 200 & PPI-worst         & 0.910 & 9.0615 & 3.290 &  \\
Lin.\ Reg. & Linear & 200 & \textbf{PPI-MOE}  & 0.910 & 2.7072 & 0.983 &  \\

Lin.\ Reg. & Linear & 500 & Conventional      & 0.945 & 1.7679 & 1.000 & *** \\
Lin.\ Reg. & Linear & 500 & PPI-best          & 0.945 & 1.7679 & 1.000 & *** \\
Lin.\ Reg. & Linear & 500 & PPI-mean          & 0.933 & 3.4479 & 1.950 & . \\
Lin.\ Reg. & Linear & 500 & PPI-worst         & 0.940 & 5.8670 & 3.318 & ** \\
Lin.\ Reg. & Linear & 500 & \textbf{PPI-MOE}  & 0.930 & 1.7563 & 0.993 & . \\

Lin.\ Reg. & Nonlinear & 200 & Conventional     & 0.905 & 7.1322 & 1.000 &  \\
Lin.\ Reg. & Nonlinear & 200 & PPI-best         & 0.920 & 4.7344 & 0.664 &  \\
Lin.\ Reg. & Nonlinear & 200 & PPI-mean         & 0.914 & 8.4522 & 1.185 &  \\
Lin.\ Reg. & Nonlinear & 200 & PPI-worst        & 0.885 & 14.4147 & 2.021 &  \\
Lin.\ Reg. & Nonlinear & 200 & \textbf{PPI-MOE} & 0.910 & 4.5400 & 0.636 &  \\

Lin.\ Reg. & Nonlinear & 500 & Conventional     & 0.970 & 4.5206 & 1.000 & . \\
Lin.\ Reg. & Nonlinear & 500 & PPI-best         & 0.945 & 2.9830 & 0.660 & *** \\
Lin.\ Reg. & Nonlinear & 500 & PPI-mean         & 0.948 & 5.4124 & 1.197 & *** \\
Lin.\ Reg. & Nonlinear & 500 & PPI-worst        & 0.940 & 9.2308 & 2.042 & ** \\
Lin.\ Reg. & Nonlinear & 500 & \textbf{PPI-MOE} & 0.950 & 2.8814 & 0.637 & *** \\

\midrule
\multicolumn{8}{l}{\textbf{Logistic regression coefficient (\(\hat{\beta}_1\))}} \\
\midrule
Log.\ Reg. & Linear & 500 & Conventional      & 0.920 & 3.8692 & 1.000 & . \\
Log.\ Reg. & Linear & 500 & PPI-best          & 0.985 & 13.3119 & 3.441 & . \\
Log.\ Reg. & Linear & 500 & PPI-mean          & 0.990 & 102660.4034 & 26533.991 &  \\
Log.\ Reg. & Linear & 500 & PPI-worst         & 0.990 & 325903.3788 & 84231.167 &  \\
Log.\ Reg. & Linear & 500 & \textbf{PPI-MOE}  & 0.970 & 6.6481 & 1.718 & ** \\

Log.\ Reg. & Linear & 1000 & Conventional      & 0.885 & 2.1308 & 1.000 &  \\
Log.\ Reg. & Linear & 1000 & PPI-best          & 0.930 & 2.6450 & 1.241 & ** \\
Log.\ Reg. & Linear & 1000 & PPI-mean          & 0.952 & 7588.0250 & 3561.092 & *** \\
Log.\ Reg. & Linear & 1000 & PPI-worst         & 0.975 & 45510.8492 & 21357.191 & ** \\
Log.\ Reg. & Linear & 1000 & \textbf{PPI-MOE}  & 0.915 & 2.9004 & 1.361 & . \\

Log.\ Reg. & Nonlinear & 500 & Conventional     & 0.945 & 0.4654 & 1.000 & *** \\
Log.\ Reg. & Nonlinear & 500 & PPI-best         & 0.940 & 0.3535 & 0.760 & *** \\
Log.\ Reg. & Nonlinear & 500 & PPI-mean         & 0.951 & 0.4392 & 0.944 & *** \\
Log.\ Reg. & Nonlinear & 500 & PPI-worst        & 0.975 & 0.6089 & 1.308 & ** \\
Log.\ Reg. & Nonlinear & 500 & \textbf{PPI-MOE} & 0.960 & 0.2989 & 0.642 & *** \\

Log.\ Reg. & Nonlinear & 1000 & Conventional     & 0.935 & 0.3218 & 1.000 & ** \\
Log.\ Reg. & Nonlinear & 1000 & PPI-best         & 0.950 & 0.2401 & 0.746 & *** \\
Log.\ Reg. & Nonlinear & 1000 & PPI-mean         & 0.944 & 0.2973 & 0.924 & *** \\
Log.\ Reg. & Nonlinear & 1000 & PPI-worst        & 0.925 & 0.4074 & 1.266 & . \\
Log.\ Reg. & Nonlinear & 1000 & \textbf{PPI-MOE} & 0.940 & 0.2068 & 0.643 & *** \\

\end{longtable}
\normalsize
\end{document}